%% file: main.tex
\documentclass[11pt,a4paper,openright,pdftex,twoside]{report}
\usepackage{lmodern}
\usepackage[english]{babel}
\usepackage[utf8]{inputenc}
\usepackage[T1]{fontenc}
\usepackage[top=1.5in, bottom=1.5in, left=1.5in, right=1.0in]{geometry}
\usepackage[pdftex]{graphicx}
\usepackage[below]{placeins}
\usepackage[usenames]{color}
\usepackage{amssymb}
\usepackage{amsmath}
\usepackage[hang]{footmisc}
\usepackage{listings}
\usepackage{enumerate}
\usepackage{multirow}
\usepackage{multicol}
\usepackage[square,sort]{natbib}
\setcitestyle{aysep={}}
\usepackage{booktabs}       % professional-quality tables
\usepackage{subcaption}     % for subfigure
\usepackage{scrextend}
\usepackage{array}
\usepackage{fancyhdr}
\usepackage[toc,page]{appendix}
\usepackage{makeidx}
\usepackage{algorithm}
\usepackage{algorithmic}
\usepackage[pdftex,bookmarks,pdfpagelabels,hidelinks]{hyperref}
\usepackage[acronym]{glossaries}

\loadglsentries{abbreviations}
\makeglossaries

\graphicspath{{./}}

\makeindex

\usepackage{phdthesis}

\usepackage{afterpage}
\newcommand\blankpage{
    \null
    \thispagestyle{empty}
    \newpage}

\newtheorem{lemma}{Lemma}
\newtheorem{definition}{Definition}
\newcommand{\vect}[1]{\mathbf{#1}}

\renewcommand{\arraystretch}{1.1}

\AtBeginDocument{
  \renewcommand{\contentsname}{Table of contents}
  \renewcommand{\listtablename}{List of tables}
  \renewcommand{\listfigurename}{List of figures}
}

\begin{document}

  \include{titlepage}

  \pagenumbering{roman}

  \include{abstract_pl}
  \include{abstract_eng}
  \include{acknowledgements}
  \include{notation}

  \cleardoublepage
  \phantomsection
  \addcontentsline{toc}{chapter}{\contentsname}
  \tableofcontents
  \cleardoublepage

  \pagenumbering{arabic}

  \include{introduction}
  \include{background}
  \include{rp}
  \include{rp_layer}
  \include{rp_init}
  \include{conclusions}

  \appendix
  \include{datasets}

  \cleardoublepage
  \phantomsection
  \addcontentsline{toc}{chapter}{\listtablename}
  \listoftables

  \cleardoublepage
  \phantomsection
  \addcontentsline{toc}{chapter}{\listfigurename}
  \listoffigures

  \cleardoublepage
  \phantomsection
  \addcontentsline{toc}{chapter}{Acronyms}
  \printglossary[type=\acronymtype,style=long]

  \cleardoublepage
  \phantomsection
  \addcontentsline{toc}{chapter}{\bibname}
  \bibliographystyle{ACM-Reference-Format}
  \bibliography{bibliography}
  \printindex

  \include{publications}

\end{document}

%% file: abbreviations.tex
\newacronym{URL}{URL}{uniform resource locator}
\newacronym{GPGPU}{GPGPU}{general-purpose computing on graphics processing units}

\newacronym{BOW}{BOW}{bag-of-words}
\newacronym{TF-IDF}{TF-IDF}{term frequency–inverse document frequency}
\newacronym{AUC}{AUC}{area under the precision-recall curve}

\newacronym{MLP}{MLP}{multilayer perceptron}
\newacronym{SGD}{SGD}{stochastic gradient descent}
\newacronym{MSE}{MSE}{mean square error}
\newacronym{CE}{CE}{cross entropy}
\newacronym{BN}{BN}{batch normalization}
\newacronym{DBN}{DBN}{deep belief network}
\newacronym{RBM}{RBM}{restricted Boltzmann machine}
\newacronym{CD}{CD}{contrastive divergence}
\newacronym{ReLU}{ReLU}{rectified linear unit}
\newacronym{LReLU}{LReLU}{leaky rectified linear unit}
\newacronym{NReLU}{NReLU}{noisy rectified linear unit}
\newacronym{CNN}{CNN}{convolutional neural network}
\newacronym{DNN}{DNN}{deep neural network}
\newacronym{SI}{SI}{sparse initialization}
\newacronym{ResNet}{ResNet}{residual neural network}

\newacronym{RP}{RP}{random projection}
\newacronym{OSE}{OSE}{oblivious subspace embedding}
\newacronym{JLT}{JLT}{Johnson-Lindenstrauss transform}
\newacronym{SRHT}{SRHT}{subsampled randomized Hadamard transform}
\newacronym{RW-FNN}{RW-FNN}{random weight feedforward neural network}

\newacronym{PCA}{PCA}{principal component analysis}
\newacronym{SVD}{SVD}{singular value decomposition}
\newacronym{IG}{IG}{information gain}
\newacronym{LR}{LR}{logistic regression}
\newacronym{SVM}{SVM}{support vector machine}
\newacronym{PmSVM}{PmSVM}{power mean support vector machine}
\newacronym{PmSVM-LUT}{PmSVM-LUT}{power mean support vector machine with look-up tables}
\newacronym{LDA}{LDA}{linear discriminant analysis}
\newacronym{MDS}{MDS}{multidimensional scaling}

\newacronym{MMC}{MMC}{maximum margin criterion}
\newacronym{CCA}{CCA}{canonical correlations analysis}
\newacronym{MAF}{MAF}{maximum autocorrelation factors}
\newacronym{SFA}{SFA}{slow feature analysis}
\newacronym{SDR}{SDR}{sufficient dimensionality reduction}
\newacronym{LPP}{LPP}{locality preserving projections}
\newacronym{ICA}{ICA}{independent component analysis}
\newacronym{IMMC}{IMMC}{incremental maximum margin criterion}
\newacronym{IPCA}{IPCA}{incremental principal component analysis}
\newacronym{CCIPCA}{CCIPCA}{candid covariance-free incremental principal component analysis}
\newacronym{IDR/QR}{IDR/QR}{incremental dimension reduction via QR decomposition}
\newacronym{ILDA}{ILDA}{incremental linear discriminant analysis}
\newacronym{SRDA}{SRDA}{spectral regression discriminant analysis}
\newacronym{SMMP}{SMMP}{sparse matrix multiplication package}
\newacronym{VW}{VW}{Vowpal Wabbit}
\newacronym{BM}{BM}{block minimization}
\newacronym{SBM}{SBM}{selective block minimization}
\newacronym{ADMM}{ADMM}{alternating direction method of multipliers}
\newacronym{DADM}{DADM}{dual alternating direction method}
\newacronym{AMM}{AMM}{adaptive multi-hyperplane machine}

\newacronym{CSR}{CSR}{compressed sparse row}
\newacronym{CSC}{CSC}{compressed sparse column}
\newacronym{SNP}{SNP}{single nucleotide polymorphism}

%% file: titlepage.tex
\begin{titlepage}
    \begin{center}
        \vspace{2cm} 
        \textbf{\large Akademia Górniczo-Hutnicza} \\
        \vspace{0.2cm}
        \textbf{\large im. Stanisława Staszica w Krakowie} \\
        \vspace{0.6cm}
        {\large Wydział Informatyki, Elektroniki i Telekomunikacji} \\
        \vspace{0.2cm}
        {\large Katedra Informatyki} \\
        \vspace{1.0cm}
        \includegraphics[width=66pt]{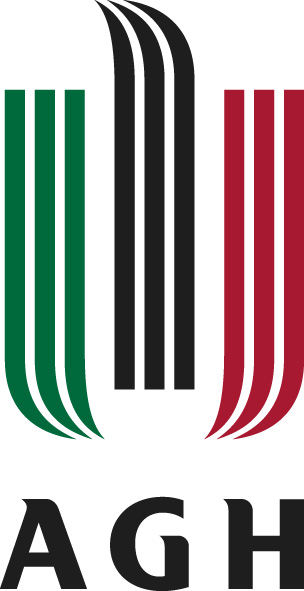} \\
        \vspace{1.8cm}

        {\large \textsc{Rozprawa doktorska}} \\
        \vspace{1.0cm}
        \textbf{\LARGE Zastosowania metody rzutu przypadkowego \\
        \vspace{0.2cm}
                       w głębokich sieciach neuronowych} \\
        \vspace{1.0cm}
        {\Large \textsc{Piotr Iwo Wójcik}} \\
        \vspace{2.0cm}

    \end{center}

    \noindent
    \textbf{Promotor:} \\
    \indent prof. dr hab. inż. Witold Dzwinel \\
    \vspace{0.5cm} \\
    \textbf{Promotor pomocniczy:} \\
    \indent dr inż. Marcin Kurdziel \\
    \vspace{0.5cm} \\
    \begin{center}
        Kraków, 2018
    \end{center}
    \afterpage{\blankpage}
\end{titlepage}

\begin{titlepage}
    \begin{center}
        \vspace{2cm} 
        \textbf{\large AGH} \\
        \vspace{0.2cm}
        \textbf{\large University of Science and Technology in Kraków } \\
        \vspace{0.6cm}
        {\large Faculty of Computer Science, Electronics and Telecommunications} \\
        \vspace{0.2cm}
        {\large Department of Computer Science} \\
        \vspace{1.0cm}
        \includegraphics[width=66pt]{agh.jpg} \\
        \vspace{1.2cm}

        {\large \textsc{Dissertation for the degree of\\ Doctor of Philosophy}} \\
        \vspace{0.8cm}
        \textbf{\LARGE Random Projection \\
        \vspace{0.2cm}
                       in Deep Neural Networks} \\
        \vspace{1.0cm}
        {\Large \textsc{Piotr Iwo Wójcik}} \\
        \vspace{2.2cm}

    \end{center}

    \noindent
    \textbf{Supervisor:} \\
    \indent Witold Dzwinel, Ph.D., Professor \\
    \vspace{0.5cm} \\
    \textbf{Co-supervisor:} \\
    \indent Marcin Kurdziel, Ph.D. \\
    \vspace{0.5cm} \\
    \begin{center}
         Kraków, 2018
    \end{center}
    \afterpage{\blankpage}
\end{titlepage}

%% file: abstract_pl.tex
\cleardoublepage
\chapter*{Streszczenie}

Niniejsza praca prezentuje zastosowania metody rzutu przypadkowego (RP) w głębokich sieciach neuronowych. W pracy
skupiono się na dwóch obszarach, w których użycie metody RP poprawia ich skuteczność: na efektywnym uczeniu głębokich
sieci na danych wysokowymiarowych oraz na inicjalizacji parametrów sieci.~Rozważono kilka klasycznych oraz niedawno
zaproponowanych konstrukcji macierzy RP: macierze Gaussa, Achlioptasa i Li oraz metody subsampled randomized Hadamard
transform (SRHT) i Count Sketch.

W pierwszym z rozważanych obszarów zastosowań metoda RP jest włączana do architektury sieci jako warstwa wejściowa
(warstwa RP). Umożliwia to efektywne uczenie głębokich sieci neuronowych na danych pozbawionych struktury,
reprezentowanych przez rzadkie, wysokowymiarowe wektory cech.~Do tej pory analiza takich danych przy pomocy sieci
neuronowych była trudna, lub wręcz niemożliwa, ze względu na wysoki koszt obliczeniowy wynikający z ogromnej liczby wag
w pierwszej warstwie sieci.~W pracy pokazano, że dzięki użyciu warstwy wejściowej której wagi zostały zainicjalizowane
elementami macierzy RP możliwe jest efektywne trenowanie głębokich sieci na tego typu danych. Zostały rozważone dwa
warianty zaproponowanej warstwy RP: z ustalonymi wagami oraz z wagami douczanymi w trakcie treningu.~Przedstawiono
również kilka modyfikacji architektury sieci oraz metod jej trenowania, dzięki którym możliwe jest uczenie sieci na
danych zawierających dziesiątki milionów przykładów uczących o wymiarowości przekraczającej miliony cech. Pozwoliło to
uzyskać wyniki porównywalne lub lepsze od najlepszych wyników publikowanych w literaturze dla kilku dużych problemów
klasyfikacji danych wielowymiarowych.~Eksperymenty z różnymi konstrukcjami RP pokazały również, że najlepsze wyniki
osiągają sieci z douczaną warstwą RP typu Count Sketch.

W drugim obszarze zastosowań macierz RP wykorzystana jest do inicjalizacji wag sieci neuronowej.~Inicjalizacja
parametrów sieci przy pomocy elementów macierzy rzutu przypadkowego pozwoliła poprawić skuteczność residualnych sieci
konwolucyjnych -- modeli osiągających obecnie najlepsze wyniki w dziedzinie rozpoznawania obrazów.~Eksperymenty
wykazały, że najwyższą skuteczność osiągają sieci inicjalizowane gęstymi macierzami RP, których kolumny są bliskie
ortogonalnym (np. konstrukcja SRHT).

%% file: abstract_eng.tex
\cleardoublepage
\chapter*{Abstract}

This work investigates the ways in which deep learning methods can benefit from random projection (RP), a classic linear
dimensionality reduction method. We focus on two areas where, as we have found, employing RP techniques can improve deep
models: training neural networks on high-dimensional data and initialization of network parameters. We consider several
recently proposed RP schemes: Gaussian, Achlioptas’, Li’s, subsampled randomized Hadamard transform (SRHT) and Count
Sketch-based constructions.

Training deep neural networks (DNNs) on sparse, high-dimensional data with no exploitable structure poses a major
computational challenge. It implies a network architecture with an input layer that has a huge number of weights, which
often makes training infeasible.~We show that this problem can be solved by prepending the network with an input layer
whose weights are initialized with an RP matrix. We study cases where the weights of this RP layer are either fixed or
learned during training. Furthermore, we propose several modifications to the network architecture and training regime
that makes it possible to efficiently train DNNs with learnable RP layer on data with as many as tens of millions of
input features and training examples. In comparison to the state-of-the-art methods, neural networks with RP layer
achieve competitive performance or improve the results on several extremely high-dimensional real-world datasets.~Our
results also demonstrate that, out of the evaluated RP methods, Count Sketch is the overall best construction for DNNs
with RP layer.

The second area where the application of RP techniques can be beneficial for training deep models is weight
initialization. Specifically, we study setting the initial weights in DNNs to elements of various RP matrices instead of
drawing them from a scaled normal distribution, as is done in current state-of-the-art initialization techniques. Such
RP initialization enables us to train deep networks to higher levels of performance.~In particular, our results show
that dense orthogonal RP initialization schemes, such as SRHT, improve the performance of residual convolutional neural
networks.

%% file: acknowledgements.tex
\cleardoublepage
\phantomsection
\addcontentsline{toc}{chapter}{Acknowledgments}
\chapter*{Acknowledgments}

I would like to express my deepest gratitude to Professor Witold Dzwinel for guiding and encouraging me during my
research efforts. I also wish to thank Doctor Marcin Kurdziel for helpful discussion and invaluable advice that greatly
improved this thesis.

I dedicate this thesis to my wife, Joanna. I know she will never read past this page, but still.

The research for this thesis was supported by:

\begin{itemize}
\item the Polish National Science Centre (NCN) grant No. \mbox{DEC-2013/09/B/ST6/01549} ``Interactive Visual Text
Analytics (IVTA): Development of novel, user-driven text mining and visualization methods for large text corpora
exploration.''
\item the ``HPC Infrastructure for Grand Challenges of Science and Engineering'' Project, co-financed by the European
Regional Development Fund under the Innovative Economy Operational Programme,
\item the PL-Grid Infrastructure.
\end{itemize}

%% file: notation.tex
\cleardoublepage
\phantomsection
\addcontentsline{toc}{chapter}{Notation}
\chapter*{Notation}

{\renewcommand{\arraystretch}{1.4}% for the vertical padding
\begin{tabular}{cl}

% \multicolumn{2}{c}{\textbf{Numbers and arrays}} \\
$x$ & a scalar \\
$\vect{x}$ & a vector \\
$\mathbf{X}$ & a matrix \\
$\mathbb{X}$ & a set \\
$\mathbb{R}$ & the set of real number \\
$\left\vert{\mathbb{X}}\right\vert$ & the number of elements in $\mathbb{X}$ \\
$[x,y]$ & the real interval including $x$ and $y$ \\
$\mathbf{X}^{T}$ & transpose of matrix $\mathbf{X}$ \\
$\|\vect{x}\|_1$ & L1 norm of vector $\vect{x}$ \\
$\|\vect{x}\|_2$ & L2 norm of vector $\vect{x}$ \\

$x_i$ & $i$-th element of vector $\vect{x}$ \\
$X_{ij}$ & element $i,j$ of matrix $\mathbf{X}$ \\
$\mathbf{X}_{i \cdot}$ & $i$-th row of matrix $\mathbf{X}$ \\
$\mathbf{X}_{\cdot j}$ & $j$-th column of matrix $\mathbf{X}$ \\

$\vect{x}^{(i)}$ & $i$-th example from a dataset \\
$y^{(i)}$ & label (target output) for $i$-th example \\
$\mathbf{X}$ & design matrix of a dataset with example $\vect{x}^{(i)}$ in row $\mathbf{X}_{i \cdot}$ \\

$\frac{\partial y}{\partial x}$ & partial derivative of $y$ with respect to $x$ \\
$\nabla_{x}y$ & gradient of $y$ with respect to $x$ \\
$f(\vect{x}; \vect{y})$ & a function of $\vect{x}$ parametrized by $\vect{y}$ \\

$a \sim P$ & random variable $a$ has distribution $P$ \\
$\mathcal{N}\left(m, s^2\right)$ & a Gaussian distribution with mean $m$ and variance $s^2$ \\

\end{tabular}
\renewcommand{\arraystretch}{1.1}
\afterpage{\blankpage}

%% file: introduction.tex
\cleardoublepage
\chapter{Introduction}
\label{cha:introduction}

In this work we investigate the ways in which deep learning methods can benefit from \gls{RP}, a classic linear
dimensionality reduction method. In particular, we focus on two areas where, as we have found, employing \gls{RP}
techniques can enhance deep models.

In the first application of \acrlong{RP}, we make use of its original purpose, i.e., reducing the dimensionality of the
input data. We show how this can be useful in the problem of learning from data that is represented by sparse,
unstructured, high-dimensional feature vectors\footnote{While the term ``high-dimensional'' is sometimes used to refer
to data described with at least four features, here we consider a feature vector high-dimensional when its
dimensionality is on the order of millions.}. This type of data often arises in areas such as social media, web
crawling, gene sequencing or biomedical analysis. Currently, training \glspl{DNN} or other complex nonlinear models is
practically infeasible for similar applications. Therefore, simpler but faster linear approaches, such as
\gls{SVM} or \gls{LR} classifiers~\citep{yuan2012recent} are usually employed. Importantly, these methods are capable of
efficiently processing sparse, high-dimensional input data. With the assistance of \gls{RP}, we hope to narrow this gap
and enable deep networks to be trained on such problematic type of data.

The dimensionality of the input data in most modern neural network applications is relatively low. For example, networks
trained for speech recognition tasks employ input vectors with the size on the order of hundreds of
dimensions~\citep{graves2013speech}. Learning with larger input dimensionality typically requires some structure in the
input data. This is the case in \glspl{CNN} trained on images, which can work with up to hundred thousand input pixels.
This architecture takes advantage of the spatial structure of images by exploiting the local pixel connectivity and
sharing the weights between spatial locations, which greatly reduces the number of learnable parameters. However, with
no exploitable structure in the data, training \glspl{DNN} on high-dimensional data poses a severe computational
problem. The reason for this is the implied network architecture and in particular, a huge input layer, which may
contain billions of weights. Even with recent advances in \gls{GPGPU}, training networks with that number of parameters
is infeasible.

We show that this problem can be solved by incorporating \acrlong{RP} into the network architecture. In particular, we
propose to prepend the network with an input layer whose weights are initialized to elements of an \gls{RP} matrix. We
study cases where the weights of this \gls{RP} layer are either fixed during training or finetuned with error
backpropagation. Our results demonstrate that, in comparison to the state-of-the-art methods, neural networks with
\gls{RP} layer achieve competitive performance on extremely high-dimensional real-world datasets.

The second, less conventional area, where we have found the application of \gls{RP} techniques to be beneficial for
training deep models is weight initialization. Specifically, we initialized the weights in deep networks with various
\gls{RP} matrices instead of drawing them from a scaled normal distribution, as is done in the current state-of-the-art
initialization technique~\citep{he2015delving}. Such \acrlong{RP} initialization enabled us to train deep networks to
higher levels of performance: our experiments suggest that particularly deep \glspl{CNN} can benefit from the introduced
method.

\section{Thesis statement}

The goal of this dissertation is to show that \acrlong{RP} methods can be beneficial in training \acrlongpl{DNN}. The
dissertation thesis is:

\hskip 1em
\begin{addmargin}[4em]{4em}
\textit{Random Projection enables training Deep Neural Networks on sparse, unstructured data with millions of
dimensions. Furthermore, when used as a weight initialization method it improves the network performance.}\\
\end{addmargin}

Primarily, the dissertation presents how we can efficiently incorporate \gls{RP} as an input layer in deep networks.
This broadens their applicability to types of input data that currently can only be learned with fast linear
classifiers. Additionally, the dissertation shows that \gls{RP} can be successfully applied as a method for initializing
weights in deep models.

\section{Research contribution}

The main contributions of this dissertation are:
\begin{itemize}
\item a review of the challenges and existing approaches to training \glspl{DNN} on large-scale data that is sparse,
high-dimensional and unstructured;
\item the proposition of the fixed-weight \acrlong{RP} layer that enables efficiently training deep networks on sparse,
high-dimensional, unstructured data;
\item the proposition of network architectures and training regimes that make finetuning the weights in the \gls{RP}
layers feasible, even on large-scale datasets;
\item the proposition of initializing weights in deep networks with \gls{RP} matrices;
\item an implementation of the proposed methods and their experimental evaluation on both synthetic and real-world
large-scale datasets, including a comparison with the current state-of-the-art approaches.
\end{itemize}

\section{Thesis structure}

The dissertation is organized as follows.

Chapter~\ref{cha:background} consists of two parts. In the first part, we introduce deep learning models and related
training methods, which we extensively use in this work. In the second part, we present a particularly difficult type of
data for neural network models -- data that is sparse, high-dimensional and unstructured. We survey existing techniques
that, by reducing the data dimensionality, can make training deep networks on such data possible.

In Chapter~\ref{cha:rp} we present in detail one of these methods, which is the core of the network architecture that we
introduce in the following chapter -- \acrlong{RP}. We review several important \gls{RP} constructions: Gaussian,
Achlioptas', Li's, \acrlong{SRHT} and Count Sketch. We analyze their properties, focusing on the embedding quality and
computational cost of performing the projection.

In Chapter~\ref{cha:rp_layer} we show how to incorporate \gls{RP} into the architecture of \glspl{DNN} to enable them to
learn from sparse, high-dimensional, unstructured data. We evaluate the performance of such networks on synthetic and
real-world datasets. We compare the effectiveness and computational cost of our approach with competing state-of-the-art
techniques. Finally, we discuss selected important implementation details.

In Chapter~\ref{cha:rp_init} we motivate and study initializing weights in \glspl{DNN} with elements of \gls{RP}
matrices. We evaluate \gls{RP} initialization in \glspl{CNN} and in pretrained, fully-connected networks on several
real-world datasets.

Finally, in Chapter~\ref{cha:conclusions} we conclude the dissertation and discuss further directions for research.

%% file: background.tex
\cleardoublepage
\chapter{Background}
\label{cha:background}

In this chapter we first briefly introduce several important \gls{DNN} models, algorithms and architectures.
Specifically, we focus on models that we employ later in this work: \glspl{MLP}, \glspl{DBN}, Autoencoders and
\glspl{CNN}. In the second part we focus on the problem of training deep networks on data that is simultaneously sparse,
unstructured and high-dimensional. We show where data with these three properties may arise and why learning from it
proves to be a challenging task. We explain how this can be leveraged by using fast dimensionality reduction techniques.
Finally, we review existing dimensionality reduction approaches that are capable of efficiently processing sparse,
high-dimensional data.

\input{background_dl}
\input{background_dr}

%% file: background_dl.tex
\section{Deep neural networks}

For years neural networks have been attracting the attention of researchers in both academia and industry. Unlike
conventional machine learning techniques, they do not require handcrafted features, but instead discover features during
learning. Yet, for a long time, training networks with a larger number of layers, called deep networks, was
unsuccessful, and simpler machine learning algorithms, like support vector machines~\citep{cortes1995support}, were more
useful in practical applications. However, advances from the last decade led to a resurgence of interest in neural
networks. Since then, \glspl{DNN} have demonstrated impressive results, significantly pushing the state of the art on
many difficult tasks such as image recognition~\citep{simonyan2014very,huang2016deep}, speech
recognition~\citep{hinton2012speech,sercu2016very} or sequence
modeling~\citep{mikolov2013distributed,sutskever2011generating,graves2013speech,sutskever2014sequence}.

During the recent years many types of artificial neural networks have been proposed, e.g., feedforward neural networks,
recurrent neural networks, radial basis function networks or convolutional neural networks~\citep{goodfellow2016deep}.
Here we focus on networks that fall into the first, arguably most popular category, i.e., feedforward networks. Below we
briefly introduce the most important feedforward models and architectures used in this work.

\subsection{Multilayer perceptron}

Deep feedforward networks are the backbone of modern deep learning methods. The most important architecture in this
category is the \acrlong{MLP}. In fact, some authors consider the terms \textit{multilayer perceptron} and \textit{deep
feedforward network} as synonyms~\citep{goodfellow2016deep}.

The aim of an \gls{MLP} is to find the best approximation $f^*$ of a function $f$ that maps the information given to the
network on input into the desired output. For example, in the image classification task, the input may correspond to
pixel intensities of an image and the output may correspond to the category of the input image. In this case, by using
observed data, \gls{MLP} learns to map input images to the output categories in order to be able to predict the
categories of previously unseen images.

\begin{figure}[htb!]
  \centering
  \includegraphics[width=0.8\linewidth]{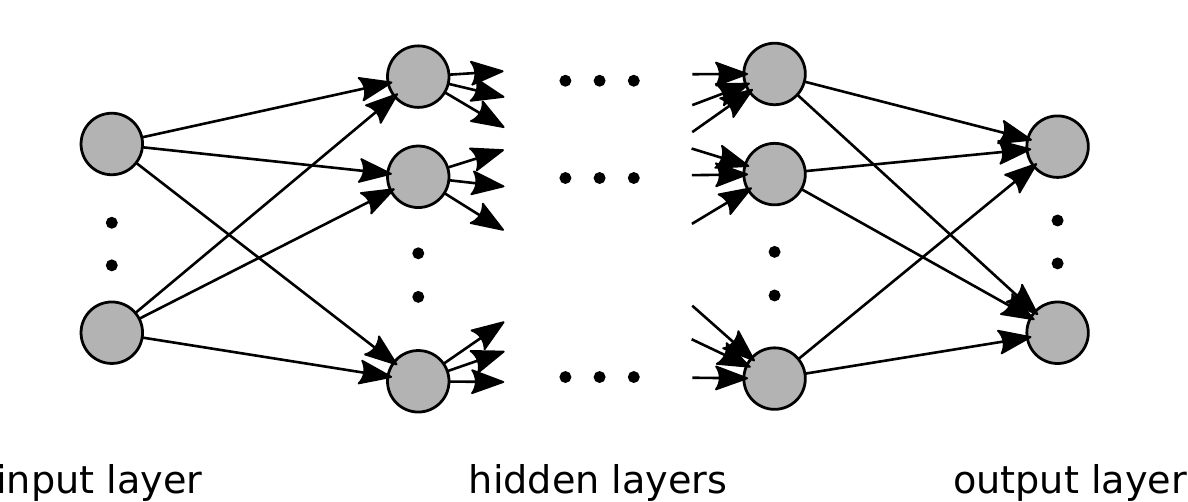}
  \caption{A schematic representation of a multilayer perceptron.}
  \label{fig:mlp}
\end{figure}

The mapping from the input to the output is realized by feeding the input signal through multiple layers of
computational nodes. Each node in one layer has weighted connections directed to the nodes of the subsequent layer
(Fig.~\ref{fig:mlp}). \Acrlong{MLP} is called a feedforward network because of the flow of computations that are
performed when processing information. Specifically, the input data is first fed into the input layer, where the inputs
are multiplied by connection weights as they are passed to the first hidden layer. After the information is processed in
the first hidden layer, it is again multiplied and passed to the subsequent layer. This process is repeated until the
output layer is reached. Importantly, the information flows in one direction, forward, because there are no backward
connections. In this aspect feedforward neural networks differ from recurrent neural networks.

\subsubsection{Activation function}

Except for the input nodes, each node is a neuron that performs a simple computation:
\begin{equation}
  y = \phi(z), \quad
  z = \sum_i{w_i x_i} + b,
\end{equation}
where $y$ is the output, $x_i$ is the $i$-th input, $w_i$ is its corresponding weight, $b$ is the bias, and $\phi$ is
the activation function. Historically, a popular choice for the activation function, also called the transfer function,
was the logistic sigmoid function:
\begin{equation}
  \phi(z) = \frac{1}{1 + e^{-z}}
\end{equation}
or the hyperbolic tangent function:
\begin{equation}
  \phi(z) = \tanh(z) = \frac{e^{z} - e^{-z}}{e^{z} + e^{-z}}.
\end{equation}
Nowadays the recommendation is to use the rectifier linear function~\citep{nair2010rectified}, defined as:
\begin{equation}
  \phi(z) = \max\{0, z\}.
\end{equation}
Neurons employing this activation function are commonly called \glspl{ReLU}. There are also several variants of the
rectifier linear function, e.g., the leaky rectifier linear function~\citep{maas2013rectifier}, defined as:
\begin{equation}
  \phi(z) = \max\{az, z\},
\end{equation}
where $a$ is usually small, e.g., $a=0.01$. A \gls{LReLU} works similarly to a \acrlong{ReLU}, but propagates a small,
non-zero gradient when the unit is not active.

\subsubsection{Training with stochastic gradient descent}

In order for the \gls{MLP} to accurately approximate the mapping function $f$, its parameters $\boldsymbol{\theta}$,
i.e., weights $\vect{w}$ and biases $\vect{b}$, have to be adjusted. This process of adjusting network parameters is
called network training. \Acrlong{MLP} networks are usually trained in a supervised manner. That is, during training the
network is presented with data examples $\vect{x}^{(i)}$ with known labels $y^{(i)}$. By knowing the desired output and
the output computed by the network $f^*(\vect{x}^{(i)})$, the value of some per-example loss function
$J_i(\vect{x}^{(i)}, y^{(i)}; \boldsymbol{\theta})$ can be calculated. This value indicates how well the model is
approximating the mapping function $f$ for the $i$-th example. The cost function, also called the objective function is
the average loss over individual examples:
\begin{equation}
  J(\mathbf{X}, \mathbf{y}; \boldsymbol{\theta}) = \frac{1}{n} \sum_{i=1}^n J_i(\vect{x}^{(i)}, y^{(i)};
  \boldsymbol{\theta}),
\end{equation}
where $\mathbf{X}$ is the design matrix of the dataset\footnote{A design matrix of a dataset is a matrix in which each
row represents a training example.}, $\mathbf{y}$ is the vector of labels, and $n$ is the number of training examples.
The goal of the training is to minimize the value of the objective function on the training examples $\mathbf{X}$ by
adjusting the values of parameters $\boldsymbol{\theta}$. Unfortunately, this cannot be done analytically. However, it
is possible to compute the gradient of the objective function with respect to each parameter $\theta \in
\boldsymbol{\theta}$: $\frac{\partial J}{\partial \theta}$. This can be done starting from the output layer, by applying
the chain rule for derivatives. The procedure for calculating the gradients is called
\textbf{backpropagation}~\citep{rumelhart1988learning}. When the gradients are known, we can use a gradient-based
optimization algorithm to find a configuration of parameters $\boldsymbol{\theta}$ that minimizes the objective
function. In practice, the \textbf{\gls{SGD} algorithm}, a stochastic approximation of the gradient descent optimization
method, works surprisingly well. In \gls{SGD} the true gradient $\nabla_{\boldsymbol{\theta}} J(\mathbf{X},
\mathbf{y}; \boldsymbol{\theta})$, i.e., the average gradient calculated over all training examples, is approximated by
a gradient computed on a single training example. Pseudocode for \gls{SGD} is presented in Algorithm~\ref{alg:sgd}.
\begin{algorithm}
\caption{Stochastic gradient descent pseudocode.}
\label{alg:sgd}
\begin{algorithmic}
    \STATE initialize network parameters $\boldsymbol{\theta}$
    \STATE choose learning rate $\gamma$
    \WHILE{not converged}
        \STATE shuffle training examples
        \FOR{each training example $(\vect{x}^{(i)}, y^{(i)})$}
            \STATE compute $\nabla_{\boldsymbol{\theta}} J_i(\vect{x}^{(i)}, y^{(i)}; \boldsymbol{\theta})$ with
              backpropagation
            \STATE $\boldsymbol{\theta} \Leftarrow \boldsymbol{\theta} - 
              \gamma \nabla_{\boldsymbol{\theta}} J_i(\vect{x}^{(i)}, y^{(i)}; \boldsymbol{\theta})$
        \ENDFOR
    \ENDWHILE
\end{algorithmic}
\end{algorithm}

However, performing the parameters update with a single training example is computationally inefficient. To speed up the
training, most practitioners compute the average gradient over several training examples and then update the weights.
This modification is called the mini-batch \acrlong{SGD}~\citep{bottou1998online}. Employing mini-batch \gls{SGD}
instead of the classic \gls{SGD} is beneficial in two ways. First, mini-batch \gls{SGD} can be parallelized more
efficiently than \gls{SGD}. This is because the majority of operations in mini-batch \gls{SGD} involve matrix-matrix
operations, while \gls{SGD} utilizes mostly vector-matrix operations. Performing one matrix-matrix operation, e.g.,
matrix multiplication to compute activations for a mini-batch of 100 examples, is significantly faster than an
equivalent number of vector-matrix multiplications. Second, mini-batch \gls{SGD} leads to smoother convergence since the
gradient estimates are less noisy than the gradients estimated using individual examples~\citep{bousquet2008tradeoffs}.

\subsubsection{Momentum method}

The momentum method~\citep{polyak1964some} is a particularly important \gls{SGD} extension, which usually improves the
speed of convergence of \glspl{DNN}. \Acrlong{SGD} with momentum stores the values of the parameter updates at each
iteration and uses them in the next update. The parameter update in \gls{SGD} with momentum is given by:
\begin{equation}
  \begin{split}
  & \mathbf{v} \Leftarrow \mu \mathbf{v} - \gamma \nabla_{\boldsymbol{\theta}} J(\mathbf{X}, \mathbf{y};
  \boldsymbol{\theta}) \\
  & \boldsymbol{\theta} \Leftarrow \boldsymbol{\theta} + \mathbf{v},
  \end{split}
\end{equation}
where $\vect{v}$ is the velocity vector of the same size as the parameter vector $\boldsymbol{\theta}$ and $\mu$ is an
additional hyperparameter, usually referred to as \textit{momentum}. In the physical interpretation $\mu$ is similar to
the friction coefficient. Its value is usually set between $0.5$ and $0.99$. In practice, by taking into account
gradients from the previous updates, the momentum method accelerates \gls{SGD} learning, especially for gradients
$\nabla_{\boldsymbol{\theta}} J(\mathbf{X}, \mathbf{y}; \boldsymbol{\theta})$ that are noisy or small but consistent.

\subsubsection{Output units and cost functions}

In a supervised learning problem, the cost function quantifies the error a network makes by comparing the network
prediction with the expected output. The way in which a network represents its output determines the type of the loss
function. Therefore, we discuss output unit types together with corresponding loss functions. The most popular choices
of the output units include linear units, sigmoid units and softmax units.

\paragraph{Linear units.}

Linear units do not employ a nonlinear transfer function and return real-valued outputs. This makes them suitable for
regression tasks. Because linear units do not saturate, they can be used with any gradient-based optimization algorithm.
Most often the \gls{MSE} cost function, also known as the quadratic cost, is used along with linear outputs. The loss
for a singe example $(\vect{x}^{(i)}, y^{(i)})$ is then defined as:
\begin{equation}\label{eq:mse}
  J_i(\vect{x}^{(i)}, y^{(i)}; \boldsymbol{\theta}) = \frac{1}{2} \sum_{j} \left(a^{(i)}_j - y^{(i)}_j\right)^2,
\end{equation}
where $a^{(i)}_j$ represents the activation value of the $j$-th neuron in the output layer when the network is presented
with $i$-th example, and $y^{(i)}_j$ is the $j$-th component of the desired output for the $i$-th example.

\paragraph{Sigmoid units.}

Sigmoid units, on the other hand, are more suited for the classification task. Specifically, a single sigmoid unit in an
output layer can be used for binary classification since its output $\sigma(z) = \frac{1}{1 + e^{-z}} \in [0, 1]$ can be
interpreted as class probability. In this context, rather than \gls{MSE}, a more appropriate cost function is the
\gls{CE} loss:
\begin{equation}\label{eq:ce_binary}
  J_i(\vect{x}^{(i)}, y^{(i)}; \boldsymbol{\theta}) = -\left( a^{(i)} \ln y^{(i)} + (1- a^{(i)})\ln(1 - y^{(i)})\right),
\end{equation}
where $y^{(i)}$ is the desired output, and $a^{(i)}$ is the output produced by the sigmoid unit.

\paragraph{Softmax units.}

For the multi-class classification task, the network has to learn a categorical distribution over $n$ possible
categories. In these settings, the softmax function~\citep{bridle90probabilistic} is a perfect choice. To represent a
valid probability distribution, each unit in a softmax layer is required to output a value in the $[0, 1]$ interval and
the outputs must sum up to $1$. A popular function that satisfies these conditions is softmax:
\begin{equation}
  y(\mathbf{z})_i = \frac{e^{z_i}}{\sum_{j=1}^n e^{z_j}},
\end{equation}
where $\vect{z} = (z_1, \ldots, z_n)$ is a vector of inputs to neurons in the softmax layer. For multiclass
classification, the \gls{CE} cost from Eq.~\ref{eq:ce_binary} can be extended to:
\begin{equation}
  J_i(\vect{x}^{(i)}, y^{(i)}; \boldsymbol{\theta}) = -\sum_{j=1}^C a^{(i)}_j \ln{y^{(i)}_j},
\end{equation}
where $C$ is the total number of classes.

\subsubsection{Regularization}

The number of parameters in a neural network can be very high, which often makes the training process prone to
overfitting. One way to avoid overfitting is to employ regularization. There are several means of regularizing neural
networks. These include, for example, penalizing the magnitude of network parameters or employing more complex
techniques, such as dropout.

\paragraph{Parameter penalty.}

Probably the simplest way to regularize a neural network model is to impose a penalty on the magnitudes of its
parameters. This can be realized by adding a parameter norm penalty term $\omega(\boldsymbol{\theta})$ to the cost
function:
\begin{equation}
  \tilde{J}(\mathbf{X}, \mathbf{y}; \boldsymbol{\theta}) = J(\mathbf{X}, \mathbf{y}; \boldsymbol{\theta}) +
  \lambda\omega(\boldsymbol{\theta}),
\end{equation}
where $\lambda$ is a hyperparameter controlling the regularization strength. The most common parameter penalty norms are
the L2 regularization $\omega(\boldsymbol{\theta}) = \frac{1}{2} \|\vect{w}\|_2^2$ and L1 regularization
$\omega(\boldsymbol{\theta}) = \|\vect{w}\|_1$. Note that typically only the weights $\vect{w}$ (and not the biases) are
penalized~\citep{goodfellow2016deep}.

\paragraph{Dropout.}

Regularization techniques such as parameter norm penalties are not specific to neural networks. They have been used
extensively, for example, in linear regression or logistic regression models, prior to the advent of deep learning.
Dropout~\citep{srivastava2014dropout}, however, is a recent regularization technique tailored specifically for reducing
overfitting in \glspl{DNN}.

Dropout can be applied during training with a mini-batch-based learning algorithm, such as mini-batch \acrlong{SGD}. It
amounts to disabling a randomly selected subset of units each time a mini-batch is processed. The neurons are kept alive
with probability $d$, which is usually set to $0.5$ for hidden units and $0.8$ for input units.

Dropout can be viewed as an extreme form of bagging~\citep{breiman1996bagging} -- an ensemble learning technique in
which each member of the ensemble is trained with a different subsample of the input data. For each mini-batch, dropout
creates a different network that is trained on examples from just this single mini-batch. At test time, network with
scaled weights is used and no units are dropped. Mathematically, this approximates ensemble
averaging~\citep{warde2013empirical}.

\paragraph{Batch normalization.}

\Gls{BN}~\citep{ioffe2015batch} is another recently introduced technique that acts as a regularizer. \Gls{BN} addresses
the covariate shift problem, i.e., its goal is to assure that the distribution of layer inputs does not change during
training. This is achieved by performing a zero-mean unit variance normalization for each mini-batch. Specifically, for
a mini-batch containing $m$ examples $(\vect{x}^{(1)}, \ldots, \vect{x}^{(m)})$, where every example is a
$d$-dimensional vector $\vect{x}^{(i)} = (x_1^{(i)}, \ldots, x_d^{(i)})$, each dimension $k$ is normalized separately:
\begin{equation}
  \mu_k = \frac{1}{m} \sum_{i=1}^m x_k^{(i)}, \quad
  \sigma_k^2 = \frac{1}{m} \sum_{i=1}^m (x_k^{(i)} - \mu_k)^2, \quad
  \hat{x}_k^{(i)} = \frac{x_k^{(i)} - \mu_k}{\sqrt{\sigma_k^2 + \epsilon}}, \quad
  y_k^{(i)} = \gamma_k \hat{x}_k^{(i)} + \beta_k,
\end{equation}
where $\mu_k$ and $\sigma_k^2$ are the mean and variance along $k$-th mini-batch dimension, respectively. Parameters
$\gamma_k$ and $\beta_k$ are learned along with other network parameters and correspond to the scale and shift for the
normalized $\hat{x}_k$. To avoid dividing by zero, a small constant $\epsilon > 0$ is introduced. The \gls{BN}
transformation is differentiable, and therefore it is possible to backpropagate the gradients through the normalization
parameters. \Acrlong{BN} can significantly speed up the training and in some cases even replace
dropout~\citep{ioffe2015batch}.

\subsubsection{Weight initialization}

The question how the weights in a neural network should be initialized is not trivial and has prompted a vigorous
research during the recent years~\citep{hinton2006reducing,glorot2010understanding,martens2010deep,glorot2011deep,
sutskever2013importance,he2015delving}. The spectrum of techniques attempting to solve this problem is wide and ranges
from simply setting the weights to random numbers drawn from scaled distributions to more complex approaches, such as
pretraining with \glspl{DBN} or transfer learning. We elaborate on this topic in Section~\ref{sec:weight_init}.

\subsection{Deep belief network}

The renaissance of deep learning in the 2000s began with the discovery that greedy layer-wise pretraining can be used to
find a combination of initial parameters that make training deep networks possible. The first architecture that
succeeded in this task was the \acrlong{DBN}~\citep{hinton2006reducing}. \Acrlong{DBN} is composed of stacked
\glspl{RBM}. Its training consists of first performing layer-by-layer unsupervised pretraining and then finetuning the
network with error backpropagation.

\subsubsection{Restricted Boltzmann machine}

\Acrlong{RBM}~\citep{smolensky1986information} is a generative model that learns a probability distribution over a set
of observations. It is composed of two groups of units, visible and hidden, that are arranged in a bipartite graph. The
visible units correspond to features of the observations, and the hidden units represent latent factors that model the
dependencies between these features. Each visible unit is connected to every hidden unit with a symmetrical weighted
connection. In the simplest case, visible and hidden units are binary. That is: $v_i, h_j \in \left\{0, 1\right\}$, $i=1
\ldots n$, $j=1 \ldots m$, where $n$ is the number of visible units, and $m$ is the number of hidden units.
\Acrlongpl{RBM} work by simultaneously updating the states of all hidden units given the states of visible units and
vice versa. The updates for binary visible and hidden units are stochastic:
\begin{equation}
  p\left(v_i=1 | \vect{h}\right) = \left(1 + e^{-\left(\vect{h}\vect{W}_{i \cdot}^\mathrm{T} + a_i\right)}\right)^{-1},
  \quad
  p\left(h_j=1 | \vect{v}\right) = \left(1 + e^{-\left(\vect{v}\vect{W}_{\cdot j} + b_j\right)}\right)^{-1},
\end{equation}
where $\vect{a}$ is a vector of visible unit biases, $\vect{b}$ is a vector of hidden unit biases, and $\vect{W}_{i
\cdot}$, $\vect{W}_{\cdot j}$ are the i-th row and the j-th column of the weight matrix $\vect{W}$, respectively. Note,
however, that certain other activation functions can also be used with \glspl{RBM}, e.g., to model non-binary
vectors~\citep{hinton2012practical}. For example, to deal with real-valued input, binary visible units can be replaced
by linear units with independent Gaussian noise~\citep{freund1992unsupervised,welling2004exponential}.

\subsubsection{Contrastive divergence}

In the \gls{RBM} model the goal of training is to maximize the product of probabilities that the model assigns to
observations from a training set. To do this, \gls{RBM} training algorithms approximate the gradient of the
log-likelihood of training vectors with respect to the weights and biases. This gradient is then used inside a gradient
descent procedure to update the weights. One of the most commonly used gradient approximation methods is the \gls{CD}
algorithm~\citep{hinton2002training}. A training step in \gls{CD} begins with taking a sample of visible and hidden
units over the training data. The algorithm thus picks a random training example $\vect{v}^{\left(p\right)}$ and then
takes a sample $\vect{h}^{\left(p\right)}$ of hidden units according to the activation probabilities
$p\left(h^{\left(p\right)}_j=1 | \vect{v}^{\left(p\right)}\right)$. Next, \gls{CD} takes an approximate sample
$\left(\vect{v}^{\left(n\right)}, \vect{h}^{\left(n\right)}\right)$ from the \gls{RBM} model by performing alternating
Gibbs sampling of the visible and hidden units, starting the chain from the hidden configuration
$\vect{h}^{\left(p\right)}$. The gradient is then approximated as:
\begin{equation}
  \begin{aligned}
    \frac{\partial \log p\left(\vect{v}^{\left(p\right)}\right)}{\partial \vect{W}} &
      = {\vect{v}^{\left(p\right)}}^\mathrm{T}\vect{h}^{\left(p\right)}
      - {\vect{v}^{\left(n\right)}}^\mathrm{T}\vect{h}^{\left(n\right)} \\
    \frac{\partial \log p\left(\vect{v}^{\left(p\right)}\right)}{\partial \vect{a}} &
      = \vect{v}^{\left(p\right)} - \vect{v}^{\left(n\right)} \\
    \frac{\partial \log p\left(\vect{v}^{\left(p\right)}\right)}{\partial \vect{b}} &
      = \vect{h}^{\left(p\right)} - \vect{h}^{\left(n\right)}.
  \end{aligned}
\end{equation}
In its fastest variant \gls{CD} performs only one Gibbs step - the so-called \gls{CD}\textsubscript{1} algorithm.
\gls{CD}\textsubscript{1} was used by \citet{hinton2006reducing} to train \glspl{DBN}, i.e., stacked \glspl{RBM} where
the first \gls{RBM} models the observed data and each subsequent \gls{RBM} models outputs from the previous layer. This
procedure was used to obtain initial weights for deep autoencoders and deep \gls{MLP} networks. Networks initialized in
this manner were then fine-tuned with error backpropagation, ultimately achieving state-of-the-art performance on
several dimensionality reduction and classification tasks.

The pretraining procedure described in~\citep{hinton2006reducing} was further developed by \citet{nair2010rectified}
with the introduction of \glspl{NReLU}, i.e., units with an activation function given by:
\begin{equation}
  \mathrm{NReLU}\left(x\right) = \max\left\{0, x + \mathcal{N}\left(0, \left(1 + e^{-x}\right)^{-1}\right)\right\}.
\end{equation}
\Acrlongpl{NReLU} replace binary hidden units during layer-wise pretraining. Afterwards, when the network is fine-tuned
with error backpropagation, hidden layers employ a deterministic variant of the above activation function, i.e., the
standard rectified linear function.

\subsection{Autoencoder}

The autoencoder is an \gls{MLP} whose aim is to reconstruct at the output the same information that it was given as
input. Therefore, autoencoders must have the same number of input and output units. Autoencoder's middle layer -- called
the coding layer -- usually has the smallest number of units. This is the case in undercomplete autoencoders, whose task
is to construct a compact representation of the data, for example for dimensionality reduction, data embedding or
visualization.

The autoencoder is composed of two parts, the encoder and the decoder. For an observation $\vect{x}$ presented to the
network on the input, the encoder calculates its representation $\vect{c} = e(\vect{x})$, and the decoder calculates the
reconstruction $d(\vect{c})$. Autoencoders are trained to minimize a loss function, whose aim is to penalize the
dissimilarity between $\vect{x}$ and $d(e(\vect{x}))$. By modifying the cost function, the training may also make the
representation on the coding layer have various desired properties, such as sparsity~\citep{ranzato2007} or being stable
to small changes in the input~\citep{rifai2011contractive}. After training, the encoder part of the network can be used
to extract codes for new inputs.

Autoencoders have been developed and studied for a long time before the advent of deep
learning~\citep{yann1987modeles,bourlard1988auto}. Originally these models were, however, composed of only three layers:
the input layer, the coding layer and the output layer. The discovery of generative
pretraining~\citep{hinton2006reducing} opened a possibility of training much deeper models.

\subsection{Convolutional neural network}

A \gls{CNN}~\citep{lecun1990handwritten,lecun1998gradient} can be thought of as an \gls{MLP}-variant specialized for
processing spatially structured data. The most important example of such data, on which \glspl{CNN} perform
exceptionally well, is the image data. By knowing that the input data is organized in a grid-like structure, \glspl{CNN}
can greatly reduce the number of learnable parameters, and thus speed up the training. This is achieved mostly by
enforcing lower neuron connectivity, weights sharing and pooling.

Apart from the classical fully-connected layers, \glspl{CNN} make use of two specific types of layers: convolutional
layers and pooling layers. The purpose of a convolutional layer is to detect local features in the input volume. Neurons
in convolutional layers are grouped into feature maps. However, instead of being connected to all units from the
previous layer, each neuron is connected only to a small region of the input, called the receptive field. Weights of
these connections form a filter and are shared among units from the same feature map. During the forward pass, filters
from each feature map are convolved with the input to produce the activation maps. Convolutional layers are interleaved
with pooling layers. Their function is to reduce the computation burden for subsequent layers, by down-sampling the data
representation. Apart from controlling the overfitting, pooling layers additionally make the spatially reduced
representation invariant to translation of the input. The most commonly used pooling method is max
pooling~\citep{zhou1988computation}, which amounts to partitioning the feature map from the previous layer into
non-overlapping regions and returning maximum value for each region. Typically, a \gls{CNN} employs a few stacks of
convolutional layers with \gls{ReLU} activations, followed by a pooling layer. This pattern can repeat itself several
times before a transition to one or more fully-connected layers. Similarly to an \gls{MLP}, a \gls{CNN} can be trained
with \gls{SGD} and backpropagation, as all operations performed by its layers are differentiable.

\glspl{CNN} were developed and used long before the advent of deep learning. One of the first successful application of
\glspl{CNN} was the LeNet architecture employed for digit recognition~\citep{lecun1998gradient}. However, the popularity
of CNNs really skyrocketed after the ImageNet Large Scale Visual Recognition Challenge (ILSVRC) competition in 2012,
where deep \glspl{CNN} dominated its competitors on a large-scale image classification
task~\citep{krizhevsky2012imagenet}. Since then, deep convolutional networks are the model of choice for almost all
image recognition tasks. Currently, the state-of-the-art results are achieved with a \gls{CNN} variant called
\glspl{ResNet}~\citep{he2015deep}. These networks employ very deep architectures (e.g. 152-layer networks
from~\citep{he2015deep,huang2016deep}) and introduce, so called, skip connections, which can bypass several layers.

%% file: background_dr.tex
\section{Challenges in training neural networks on sparse, high-dimensional data}
\label{sec:dr}

In this section, we focus on the problem of training \glspl{DNN} on data that is unstructured, sparse and
high-dimensional. We show why these three data properties, when occurring simultaneously, can make the training
computationally challenging or, in some cases, even infeasible. We discuss how the above problem can be overcome by
employing dimensionality reduction of the original data prior to network training. We survey existing dimensionality
reduction approaches, focusing on methods that are capable of processing sparse, high-dimensional data.

% http://jmlr.org/papers/volume15/tan14a/tan14a.pdf
% With the rapid development of the Internet, big data of large volume and ultrahigh dimensionality have emerged in
% various machine learning applications, such as text mining and information retrieval (Deng et al., 2011; Li et al.,
% 2011, 2012). For instance, Weinberger et al. (2009) have studied a collaborative email-spam filtering task with 16
% trillion (1013) unique features. The ultrahigh dimensionality not only incurs unbearable memory c 2014 Mingkui Tan,
% Ivor W. Tsang and Li Wang. Tan, Tsang and Wang requirements and high computational cost in training, but also
% deteriorates the generalization ability because of the “curse of dimensionality” issue (Duda et al., 2000.; Guyon and
% Elisseeff, 2003; Zhang and Lee, 2006; Dasgupta et al., 2007; Blum et al., 2007). Fortunately, for many data sets with
% ultrahigh dimensions, most of the features are irrelevant to the output. Accordingly, dropping the irrelevant features
% and selecting the most relevant features can vastly improve the generalization performance (Ng, 1998). Moreover, in
% many applications such as bioinformatics (Guyon and Elisseeff, 2003), a small number of features (genes) are required
% to interpret the results for further biological analysis. Finally, for ultrahigh-dimensional problems, a sparse
% classifier is important for faster predictions.

\subsection{Sparse, high-dimensional, unstructured data}
\label{sec:sparse_high_unstuct_data}

We begin by describing the type of data we are interested in, along with the challenges it entails. We also look at
domains and applications where such data may arise.

\paragraph{Data dimensionality.}

A common and intuitive way to represent a given dataset is by using the vector-space
model~\citep{salton1979mathematics}. In the vector-space model, the observations are represented by an~$n \times d$
matrix called the design matrix, in which each of the~$n$ rows corresponds to an observation that is described by~$d$
attributes (also called features or variables). The interpretation of the attributes depends, of course, on the nature
of the dataset. For a set of images, an observation refers to an image that is defined by a list of pixel intensities or
higher-level features, whereas text, for example, is often represented as a multiset of its words -- the so-called
\gls{BOW} representation. Regardless of the feature interpretation, their number $d$, i.e., \textbf{the data
dimensionality}, plays an important role in determining the applicability of machine learning and data mining methods.

High-dimensionality is a ubiquitous property of modern real-world datasets. Data having hundreds or even millions of
features arise in various application domains, e.g., 2D/3D digital image processing, bioinformatics, e-commerce, web
crawling, social media, mass spectrometry, text analysis or speech processing.

\paragraph{Data sparsity.}

Sparsity is a common property of many high-dimensional datasets. It is defined as the number of zero-valued elements in
the~$n\times d$ design matrix divided by the total number of elements~$nd$. However, when working with highly sparse
datasets a more convenient term to use is \textbf{data density}, which is equal to one minus the
sparsity~\citep{herlocker2004evaluating}.

Data can be sparse for two main reasons. Zeros in the design matrix may simply represent missing measurements, also
denoted as null values or ``NA'' values. This is the case, for example, in recommender system data, where rows of the
design matrix correspond to users and columns correspond to items. Each row of such data matrix contains user's ratings
of $d$ items. As $d$ is often extremely large, even the most active users are only able to rate just a small subset of
all items. Therefore, most elements of the data matrix are missing and are often represented by zeros. The second reason
for data sparsity stems from the type of data representation. For example, sparsity can be introduced as a result of
binarization or discretization techniques. It may also result from converting categorical variables to a one-hot
representation or when converting text data to \gls{BOW} representation.

From the computational point of view processing sparse data has both advantages and disadvantages. On the one hand,
sparsity is beneficial as it enables storing and manipulating large data in a compressed format. On the other hand,
efficient manipulation of sparse matrices requires specialized algorithms and data structures that are capable of taking
advantage of the sparse representation. Moreover, sparse data often entails using careful normalization during
preprocessing.

\paragraph{Data structure.}

The third important property of datasets we are interested in is the lack of structure. The terms ``structured data''
and ``unstructured data'' are, however, imprecise and may have different meanings, depending on the area in which they
are used. Here, we consider data as unstructured if its structure is not helpful for our task, i.e., training a neural
network model. For example, images are a typical example of structured data, as their spatial structure, i.e., pixel
adjacency, can be exploited when designing the network architecture (as convolutional networks do). On the other hand,
text data in the \gls{BOW} representation is unstructured: the context and word order do not matter and there is no
apparent similarity between words, as they correspond to indices in a vocabulary.

\subsubsection{Where does such data arise?}

There are many kinds of data that exhibit the above properties. Millions or even billions of input features, high
sparsity and lack of structure can be found in applications such as natural language processing, malware detection,
recommendation and ranking systems, bioinformatics and high energy physics.

\paragraph{Text data.}

High dimensionality and sparsity of text data is usually the result of employing the \gls{BOW} model. In this model, the
text is represented as a multiset of its tokenized words. Therefore, a collection of text documents can be represented
in the vector-space model as an $n \times d$ data matrix, with $n$ equal to the number of documents and $d$ equal to the
size of the dictionary. The dictionary is the set of all unique words appearing at least once in the corpus. Since the
great majority of documents typically use a tiny subset of the dictionary, the data matrix is often very sparse.
Although being simplistic, i.e., not taking grammar or word order into account, the \gls{BOW} model is still a popular
representation of text data. One extension of the \gls{BOW} model capable of capturing phrases and multi-word
expressions is the bag of n-grams representation. Instead of building a multiset of single words, it counts the
occurrences of n-grams (usually bigrams or trigrams) of either words or characters. However, this greatly increases the
dictionary size, especially when constructing representations of large text corpora.

\paragraph{Biological data.}

Certain types of biological data are also high-dimensional, sparse and unstructured. One example is thresholded
microarray data. Microarray experiments measure expression levels of tens of thousands of genes (features)
simultaneously. While these measurements are initially stored as dense real-valued matrices, it is not uncommon to
preprocess them and only store the discretized values for genes that are significantly up- or down-regulated between
different experimental conditions. In this representation, the microarray data becomes highly sparse, as usually, just a
small fraction of all genes are up- or down-regulated.

Another example of sparse high-dimensional biological data is the \gls{SNP} data. A \gls{SNP} is a variation of a single
base-pair at a specific location in the DNA sequence among individuals of the same species. Dimensionality of \gls{SNP}
data can be very high, as the number of identified and mapped SNP sites often reaches hundreds of thousands or millions.
Raw \gls{SNP} data is mostly dense as the majority of SNPs occur at a frequency of more than 1\%. \Acrlongpl{SNP}
occurring less often are considered as low frequency variants or ``mutations''~\citep{barnes2002snp}. Therefore, studies
that focus on these low-frequency variants, such as cancer research, use sparse data (see,
e.g.,~\citep{vural2016classification}).

A significant challenge in training neural networks on biological data stems from the disproportion between the number
of available training examples and example dimensionality. In particular, for most biological datasets the number of
features is up to several orders of magnitude greater than the number of examples. This is the result of treating
samples or patients as observations (examples) and genes or proteins as variables. This perspective is common, e.g. in
the identification of significantly expressed genes, cancer classification and other
studies~\citep{clarke2008properties}.

\paragraph{Web data.}

The Internet is an important source of various types of large-scale data. In particular, huge amounts of data of
different nature can be extracted from web pages using web crawlers. Many archives of web crawl data are publicly
available for research and analysis. Such archives contain terabytes or even petabytes of raw web page data and metadata
collected over several years of web crawling\footnote{See, e.g., Common Crawl: \url{http://commoncrawl.org/}}. However,
because of storage and computational costs, these amounts of unprocessed data are usually unfit for research purposes.
Fortunately, many projects, such as, e.g., Web Data Commons\footnote{\url{http://webdatacommons.org/}}, provide datasets
extracted from raw web crawl data. Most often web crawl data is sparse and high-dimensional because of the use of the
\gls{BOW} model for text data or feature binarization and discretization techniques.

One interesting example of large-scale data closely related to web crawling is the \gls{URL} reputation data. Features
in \gls{URL} reputation datasets are a combination of lexical features, such as \gls{BOW} representation of tokens in
the \gls{URL}, with host-based features, such as DNS, WHOIS, AS and IP related information or blacklist
membership~\citep{ma2009identifying}. Because of large amount of examples and dimensionality reaching millions of
features, this type of data can be used for evaluating online learning methods.

Another example of sparse, high-dimensional web data is advertisement click-through rates prediction data. Here, the
main cause of high dimensionality and sparsity is the use of one-hot-encoding representation for categorical
features~\citep{richardson2007predicting,lee2012estimating}.

\paragraph{Other types of data.}

Other, more exotic machine learning data include, for example, logs of student interactions with intelligent tutoring
systems (KDD Cup 2010 data). The dimensionality of such datasets after preprocessing can reach tens of millions of
features~\citep{yu2010feature}. Another example of sparse, high-dimensional data is link data, i.e., graphs represented
by adjacency matrices, where each feature corresponds to a weight or absence/presence of a link between nodes in a large
network.

\subsection{Learning from sparse, high-dimensional data}

In many cases training neural networks on data characterized above can be computationally challenging or even
infeasible. Note that this does not necessarily mean that neural networks cannot be trained on very-high-dimensional
data. Often, when the data is structured the number of learnable parameters, i.e., weights, can be greatly reduced.
\glspl{CNN}, for example, reduce the number of parameters by exploiting local connectivity and sharing the weights
between spatial locations. However, when the input data has no obvious structure it is difficult to constrain the
network architecture. In such scenario, learning directly from unprocessed data ties the number of input units in the
first layer to the data dimensionality. As a result, when the number of input dimensions exceeds tens of thousands, the
number of weights in the first fully-connected layer is so large that the training becomes practically infeasible.

Fortunately, the above problem can be overcome by first \textbf{reducing the
dimensionality}~\citep{van2009dimensionality}\footnote{The purpose of dimensionality reduction is to create a meaningful
lower-dimensional representation of the original data. Dimensionality reduction limits the influence of the so-called,
curse of dimensionality, which greatly facilitates or even enables, e.g., classification, clustering, visualization or
compression of high-dimensional datasets. The term ``curse of dimensionality'', originally coined
in~\citep{bellman1961adaptive}, broadly relates to problems that arise when the dimensionality of the analyzed data
becomes high. In the context of machine learning, it refers to the difficulty of searching high-dimensional spaces and
finding structure in data embedded in such spaces~\citep{duda2012pattern,donoho2000high}.} of the input space to a
manageable size and then training a network on a representation with fewer dimensions. While in recent years a plethora
of dimensionality reduction methods have been developed~\citep{jovic2015review}, few of them are suited for handling
sparse, high-dimensional data. We focus on several such methods that can be applied as a preprocessing step before
network training.

Note that even with the performance considerations put aside, performing dimensionality reduction prior to network
training has its merits. In principle, the transformation realized by many dimensionality reduction techniques can as
well be performed by one or more layers of the network (see for example \gls{PCA} realized by a neural network
layer~\citep{oja1982simplified} or autoencoders~\citep{hinton2006reducing}). However, this approach results in larger
network architectures that require more data to be trained and are more prone to overfitting. Therefore, we focus on
performing the dimensionality reduction procedure separately, before the network training.

Dimensionality reduction methods can be divided into two categories: feature selection and feature extraction. The aim
of feature selection is to limit the number of features by only keeping the most relevant ones and discarding the
others. Feature extraction, on the other hand, constructs new derived features by transforming the original input
variables. This transformation can be either linear or nonlinear. Feature selection is also employed in problems where
the aim is not the dimensionality reduction itself but rather the identification of influential feature subsets: e.g.,
in bioinformatics for finding genes related to resistance to a pathogen~\citep{guyon2002gene}.

\paragraph{Feature selection methods.}

Feature selection is based on a premise that some features might be unnecessary by being either redundant or irrelevant,
and thus can be removed from the feature set. Which features are considered relevant depends on the type of the feature
selection algorithm. Traditionally, three approaches to feature selection were proposed: filter methods, wrapper methods
and embedded methods~\citep{guyon2003introduction}.

Filter methods try to assess feature relevancy only from the data, without evaluating the influence of the selected
feature subset on the performance of the trained model. This approach is computationally much less expensive than the
wrapper approach but produces feature subsets that are not tailored to a specific model. Most filter methods are
univariate, i.e., they rank variables according to their individual predictive power, which was shown to yield inferior
models compared to methods that rank subsets of features~\citep{guyon2003introduction}. In order to alleviate this
problem a number of multivariate filter methods have also been proposed~\citep{saeys2007areview}.

In the wrapper approach selection of relevant features is performed with regard to the model performance. Wrapper
methods view the feature selection process as a search problem, where the search space is defined by all possible
feature subsets. Different feature combinations from the feature set $\mathbb{F}$ are assigned scores that are based on
the performance of models trained on these combinations. For practical applications, where
$\left\vert{\mathbb{F}}\right\vert$ is large, evaluating all possible feature subsets is infeasible, as the number of
subsets grows exponentially with $\left\vert{\mathbb{F}}\right\vert$. When an exhaustive search is impossible, a wide
range of search strategies can be applied, including best-first, genetic algorithms, simulated annealing, particle swarm
optimization or branch-and-bound~\citep{kohavi1997wrappers}. These methods may still be computationally prohibitive when
training a single model on a particular feature subset is costly. In this case, greedy search strategies including,
e.g., forward selection or backward elimination, may prove useful. Apart from being computationally expensive, these
methods are also prone to overfitting, especially for large $\left\vert{\mathbb{F}}\right\vert$.

Similarly to wrapper methods, embedded methods~\citep{guyon2003introduction} rely on evaluating a classifier (or another
model) on candidate feature subsets. However, they incorporate feature selection into the training method itself, i.e.,
they learn which features contribute to the model's performance while the model is being created. In this way, they
avoid expensive retraining of the whole model after every modification in the feature subset.

\paragraph{Feature extraction methods.}

Unlike feature selection methods, feature extraction approaches construct new features by performing a transformation of
the original high-dimensional data into a lower-dimensional space~\citep{van2009dimensionality}. Based on the type of
this transformation, feature extraction methods can be classified into linear and nonlinear methods. Linear feature
extraction methods include, among others, \gls{PCA}~\citep{pearson1901liii,jolliffe2002principal}, \acrlong{RP},
\gls{LDA}~\citep{friedman2001elements}, \gls{MDS}~\citep{torgerson1952multidimensional} and
\acrlong{MMC}~\citep{li2006efficient}. Nonlinear methods include, for example, ISOMAP~\citep{tenenbaum2000global},
locally linear embedding~\citep{roweis2000nonlinear}, autoencoders~\citep{hinton2006reducing}, Sammon
mapping~\citep{sammon1969nonlinear} or t-SNE~\citep{maaten2008visualizing}. In general, linear dimensionality reduction
techniques are computationally more efficient than nonlinear methods, but often perform worse on complex, real-world
data. For a comparative overview of popular feature extraction methods
see~\citep{van2009dimensionality,storcheus2015survey}.

% Feature extraction works because most datasets are embedded in a high-dimensional space but their data points lay on a
% manifold with much fewer dimensions. In the ideal case, the number of dimensions used to represent the reduced data
% should correspond to the intrinsic dimensionality of the data~\citep{bennett1965representation}. The intrinsic
% dimensionality relates to the minimum number of variables that are necessary to explain the variability of the
% observed data without information loss.

\subsection{Feature selection for sparse, high-dimensional data}
\label{sec:background_fs_sparse_high}

Many feature selection methods have been proposed throughout the years (for a comparative study
see~\citep{jovic2015review,kumar2014feature}). However, few of these methods are applicable to sparse, high-dimensional
data. In particular, many state-of-the-art feature selection methods require calculating pairwise correlation
coefficients between the features. This makes them infeasible for datasets with millions of features. Similarly, more
complex wrapper methods, which require training the model multiple times, are not applicable in these settings due to
their computational cost. Therefore, reducing data dimensionality using feature selection methods is viable mostly with
the filter methods. Unfortunately, existing filter methods that are suitable for dense data cannot be easily modified to
be applicable to sparse datasets~\citep{liu2005toward}.

In an influential study, \citet{forman2003extensive} evaluated several feature selection methods including Chi-square,
\gls{IG}, F1-measure, odds ration, bi-normal separation and others. However, \citeauthor{forman2003extensive} focused
specifically on the problem of feature selection for the purpose of classification of \gls{BOW}-represented text data.
In a more recent study of feature selection methods for general big data, \citet{bolon2015recent} enumerate several
popular algorithms suitable for processing high-dimensional datasets. These include mostly filter approaches and
scalable embedded methods, i.e., Chi-square, F-score, \gls{IG}, ReliefF, mRMR, \mbox{SVM-RFE}, CFS, FCBF,
\mbox{INTERACT} and Consistency. Here, we are interested in processing datasets with the number of examples $n$ and the
number of features $d$ both on the order $\ge 10^{5}$. Therefore, we only focus on the fastest feature selection
methods, whose computational complexity is not worse than~$\mathcal{O}(nd)$. These are: Chi-square~\citep{liu1995chi2},
F-score~\citep{duda2012pattern} and \gls{IG}~\citep{quinlan1986induction}. All these methods are univariate, i.e., each
of them scores the features independently.

\subsubsection{Chi-square}
\label{sec:chi2}

The Chi-square feature selection ranks features in a dataset represented by a design matrix $\mathbf{X}$ by performing
$\chi^2$ tests between the feature vectors and the class vector $\mathbf{y}$. The method is applicable to continuous
data after binning~\citep{fayyad1993discret}, as the $\chi^2$ test is only defined for categorical (nominal) data. The
Chi-square statistic for a feature $i$ is calculated as:
\begin{equation}
    \chi^2(i; \mathbf{X}, \mathbf{y}) = \sum_{j \in \mathrm{vals}(\mathbf{X}_{\cdot i})} \sum_{k=1}^{N} \frac{(O_{ijk} -
    E_{ijk})^2}{E_{ijk}},
\end{equation}
where $\mathrm{vals}(\mathbf{v})$ is a function returning a set of unique values in $\mathbf{v}$, and $N$ is the number
of classes. $O_{ijk}$ and $E_{ijk}$ are the observed and expected numbers of examples belonging to class $k$ whose
$i$-th feature has value $j$. The value of $O_{ijk}$ is calculated from the empirical data, and $E_{ijk}$ is estimated
assuming the independence of feature $i$ from the predicted class. High scores of $\chi^2(i; \mathbf{X}, \mathbf{y})$
indicate that the null hypothesis of independence should be rejected and feature $i$ and predicted class are correlated.
Chi-square feature selection returns the highest ranked features, which are likely to be relevant during classification.

Importantly, Chi-square feature selection can be implemented in a way that leverages data sparsity, making it
computationally efficient.

\subsubsection{F-score}

The Fisher score of the $i$-th feature can be defined as:
\begin{equation}
  F_\mathrm{score}(i; \mathbf{X}, \mathbf{y}) = \frac{\sum_{k=1}^N n_{k}(\mu_k^i - \mu^i)^2}{\sum_{k=1}^N
  n_{k}(\sigma_k^i)^2},
\end{equation}
where $N$ is the number of classes, $n_{k}$ is the number of examples in $\mathbf{X}$ belonging to the $k$-th class,
$\mu_k^i$ and $\sigma_k^i$ are the average and standard deviation of feature $i$ for examples belonging to class $k$,
respectively, and $\mu^i$ is the average of feature $i$ over all examples. Feature selection with Fisher score finds a
set of features that are most discriminative between the classes, i.e., have the highest $F_\mathrm{score}$ values.
Specifically, it seeks for features that maximize the distances between the means of the classes while minimizing the
variance withing each class. This criterion is also used in feature extraction, e.g., in \acrlong{LDA}. Despite being
simple, the F-score feature selection combined with random forest and \gls{SVM} has been shown to work surprisingly
well~\citep{chen2006combining}.

Several more complex feature selection methods have been developed based on the Fisher score. For example,
\citet{gu2012generalized} proposed a generalized multivariate F-score method, i.e., a method that selects a subset of
features simultaneously. However, its computational cost makes it prohibitive in our settings. Although being suitable
for sparse data, the method assumes that the data matrix has been centered. Unfortunately, centering each feature cannot
be realized without making the data matrix dense.

\subsubsection{Information gain}

Information gain for a feature $i$ is the amount of uncertainty about the predicted class that gets reduced when feature
$i$ is observed. Here we use the term \textit{information gain} as it was introduced in the context of decision
trees~\citep{quinlan1986induction}. Therefore, it is equivalent to \textit{mutual information}. Alternatively, some
authors define information gain as the Kullback–Leibler divergence (also known as information divergence or relative
entropy).

More formally, given a set of training examples $\mathbb{X}$, each of the form \mbox{$(\mathbf{x}, y) = (x_1, \ldots,
x_k, y)$}, information gain $\mathrm{IG}(i; \mathbb{X})$ is the reduction of entropy that is achieved by observing
feature $i$:
\begin{align*}
  \mathrm{IG}(i; \mathbb{X}) &= H(\mathbb{X}) - H(\mathbb{X}|i) \\
  &= H(\mathbb{X}) - \sum_{v \in \mathrm{vals}(\mathbb{X}_i)} \frac{|\{(\mathbf{x}, y) \in \mathbb{X} | x_i = v
  \}|}{|\mathbb{X}|} H(\{(\mathbf{x}, y) \in \mathbb{X} | x_i = v\}),
\end{align*}
where $\mathbb{X}_i$ is a feature vector in the dataset that corresponds to feature $i$, and $\mathrm{vals}$ is defined
as in the Chi-square method. Entropy $H$ for a dataset $\mathbb{S}$ is defined as:
\begin{equation}
  H(\mathbb{S}) = - \sum_{c = 1}^N p_c(\mathbb{S})\log{p_c(\mathbb{S})},
\end{equation}
where $N$ is the number of classes in $\mathbb{S}$ and $p_c(\mathbb{S})$ is the probability of a training example in
$\mathbb{S}$ belonging to the class $c$. $\mathrm{IG}(i; \mathbb{X})$ is equal to zero if variable represented with
feature vector $\mathbb{X}_i$ is independent from the predicted class vector. Similarly to F-score and Chi-square
methods, information gain selects features with highest scores $\mathrm{IG}(i; \mathbb{X})$, which suggest their high
correlation with the predicted class.

Note that the above definition of information gain is suitable for datasets with a discrete set of feature values. For
continuous data, several methods of discretization were developed, most notably the information theoretic
binning~\citep{fayyad1993discret}. An alternative approach is to estimate the entropy with $k$-nearest neighbor
distances~\citep{kraskov2004estimating}.

% https://arxiv.org/pdf/1409.7794.pdf
% http://www.jsoftware.us/vol11/132-JSW1542.pdf
% http://www.shivani-agarwal.net/Teaching/E0371/Papers/icml10-sparse-svm.pdf

% https://www.elen.ucl.ac.be/Proceedings/esann/esannpdf/es2011-91.pdf
% filter method
% only for BOW data

\subsection{Feature extraction for sparse, high-dimensional data}
\label{sec:background_fe_sparse_high}

In general, most feature extraction methods are computationally more demanding than filter feature selection approaches.
As performance is a key issue in our application, we focus on the most efficient linear feature extraction algorithms.

In a recent comprehensive study of commonly used linear feature extraction techniques \citet{cunningham2015linear}
discuss \gls{PCA}, \gls{MDS}, \gls{LDA}, \gls{CCA}, \gls{MAF}, \gls{SFA}, \gls{SDR}, \gls{LPP}, \gls{ICA}, probabilistic
\gls{PCA}, factor analysis and distance metric learning. However, in their analysis \citeauthor{cunningham2015linear}
focus on reducing the dimensionality of dense data. In particular, they assume that the original input data can be
easily mean-centered. This step cannot be realized for large sparse datasets, without making them fully-dense and
destroying the benefits of sparse representation. Moreover, most of the methods discussed
in~\citep{cunningham2015linear} were not developed for sparse data, and thus are unfit for such applications. These
include \gls{MDS}, \gls{LDA}, \gls{SFA}, \gls{SDR}, \gls{LPP} and \gls{ICA}. \citet{james2001functional} presented a
modified version of \gls{LDA}, so called functional \gls{LDA} and suggested that it can be extended to be applicable to
sparse data. However, they did not evaluate this modification on sparse datasets and did not specify its computational
complexity. \Gls{CCA} and \gls{MAF} both require performing eigendecomposition of the correlation or covariance matrix,
which makes their computational complexity too high for our case. \Gls{SFA} also requires expensive estimation of the
covariance matrix. To solve this problem, \citet{kompella2012incremental} propose an online version of \gls{SFA}, called
incremental \gls{SFA}, which does not rely on computing a covariance matrix. However, similarly to previous methods,
their approach is also not suited for sparse data. \Gls{MDS}, \gls{LPP}, \gls{ICA} and its numerous extensions also have
prohibitive computational complexity -- most often not lower than $\mathcal{O}(n^3)$ (assuming for simplicity that $n
\approx d$)~\citep{van2009dimensionality,he2003locality}. Distance metric learning methods like, e.g., neighbourhood
components analysis are more suited for visualization purposes, as they learn low-dimensional
embeddings~\citep{goldberger2005nca}.

Several scalable incremental feature extraction algorithms have also been proposed. These include
\gls{IMMC}~\citep{yan2004immc}, online variants of \gls{LDA}, \gls{IPCA}~\citep{li2003integrated} and
\gls{CCIPCA}~\citep{weng2003candid}. These methods were developed in the context of online learning for problems
associated with data streaming. However, their computational complexity is sometimes still too high for our purpose,
i.e., for cases when both $n$ and $d$ are on the order of millions. For example, \gls{IMMC} improves the complexity of
batch \gls{MMC}~\citep{li2006efficient} from $\mathcal{O}(min\{n^3, d^3\})$ to $\mathcal{O}(ndkc)$, where $c$ is the
number of classes. While much faster than the classical method, this is still significantly slower than, e.g., \gls{PCA}
realized via a randomized version of the block Lanczos method~\citep{halko2011algorithm} (see the section about
\gls{PCA} below). Several variants of \gls{LDA} offer faster computational time. \Gls{IDR/QR}~\citep{ye2005idr}, for
example, offers complexity of~$\mathcal{O}(ndc)$. This is achieved by applying QR decomposition instead of \gls{SVD}.
Unfortunately, \gls{IDR/QR} is not suited for sparse data. Another example of a fast \gls{LDA}-based method is
\gls{ILDA}~\citep{kim2007incremental}, which can be computed in~$\mathcal{O}(dk^2)$ -- time that is not dependent on the
number of training examples. Similarly to \gls{ILDA}, \gls{SRDA}~\citep{cai2008srda} is capable of processing sparse
data and can be computed in just $\mathcal{O}(min\{n,d\}s)$ operations, where $s$ is the average number of non-zero
features in each example. However, the application of \gls{LDA}-based methods is limited due to the so-called
\textit{singularity problem}~\citep{krzanowski1995discriminant}, which occurs when the data dimensionality exceeds the
number of examples. Several variants of \gls{PCA} that construct an incremental representation of the covariance matrix
have been proposed, e.g., \gls{IPCA} and \gls{CCIPCA}. However, these methods are also unable to efficiently process
sparse datasets.

% https://www.cse.ust.hk/~qyang/Docs/2006/jun2.pdf

\subsubsection{Principal component analysis}

\Acrlong{PCA}, one of the most widely used tools in data analysis and data mining, is also one of the most popular
linear dimensionality reduction methods. It attempts to find a feature subspace that preserves the most of the data
variability. The basic approach to computing \gls{PCA} of matrix \mbox{$\mathbf{X} \in \mathbb{R}^{n
\times d}$} involves calculating the covariance matrix $\frac{1}{n-1}\mathbf{X}^{T}\mathbf{X}$ and performing its
eigendecomposition. Then, $k$ principal components with the highest eigenvalues are used to project the data into a
lower-dimensional space. While efficient for datasets with $d<n$, this approach can be numerically inaccurate, as the
condition number of the covariance matrix is the square of the condition number of $\mathbf{X}$ (see, e.g., the
L\"{a}uchli matrix~\citep{lauchli1961jordan}). Instead, \gls{PCA} is often realized by performing \gls{SVD} of the
normalized data matrix $\mathbf{X}$, which can be computed in $\mathcal{O}(min\{nd^2,n^2d\})$. For dense datasets with
$n \sim d$, this makes it prohibitive for values of $n$ higher than several thousand. In practice, however, it is
usually sufficient to compute a reduced version of \gls{SVD}, i.e., a truncated \gls{SVD}, to determine only the $k$
largest singular values of $\mathbf{X}$~\citep{friedman2001elements}. This can be achieved by using, e.g., iterative
Lanczos' methods and can speed up the computation to $\mathcal{O}(ndk)$ while also reducing the memory footprint of the
algorithm. However, even calculating a partial \gls{SVD} is computationally prohibitive when $k$ is large and $n$ and
$d$ are on the order of millions. A solution to this challenge arises from randomized matrix algorithms, which can
reduce the computational complexity even further: from $\mathcal{O}(ndk)$ to $\mathcal{O}(nd\log
k)$~\citep{mahoney2011randomized}. This can yield a significant speedup when we are interested in reducing the
dimensionality of the data to $k$ that is on the order of thousands. Such efficient algorithms for large-scale \gls{PCA}
have been presented in e.g.,~\citep{halko2011algorithm,rokhlin2009randomized,georgiev2012randomized}.

\subsubsection{Random projection}

\Acrlong{RP} is a simple and computationally efficient linear dimensionality reduction technique. We present this method
in detail and focus on its properties and applications in Chapter~\ref{cha:rp}.

% Jiquan Ngiam, Aditya Khosla, Mingyu Kim, Juhan Nam, Honglak Lee, and Andrew Y Ng. Multimodal deep learning. In
% Proceedings of the 28th international conference on machine learning (ICML-11), pages 689–696, 2011.
% Yi Sun, Xiaogang Wang, and Xiaoou Tang. Deep learning face representation from predicting 10,000 classes. In 
% Proceedings of the IEEE Conference on Computer Vision and Pattern Recognition, pages 1891–1898, 2014.
% http://www.journal.au.edu/ijcim/2008/may2008/P1-IJCIM16n2-01.pdf
% http://hrcak.srce.hr/file/143365

%% file: rp.tex
\cleardoublepage
\chapter{Random projection}
\label{cha:rp}

\Acrlong{RP} is a computationally efficient and conceptually simple dimensionality reduction technique. The key idea
behind \gls{RP} stems from the Johnson-Lindenstrauss lemma, which states that a set of points in a high-dimensional
space can be embedded into a lower-dimensional space, with distances between these points preserved up to a certain
multiplicative factor. Surprisingly, the dimensionality of this lower-dimensional space is logarithmic in $n$ and does
not depend on the dimensionality of the original data. In other words, \gls{RP} makes it possible to compactly
approximate a dataset consisting of $n$ examples using just $\mathcal{O}(n\log{n})$ memory. This is a big advantage,
especially when processing large-scale datasets whose dimensionality is on the order of, or even exceeds the number of
examples. Most importantly, by reducing the number of features to $\mathcal{O}(\log{n})$, \gls{RP} can make many methods
that strongly depend on data dimensionality viable. In the next chapter, for example, we report experiments in which we
used \gls{RP} to train neural networks on data whose dimensionality would otherwise be prohibitively high for such
models.

The Johnson-Lindenstrauss lemma is at the core of many algorithms in signal processing, statistics and computer science.
One notable example that greatly popularized \gls{RP} is sparse signal reconstruction, also known as compressed
sensing~\citep{donoho2006compressed}. \Acrlong{RP} has also found use in various machine learning tasks, e.g.,
classification~\citep{goel2005face,arriaga2006algorithmic,rahimi2008random,paul2014random},
regression~\citep{maillard2012linear,kaban2014new} or clustering~\citep{fern2003random,boutsidis2010random}. For an
overview of applications of \gls{RP} see~\citep{indyk1998approximate,vempala2005random}.

This chapter is organized as follows. In Section~\ref{sec:jl_lemma_ose} we introduce the Johnson-Lindenstrauss lemma and
the notion of \acrlongpl{OSE}. Next, in Section~\ref{sec:rp_schemes} we present five important \gls{RP} constructions:
Gaussian, Achlioptas’, Li’s, \acrlong{SRHT} and Count Sketch. We analyze their properties, focusing on the embedding
quality, applicability to sparse data and computational cost of performing the projection.

\section{Johnson-Lindenstrauss lemma and embedding quality}
\label{sec:jl_lemma_ose}

The most important theoretical result behind \gls{RP} is the Johnson-Lindenstrauss lemma
from~\citep{johnson1984extensions}. Formally it is the following fact:
\begin{lemma}[JL-lemma~\citep{johnson1984extensions}]\label{lemma:jl}
Let $\epsilon \in (0,1)$ and $\mathbb{A}$ be a set of $n$ points in $\mathbb{R}^d$. Let $k$ be an integer and
\mbox{$k = \mathcal{O}(\epsilon^{-2}\log{n})$}. Then there exists a mapping
$f: \mathbb{R}^d \mapsto \mathbb{R}^k$ such that for any $\mathbf{a},\mathbf{b} \in \mathbb{A}$
\begin{equation}
  (1-\epsilon)\lVert \mathbf{a}-\mathbf{b} \rVert_2  \le \lVert f(\mathbf{a})-f(\mathbf{b}) \rVert_2 \le
      (1+\epsilon)\lVert \mathbf{a}-\mathbf{b} \rVert_2.
\end{equation}
\end{lemma}
That is, every dataset with $n$ examples, regardless of its dimensionality, can be represented in \mbox{$k =
\mathcal{O}(\epsilon^{-2}\log{n})$} dimensions in a way that preserves the pairwise distances between any two examples
up to a multiplicative factor $1 \pm \epsilon$, where $\epsilon$ is the distortion. This estimation is optimal both in
$n$ and $\epsilon$, i.e., without a priori knowledge of the dataset, no linear dimensionality reduction technique can
improve the JL-lemma guarantee on $k$~\citep{alon2003problems}.

Note, however, that Lemma~\ref{lemma:jl} is not constructive, i.e., it does not specify how to create the mapping $f$.
In their proof, Johnson and Lindenstrauss chose $f$ as an orthogonal transformation whose corresponding projection
matrix is neither easy nor efficient to generate for practical applications. One approach is to initialize the
projection matrix with random numbers drawn from a normal distribution and then apply an orthogonalization procedure,
such as the Gram-Schmidt method, which runs in $\mathcal{O}(dk^2)$~\citep{golub2012matrix}. In recent years multiple
more practical constructions that satisfy the JL-lemma have been proposed. Such mappings $f$ that preserve pairwise
distances between the observations are called \acrlong{RP} schemes or \glspl{JLT}. In Section~\ref{sec:rp_schemes}, we
present several important \glspl{JLT}, from historically earliest to more recent. These schemes differ in two main
aspects:
\begin{itemize}
  \item the computational complexity of constructing the projection matrix (if it is explicitly needed) and projecting
  the data,
  \item the quality of embedding they provide.
\end{itemize}
To assess and compare the embedding quality of different \glspl{JLT}, or more specifically, the distributions according
to which their projection matrices are generated, we use two important concepts, first introduced
in~\citep{sarlos2006improved}: the subspace embedding property and the \gls{OSE} property. They are the main tools for
analyzing recent \gls{RP} schemes~\citep{clarkson2013low,nelson2013osnap,woodruff2014sketching}.
\begin{definition}[$(1 \pm \epsilon)$ $\ell_2$-subspace embedding~\citep{woodruff2014sketching}]
Let $\mathbf{A}$ denote an $n \times d$ matrix and $\mathbf{S}$ denote a $d \times k$ matrix, where $k \ll d$.
$\mathbf{S}$ is a $(1 \pm \epsilon)$ $\ell_2$-subspace embedding for $\mathbf{A}$ if $\forall{\mathbf{x} \in
\mathbb{R}^n}$
\begin{equation}
  (1-\epsilon) \lVert \mathbf{x}^T\mathbf{A} \rVert_2^2
      \le
  \lVert \mathbf{x}^T\mathbf{AS} \rVert_2^2
      \le
  (1+\epsilon) \lVert \mathbf{x}^T\mathbf{A} \rVert_2^2.
\end{equation}
\end{definition}
That is, matrix $\mathbf{S}$ is a subspace embedding for $\mathbf{A}$ if for any given vector $\mathbf{x} \in
\mathbb{R}^n$ the length of vector $\mathbf{x}^T\mathbf{A}$ is similar to the length of its sketch
$\mathbf{x}^T\mathbf{AS}$. One particularly useful variant of subspace embeddings is the \acrlong{OSE}.
\begin{definition}[Oblivious subspace embedding~\citep{woodruff2014sketching}]
Let $\mathbf{A}$ denote an $n \times d$ matrix and $\mathbf{\Pi}$ denote a distribution on $d \times k$ matrices
$\mathbf{S}$, where $k$ is a function of $n$, $d$, $\epsilon$ and $\delta$. $\mathbf{\Pi}$ is an $(\epsilon, \delta)$
oblivious $\ell_2$-subspace embedding if with probability at least $1 - \delta$ matrix $\mathbf{S}$ drawn from
distribution $\mathbf{\Pi}$ is a $(1 \pm \epsilon)$ $\ell_2$-subspace embedding for $\mathbf{A}$.
\end{definition}
That is, distribution $\mathbf{\Pi}$ is an \gls{OSE} if a random matrix drawn according to $\mathbf{\Pi}$ is a subspace
embedding with high probability. \citet{nelson2014lower} proved that the optimal lower bound for $k$, in order for a
distribution to be an \gls{OSE} is $\mathcal{O}(\epsilon^{-2}n)$. Note that this does not mean that the lowest
dimensional subspace into which \gls{RP} can embed an $n$-example dataset is $\mathcal{O}(\epsilon^{-2}n)$. The
\gls{OSE}'s lower bound for $k$ is not a contradiction to the JL-lemma, because the subspace embedding property is not
equivalent with the preservation of pairwise distances between examples. However, estimating \gls{OSE}'s lower bounds
for $k$ can be useful for comparing the embedding quality of different \gls{RP} constructions.

\section{Construction of the projection matrix}
\label{sec:rp_schemes}

Let~$\mathbf{A} \in \mathbb{R}^{n \times d}$ denote a data matrix consisting of~$n$ observations in~$\mathbb{R}^d$. In
\gls{RP}, matrix~$\mathbf{A}$ is projected from a high-dimensional space~$\mathbb{R}^d$ into a lower-dimensional
space~$\mathbb{R}^k$ ($k \ll d$) using a random matrix~$\mathbf{P} \in \mathbb{R}^{d \times k}$:
\begin{equation}\label{eq:rp}
  \tilde{\mathbf{A}} = \mathbf{A}\mathbf{P}.
\end{equation}
Of course, the projection matrix $\mathbf{P}$ cannot be completely random. Recently, many constructions of $\mathbf{P}$
that combine efficient projection with good embedding quality have been proposed. We can distinguish two main lines of
research here: one focusing on fast embedding of potentially dense
data~\citep{ailon2006approximate,ailon2009fast,dasgupta2010sparse} and one aiming at embedding data that is highly
sparse~\citep{dasgupta2010sparse,clarkson2013low,nelson2013osnap,kane2014sparser}. Below we present several important
\gls{RP} schemes from both of these groups, i.e., Gaussian, Achlioptas', Li's, \acrlong{SRHT} and Count Sketch-based
projections.

\subsection{Gaussian random matrix}

The original proof by Johnson and Lindenstrauss~\citep{johnson1984extensions} used a matrix composed of properly scaled
dense orthogonal vectors. However, for practical applications generating a large matrix with dense orthogonal rows or
columns is computationally too expensive. Luckily, as shown in~\citep{indyk1998approximate,dasgupta2003elementary} the
orthogonality constraint can be dropped. This observation led to a simple \gls{RP} matrix construction, i.e., a Gaussian
random matrix, whose entries are i.i.d. samples drawn from~$\mathcal{N}(0, \frac{1}{k})$. A justification for choosing
random vectors was provided in~\citep{hecht1994context}: the probability of random vectors being orthogonal or almost
orthogonal grows quickly with the vector dimensionality.

The main disadvantage of the Gaussian projection, when compared to more recent \gls{RP} constructions, is its
computational cost -- Gaussian projection matrix can be generated in $\mathcal{O}(dk)$ and requires $\mathcal{O}(ndk)$
operations to project a dense dataset. For sparse data, the projection time can be slightly improved (see
Table~\ref{tab:rp_complexity}). Despite being computationally demanding, the Gaussian projection scheme has two
advantageous traits: its implementation is straightforward and, more importantly, it produces a high-quality sketch of
the original data matrix. In fact, the Gaussian scheme achieves the optimal \gls{OSE} lower bound for the projected
dimensionality: $k = \mathcal{O}(\epsilon^{-2}n)$~\citep{nelson2014lower}.

\subsection{Achlioptas' random matrix}

One of the historically earliest lines of research on \gls{RP} matrices focused on improving the computational time of
performing the projection. This was achieved mostly by sparsifying the projection matrix. The first construction was
given in the seminal work by \citet{achlioptas2001database}, where he proposed two versions of the projection matrix
with simple probability distributions for the matrix elements:

\begin{equation}\label{eq:rp_achl1}
P_{ij} = \sqrt{\frac{1}{k}} \cdot
  \left\{
  \begin{array}{lcl}
    1 & \text{with probability } & 1/2 \\
    -1 & \text{with probability } & 1/2
  \end{array}
  \right.,
\end{equation}
\begin{equation}\label{eq:rp_achl3}
P_{ij} = \sqrt{\frac{3}{k}} \cdot
  \left\{
  \begin{array}{lcl}
    1 & \text{with probability } & 1/6 \\
    0  & \text{with probability } & 2/3 \\
    -1 & \text{with probability } & 1/6
  \end{array}
  \right..
\end{equation}
Achlioptas argued that these matrices yield a quality of embedding comparable to the quality provided by the random
Gaussian matrix. Practical applicability of Achliptas' construction was later confirmed by experimental
results~\citep{bingham2001random,fradkin2003experiments}.

Similarly to the Gaussian \gls{RP}, Achlioptas' distributions from Eq.~\ref{eq:rp_achl1} and Eq.~\ref{eq:rp_achl3} are
easy to implement. Additionally, their sparsity can be leveraged to compute the projection faster. Furthermore, the form
of Eq.~\ref{eq:rp_achl1} and Eq.~\ref{eq:rp_achl3} enables optimizations using integer arithmetics: scaling by the
constant $\sqrt{\frac{1}{k}}$ can be performed after the matrix multiplication (this can also be done for the Li's
construction presented in the next section). These properties make Achlioptas' \gls{RP} scheme especially useful in
database applications, which was in fact his main motivation. In this work, we use only the sparser construction
(Eq.~\ref{eq:rp_achl3}).

\subsection{Li's sparse matrix}
\label{sec:rp_sparse}

Achlioptas' work was continued by \citet{li2006very}, who introduced a sparser random matrix (in the literature often
referred to as the \textit{very sparse \acrlong{RP}}), by extending Achlioptas' constructions:
\begin{equation}
P_{ij} = \sqrt{\frac{s}{k}} \cdot
  \left\{
  \begin{array}{lcl}
    1 & \text{with probability } & \frac{1}{2s} \\
    0  & \text{with probability } & 1 - \frac{1}{s} \\
    -1 & \text{with probability } & \frac{1}{2s}
  \end{array}
  \right..
\end{equation}
Note that setting~$s=1$ and $s=3$ yields Achlioptas' matrices from Eq.~\ref{eq:rp_achl1} and Eq.~\ref{eq:rp_achl3},
respectively. \citeauthor{li2006very} showed, however, that one can use~$s$ as high as~$\frac{d}{\log{d}}$, if the data
follows a  normal distribution. To maintain a robust embedding, they recommend using lower $s$: $s = \sqrt{d}$, which
still significantly sparsifies the projection matrix and greatly speeds up the projection. Therefore, in this work we
use $s = \sqrt{d}$ when performing Li's projection.

\subsection{Subsampled randomized Hadamard transform}
\label{sec:rp_srht}

While the sparsification of $\mathbf{P}$ introduced by \citeauthor{li2006very} enables faster $\tilde{\mathbf{A}}$
computation, it severely distorts the distances between the observations when $\mathbf{A}$ is also sparse. This problem
was noticed and tackled by \citet{ailon2006approximate}. They proved that projection can be performed using a sparse
matrix only if the input data vectors are ``well-spread''. Data vector $\mathbf{x}$ of length $d$ is well-spread if
$\max\limits_{1\leq i\leq d} x_i$ is close to $\frac{1}{\sqrt{d}}$, i.e., $\mathbf{x}$ does not contain few non-zero
components with large absolute values. In order to assure that the input data is well-spread,
\citeauthor{ailon2006approximate} proposed to transform it with a generalized fast Fourier transform, called the
Walsh-Hadamard transform. After the transformation, the data can be safely projected using a highly sparse matrix
without introducing large distortions. This resulted in an efficient embedding scheme suitable for sparse input data,
the so-called \gls{SRHT} or alternatively the fast Johnson-Lindenstrauss transform.

\citeauthor{ailon2006approximate} defined their projection matrix~$\mathbf{P}_{\mathrm{SRHT}}$ as a scaled product of
three matrices:
\begin{equation}
  \mathbf{P}_{\mathrm{SRHT}} = \frac{1}{\sqrt{k}}\mathbf{DHS},
\end{equation}
where:
\begin{itemize}
  \item $\mathbf{D}$ is a~$d \times d$ diagonal matrix with random entries drawn uniformly from~$\{1, -1\}$,
  \item $\mathbf{H}$ is a~$d \times d$ normalized Hadamard-Walsh matrix:~$\mathbf{H} = \sqrt{\frac{1}{d}}\mathbf{H}_d$.
  The Hadamard-Walsh matrix~\mbox{$\mathbf{H}_t \in \mathbb{R}^{t \times t}$} is defined recursively as:
  \begin{equation}
  \mathbf{H}_1 = 1, \qquad
  \mathbf{H}_t =
    \begin{bmatrix}
      \begin{array}{lr}
        \mathbf{H}_{t/2} & \mathbf{H}_{t/2}\\
        \mathbf{H}_{t/2} & -\mathbf{H}_{t/2}
      \end{array}
    \end{bmatrix},
  \end{equation}
  for any~$t$ that is a power of two. If the dimensionality of the data $d$ is not a power of two, $\mathbf{A}$ can be
  padded with columns of zeros.
  \item $\mathbf{S}$ is a sparse~$d \times k$ random matrix, whose elements $S_{ij}$ are:
\begin{equation}
S_{ij} = 
  \left\{
  \begin{array}{lcl}
    0                                                    & \text{with probability } & 1 - q \\
    \text{value drawn from } \mathcal{N}(0, \frac{1}{q}) & \text{with probability } & q
  \end{array}
  \right.,
\end{equation}
where~$q = \mathcal{O}(d^{-1}\log^2 n)$ is a sparsity parameter. Some authors, e.g., \citet{matouvsek2008variants} also
experimented with replacing the normal distribution with a distribution similar to the one proposed by Achlioptas' to
speed up the generation of $\mathbf{S}$.
\end{itemize}

The strength of \gls{SRHT} lies in the fact that the product $\mathbf{xH}$ for a $d$-dimensional input vector
$\mathbf{x}$ can be calculated in just $\mathcal{O}(d\log{d})$ operations, by using the fast Fourier transform
algorithm. Therefore, the product $\mathbf{A}\mathbf{P}_{\mathrm{SRHT}}$ can be computed in $\mathcal{O}(nd\log{d})$, as
opposed to~$\mathcal{O}(ndk)$ if the projection was done by a naive matrix multiplication. \citet{ailon2009fast} further
improved the running time of \gls{SRHT} to~$\mathcal{O}(nd\log{k})$.

Quality of the \gls{SRHT} embedding is slightly worse than the quality provided by the Gaussian matrix -- \gls{SRHT}
satisfies the \gls{OSE} property for $k = \mathcal{O}(\epsilon^{-2}(n+d)\log{n})$ with high
probability~\citep{woodruff2014sketching}.

Very recently the Hadamard-Walsh matrix has received attention in the context of constructing dense structured random
matrices whose columns are orthogonal. Such constructions, e.g., the orthogonal Johnson-Lindenstrauss transform proposed
by \citet{choromanski2017unreasonable} provide promising theoretical guarantees on the embedding quality.

\subsection{Count Sketch-based projections}

An important recent family of \gls{RP} methods stems from the Count Sketch algorithm. The Count Sketch algorithm was
initially proposed by \citet{charikar2004finding} as a method to estimate the frequency of items in a data stream.
\citet{weinberger2009feature} and \citet{shi2009hash} applied it as a dimensionality reduction technique. The explicit
form of the projection matrix was then presented by \citet{dasgupta2010sparse}. The Count Sketch projection matrix, also
called the \textit{sparse embedding matrix}, can be given as:
\begin{equation}
  \mathbf{P}_{\mathrm{CS}} = \mathbf{DC},
\end{equation}
where~$\mathbf{D}$ is defined as in \gls{SRHT}, and~$\mathbf{C}$ is a~$d \times k$ sparse matrix with each row chosen
randomly from the~$k$ standard basis vectors of~$\mathbb{R}^k$. Such distribution, where a single element in each row is
picked independently at random and randomly set to either $1$ or $-1$, while the other elements are set to zero, is also
called the Rademacher distribution.

Projection using the Count Sketch scheme can be performed without a naive multiplication of the data matrix by
$\mathbf{P}_{\mathrm{CS}}$. Instead, the Count Sketch projection can be realized in linear time, i.e.,
$\mathcal{O}(nd)$. First, the result matrix $\tilde{\mathbf{A}}$ is initialized to zeros. Then each column of the data
matrix $\mathbf{A}$ is multiplied by $-1$ with probability $50\%$ and is added to a randomly selected column of
$\tilde{\mathbf{A}}$. When the input data is sparse, the time complexity of the Count Sketch projection can be decreased
to $\mathcal{O}(\mathrm{nnz}(\mathbf{A}))$~\citep{clarkson2013low}, where $\mathrm{nnz}(\mathbf{A})$ is the number of
non-zero elements in the matrix $\mathbf{A}$. Because of its low computational cost, this projection method and its
modifications have drawn considerable attention~\citep{clarkson2013low,meng2013low,nelson2013osnap}. Random projections
that are based on hash functions similar to Count Sketch have been employed in many practical machine learning
applications, e.g., in Vowpal Wabbit developed by Microsoft~\citep{langford2007vowpal}. For an overview of related
hashing techniques see~\citep{wang2017survey}.

The only significant disadvantage of Count Sketch-based \gls{RP} is its worse lower bound for $k$: the subspace
embedding property is satisfied for $k = \mathcal{O}(\epsilon^{-2}n^2)$~\citep{meng2013low,nelson2013osnap}.

\section{Summary}

\setlength{\tabcolsep}{6pt}
\begin{table}[htb]
\begin{minipage}{\linewidth}
  \caption{Properties of random projection schemes. $\mathbf{A}$ is a $n \times d$ dataset matrix and $k$ is the
  projected dimensionality. For sparse matrices $\mathbf{A}$, $\mathrm{nnz}(\mathbf{A})$ denotes the number of non-zero
  elements in $\mathbf{A}$. Embedding quality is the dimensionality for which the oblivious subspace embedding property
  is satisfied.}
  \label{tab:rp_complexity}
  \centering
  \begin{tabular}{cccccc} \toprule
    RP scheme & \begin{tabular}{@{}c@{}} Matrix \\ construction \\ time \end{tabular} &
      \multicolumn{2}{c}{Projection time} & Embedding quality \\ \cmidrule{3-4}
                 & & Dense input    & Sparse input & \\ \midrule
    Gaussian     & $\mathcal{O}(dk)$ & $\mathcal{O}(ndk)$ & $\mathcal{O}(\mathrm{nnz}(\mathbf{A})k)$ &
      $\mathcal{O}(\epsilon^{-2}n)$ \\
    Achlioptas'  & $\mathcal{O}(dk)$ & $\mathcal{O}(ndk)$ & $\mathcal{O}(\mathrm{nnz}(\mathbf{A})k)$ &
      -\footnote{For Achlioptas' and Li's projections, we did not find any estimates of $k$ for which the OSE property
      holds.} \\
    Li's         & $\mathcal{O}(\sqrt{d}k)$ & $\mathcal{O}(n\sqrt{d}k)$ &
      $\mathcal{O}(\mathrm{nnz}(\mathbf{A})k)$\footnote{For a standard sparse-dense matrix multiplication
      implementation.} & $\text{-}^a$ \\
    SRHT         & $\mathcal{O}(dk + d\log{d})$ & $\mathcal{O}(nd\log{k})$ & $\mathcal{O}(nd\log{k})$ &
      $\mathcal{O}(\epsilon^{-2}(n+d)\log{n})$ \\
    Count Sketch & $\mathcal{O}(d)$ & $\mathcal{O}(nd)$ & $\mathcal{O}(\mathrm{nnz}(\mathbf{A}))$ &
      $\mathcal{O}(\epsilon^{-2}n^2)$ \\
    \bottomrule
  \end{tabular}
\end{minipage}
\end{table}

We summarize the important properties of the presented \gls{RP} schemes in Table~\ref{tab:rp_complexity}. Specifically,
for each scheme, we focus on the time complexity of creating the projection matrix, the time complexity of the
projection and the embedding quality. Regardless of the projection scheme, the time required to create the projection
matrix is negligible when compared to the projection time. It can, however, be a factor if the projected dataset is very
sparse, i.e., when $\mathrm{nnz}(\mathbf{A}) \sim d$, which may be the case for some real-world datasets. For \gls{RP}
schemes that employ simple random number distributions, the construction time of $\mathbf{P}$ is proportional to the
number of non-zero matrix elements $\mathrm{nnz}(\mathbf{P})$. Therefore, creating a Gaussian or Achlioptas' matrix
requires $\mathcal{O}(dk)$ operations, while generating a sparser Li's matrix requires $\mathcal{O}(\sqrt{d}k)$
operations. The construction cost of an \gls{SRHT} matrix is $\mathcal{O}(dk + d \log{d})$ -- the sum of the cost of
creating a sparse $d \times k$ matrix and the cost of computing its fast Walsh-Hadamard transform, which is loglinear in
$d$. For Count Sketch, the cost is $\mathcal{O}(d)$, since we only need to generate the position of the non-zero element
in each of $d$ rows.

For simple projection schemes, i.e., Gaussian, Achlioptas' and Li's constructions the projection time depends only on
the employed matrix multiplication algorithm, which in turn depends on the sparsity of the multiplied matrices. If both
data and projection matrices are dense we assume the use of the schoolbook matrix multiplication algorithm, which runs
in $\mathcal{O}(ndk)$ time. We are aware that there exist more elaborate and slightly faster techniques, such as the
Coppersmith and Winograd-like algorithms~\citep{greiner2012sparse,le2014powers}. However, their use is beneficial mostly
when computing a product of matrices that are square. In our case, i.e., when $n \sim d$ and $d \gg k$, these methods do
not offer much improvement over the naive implementation. When $\mathbf{A}$ is sparse we can improve the Gaussian,
Achlioptas' and Li's projection time to $\mathcal{O}(\mathrm{nnz}(\mathbf{A})k)$. Additionally, if non-zero elements in
$\mathbf{A}$ are evenly distributed among its columns, the cost of Li's projection can be further decreased to
$\mathcal{O}(\mathrm{nnz}(\mathbf{A})\mathrm{nnz}(\mathbf{P})k^{-1})$, by using certain sparse-sparse matrix
multiplication algorithms. For an overview of such methods see~\citep{yuster2005fast,greiner2012sparse}. Sparse-dense
matrix multiplication can also be applied when projecting dense data with a sparse Li's matrix. Because the number of
non-zero elements in Li's projection matrix is, on average, $\frac{dk}{\sqrt{d}} = \sqrt{d}k$, projecting dense data
with Li's scheme is \mbox{$\mathcal{O}(\mathrm{nnz}(\mathbf{P})n) = \mathcal{O}(n\sqrt{d}k)$}, where $\mathbf{P}$ is a
$d \times k$ Li's projection matrix.

In general, we can observe that more elaborate projection schemes, such as \gls{SRHT} or Count Sketch offer much better
computation time in exchange for embedding quality. However, this does not necessarily mean that these methods introduce
too high distortions in the projected space to be useful for practical applications. On the contrary -- recent
experimental results with training linear classifiers on randomly projected data indicate that Count Sketch \gls{RP} is
an excellent choice, especially when the input data is sparse~\citep{paul2014random}.

%% file: rp_layer.tex
\cleardoublepage
\chapter{Training deep networks with random projection layer}
\label{cha:rp_layer}

One factor that limits the number of learnable parameters in a neural network is the computational cost of the training.
In certain applications, this number can be vastly reduced by exploiting the structure of the data. However, for
unstructured data or data where the structure is not well defined, it might be impossible to constrain the network
architecture and limit the number of parameters. In these cases learning directly from the input data requires employing
a network architecture that uses input layer with as many units as the number of features in the input vectors. When the
input dimensionality exceeds several thousand, such an input layer becomes too large to be trained in a reasonable time.
One solution to this problem is reducing the dimensionality of the input space to a manageable size and then training a
deep network on representation with fewer dimensions. Here, we focus on performing the dimensionality reduction step by
randomly projecting the input data into a lower-dimensional space. Conceptually, this is equivalent to adding a layer,
which we call \textbf{the \acrlong{RP} layer}, in front of the network. Several computationally efficient \gls{RP}
matrix constructions have been recently proposed (see Section~\ref{sec:rp_schemes}), leading to simple dimensionality
reduction methods that scale to even millions of dimensions while introducing a controlled amount of noise. These can,
therefore, be used to efficiently train networks on input data with a huge number of dimensions.

We study two variants of \gls{RP} layers: one where the parameters of the \gls{RP} layer are fixed during training and
one where they are finetuned with error backpropagation. The first variant, further called \textbf{fixed-weight \gls{RP}
layer}, can be interpreted as training a standard network architecture on data whose dimensionality has been reduced
with \acrlong{RP}. A theoretical motivation for learning on such randomly projected data has been given
in~\citep{arriaga2006algorithmic,hegde2007efficient}. Particularly, \citeauthor{arriaga2006algorithmic} provided a clear
motivation that can be summarized in two points: (i) learning from randomly projected data is possible since \gls{RP}
preserves a lot of the input structure in the lower-dimensional space, and (ii) learning in the lower-dimensional space
should require fewer examples, and thus be faster. The second \gls{RP} layer variant, which we call \textbf{the
learnable \gls{RP} layer}, or more precisely \textbf{the finetuned \gls{RP} layer}, may improve the network performance,
compared to fixed-weight \gls{RP} layer. However, it has a significantly higher computational cost. Nevertheless, we
show that with carefully designed architecture and training regime it can be applied to real-world problems.

\glsreset{SVM}
\glsreset{LR}

By incorporating \gls{RP} into the network architecture we enable \glspl{DNN} to be trained on a particularly
challenging type of data -- data whose representation is simultaneously sparse, unstructured and high-dimensional. This
opens an opportunity for applying \glspl{DNN} in tasks where learning has previously been restricted to simple linear
methods, such as \gls{SVM} or \gls{LR} classifiers. For a short survey of application areas where sparse, unstructured,
high-dimensional data can be found see Section~\ref{sec:sparse_high_unstuct_data}.

This chapter is organized as follows. In Section~\ref{sec:nn_with_rp_layer}, we start by presenting the concept of the
fixed-weight \gls{RP} layer, its motivation and related approaches. In Section~\ref{sec:rpsynth_eval}, we evaluate the
performance of the fixed-weight \gls{RP} layer on large synthetic data. In particular, we investigate how the layer
dimensionality and properties of the learned data, such as its sparsity or the fraction of significant features
influence the training process. Additionally, we compare the performance and computational cost of our approach with
different baseline techniques. Then, in Section~\ref{sec:rp_real_world}, we conduct large-scale experiments on several
real-world datasets and show the effectiveness of neural networks with \gls{RP} layers in comparison to the
state-of-the-art methods. In Section~\ref{sec:rp_bow}, we explore the prospects of employing \glspl{DNN} with
fixed-weight \gls{RP} layers for learning from \gls{BOW} data. In Section~\ref{sec:lrp_layer}, we discuss in detail how
the weights in the \gls{RP} layer can be learned. We focus on techniques that enable us to reduce the training cost and
make this task feasible. Then, in Section~\ref{sec:lrp_experiments} we experimentally evaluate the performance of deep
networks with finetuned \gls{RP} layers on large synthetic and real-world datasets. In particular, we investigate the
influence of different normalization schemes on their performance and the prospects of applying a nonlinear activation
after the \gls{RP} layer. Finally, in Section~\ref{sec:rp_impl_details}, we discuss the important implementation
decisions and details.

Selected results from this chapter were presented in~\citep{wojcik2018training}.

% comparison of supervised learning techniques trained on reduced data with PCA and RP
% RP works well with SVM and Nearest Neighbour methods, decision trees + RP less good
% \citep{fradkin2003experiments}

\section{Fixed-weight random projection layer}
\label{sec:nn_with_rp_layer}

\begin{figure}[htb!]
  \centering
  \includegraphics[width=.7\linewidth]{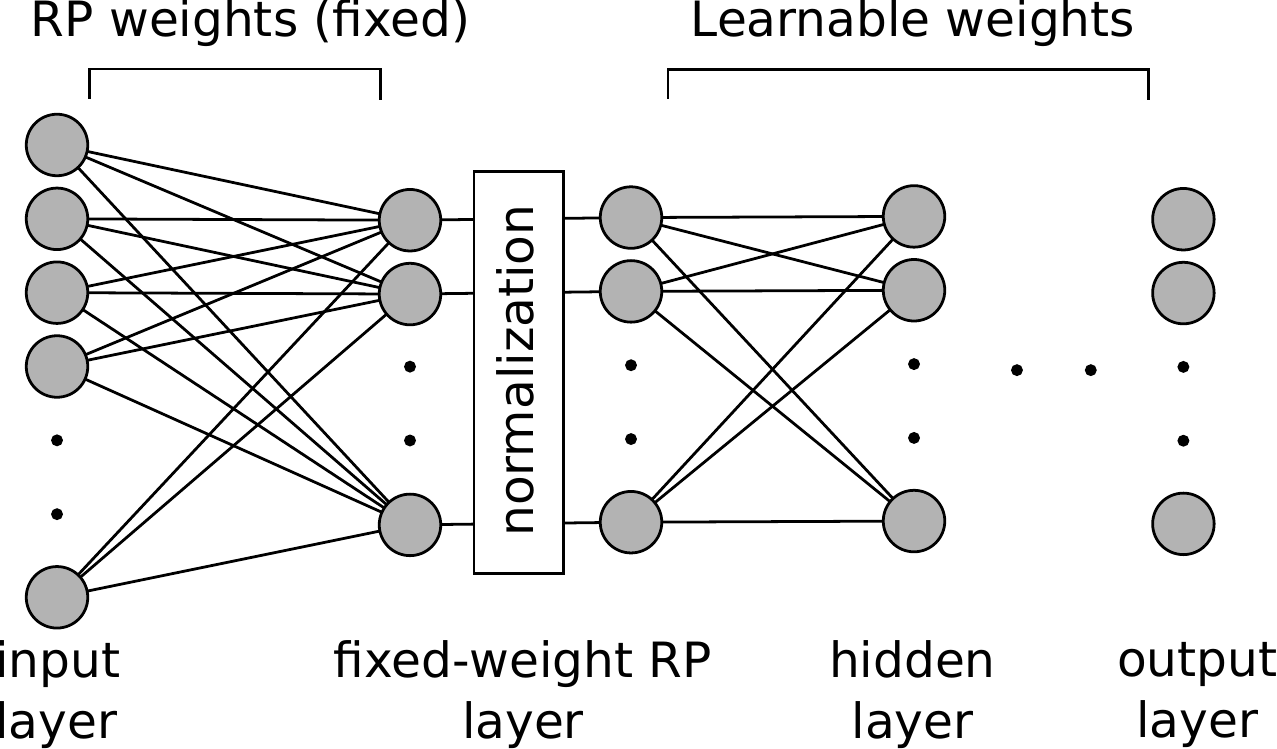}
  \caption{Neural network with fixed-weight random projection layer.}
  \label{fig:rp_network}
\end{figure}

We begin with the analysis and evaluation of \glspl{DNN} with fixed-weight \gls{RP} layers. By fixed-weight \gls{RP}
layer we mean a layer in which the weights are not updated during training, but instead, are fixed to an \gls{RP} matrix
(Fig.~\ref{fig:rp_network}). There are two main reasons for using fixed weights in the \gls{RP} layer. First, this
enables us to perform the projection and normalization of the whole dataset only once prior to the network training.
Such optimization is especially beneficial when the lower-dimensional projection fits in the operating memory, while the
original input data requires out-of-core processing. Second, for dense \gls{RP} constructions, such as Gaussian,
Achlioptas' or \gls{SRHT}, an update of the weights in the \gls{RP} layer may have a prohibitively high computational
cost: for example, dense \gls{RP} matrices for some of the tasks reported in Section~\ref{sec:rp_real_world} have up to
tens of billions of weights -- several times more than the number of weights in the largest currently used
networks~\citep{coates2013deep}\footnote{Note that the approach presented by \citet{coates2013deep} was designed
specifically to train the largest networks (with up to 11 billion parameters) and requires a cluster of GPU servers with
Infiniband connections and MPI. Deep networks that recently achieved the state-of-the-art results on popular benchmarks
are typically at least an order of magnitude smaller in terms of the parameter count. For example, the VGGNet
network~\citep{simonyan2014very} contains ``just'' 140 million learnable weights and takes 2--3 weeks to be trained on
four NVIDIA Titan Black GPUs.}. Finetuning weights of the \gls{RP} layer is more practical for sparse \gls{RP}
constructions, especially if we restrict the updates to the weights that are initially non-zero. We further investigate
this approach in Section~\ref{sec:lrp_layer}.

Layers that follow the fixed-weight \acrlong{RP} can be trained from scratch using error backpropagation. However, we
found that the performance can be improved by pretraining these layers with \glspl{DBN} and then finetuning them with
error backpropagation. Before the projected data is used to train the ``learnable'' part of the network, we normalize
each dimension to zero mean and unit variance. Initial evaluation showed that this is necessary: pretraining on
unnormalized data was unstable, especially for highly sparse datasets. One particular advantage of pretraining a
\gls{DBN} on normalized data is that, regardless of the input data type, we can use Gaussian
units~\citep{welling2004exponential} in the first layer of the ``learnable'' part of the network.

The choice of the projection matrix in the \gls{RP} layer is not trivial and depends on the dimensionality and sparsity
of the input data. In particular, these two factors have a significant impact on the computational cost of training.
While the projection time is usually negligible in comparison to the training time, this may not be the case when the
data dimensionality is very high. Fortunately, especially for large unstructured datasets, high dimensionality often
goes hand in hand with high sparsity. This is beneficial from the computational point of view since sparse
representation enables us to perform the projection faster. In particular, the performance of RP schemes that involve
matrix multiplication can be improved by fast algorithms for sparse matrix
multiplication~\citep{bank1993sparse,greiner2012sparse}. Some other RP schemes can also be optimized to take advantage
of the data sparsity~\citep{clarkson2013low}.

Another aspect to consider is the sparsity of the projection matrix itself. \Acrlong{RP} matrices that provide the best
quality of embedding are typically dense~\citep{nelson2014lower}. Unfortunately, applying dense projection schemes to
huge datasets can be computationally prohibitive. In this case, one needs to resort to more efficient projection
schemes. One possibility is to employ a projection scheme that does not require naive matrix multiplication. A good
example of such projection scheme is \gls{SRHT}. Another approach is to use a sparse projection matrix, as in, e.g.,
Li's construction. Moreover, these two approaches can be combined into a projection scheme, where the \gls{RP} matrix is
sparse and the projection does not require an explicit matrix multiplication. This results in very efficient projection
methods, such as the Count Sketch projection. However, projecting sparse data with sparse \gls{RP} matrices, regardless
if they are explicitly or implicitly constructed, can introduce significant distortions in the
embedding~\citep{ailon2006approximate}. These distortions may, in turn, affect the network accuracy. Therefore, for
large datasets, the choice of the \gls{RP} layer type is a trade-off between the network accuracy and the computational
complexity of the \gls{RP} embedding. Investigating this trade-off, apart from enabling training of neural networks on
high-dimensional data is one of the goals of this work.

\paragraph{Related approaches.}

The idea of using fixed random weights in neural networks is not new and has been incorporated into different models
proposed throughout the years. Note, however, that not every layer with random weights realizes a \acrlong{RP} (see
Chapter~\ref{cha:rp}). One important family of shallow networks employing random weights are the \glspl{RW-FNN}. These
models differ from our approach in two important aspects. First, instead of lowering the data dimensionality, they
transform the input data into a higher-dimensional space in which learning should, theoretically, be easier.
Importantly, this transformation is most often nonlinear and, in general, does not preserve the distances between the
training examples. Additionally, after randomly transforming the input, \glspl{RW-FNN} do not employ any feature
normalization. Second, \glspl{RW-FNN} cast the weight optimization problem as a standard regularized least-squares
problem, which can be solved analytically in a single step. While this approach offers a computational advantage
compared to stochastic gradient descent, it is suitable only for networks with a single hidden layer. For a more
comprehensive overview of \glspl{RW-FNN} see~\citep{scardapane2017randomness}. Predecessors of these models were
proposed in a number of early works on feedforward architectures, e.g.
in~\citep{gallant1987random,schmidt1992feedforward}. A more mature version of \glspl{RW-FNN}, called random vector
functional-link networks was introduced in~\citep{pao1992functional,pao1994learning}.

In an interesting work~\citet{arriaga2006algorithmic} suggest that the human brain may reduce the amount of information
generated by visual stimuli in a process that resembles \acrlong{RP}. They show that \gls{RP} can be realized by a
shallow neural network with weights drawn from a Gaussian distribution or just set randomly to -1 or 1 (note that this
is the denser variant of the Achlioptas' construction~\citep{achlioptas2001database}, which we described in
Section~\ref{sec:rp_schemes}). \citeauthor{arriaga2006algorithmic} use this so-called neuron-friendly \gls{RP} to show
that efficient learning is possible in the projected space. However, similarly to \glspl{RW-FNN}, they do not train
deeper models on the projected data and use a simple learning algorithm instead of error backpropagation.

To the best of our knowledge the only attempt at training \glspl{DNN} on randomly projected large-scale data, and
therefore the approach that is most relevant to our fixed-weight \gls{RP} layers, is presented in~\citep{dahl2013large}.
Therein, \citeauthor{dahl2013large} use randomly projected data as input to networks trained for the malware
classification task. Specifically, they project the original~$179,000$-dimensional data (trigrams of system API calls)
to~$4000$ dimensions and use the projected data to train a neural network with two hidden layers. With this approach,
they achieve~$43\%$ relative improvement in classification performance, compared to logistic regression trained on the
unprojected data. However, their classification task is fairly simple, with the classes being nearly linearly separable.
Unfortunately, \citeauthor{dahl2013large} only evaluate Li's matrix construction~\citep{li2006very}, which is extremely
sparse and, from our experience, is unsuited for projecting sparse n-gram data (see Section~\ref{sec:rp_bow}). It is
also worth mentioning that in their experiments unsupervised pretraining does not improve network performance, unlike in
experiments reported in our work. Additionally, \citeauthor{dahl2013large} do not employ data normalization after the
projection, whereas our experiments show that scaling each feature to zero mean and unit variance greatly helps,
especially when training networks on sparse data. Finally, \citeauthor{dahl2013large} evaluate only networks with the
sigmoid activation function and do not report results for the current state-of-the-art \gls{ReLU}
activation~\citep{nair2010rectified}.

Another recent work by \citet{choromanska2016binary} explore a similar approach to our networks with fixed-weight
\gls{RP} layer. Specifically, the authors consider networks in which the first layer uses untrained pseudo-random
weights and a nonlinear activation function. However, their approach differs from ours in two ways. First,
\citeauthor{choromanska2016binary} mostly focus on using structured pseudo-random matrices, such as various
modifications of the circulant or Toeplitz matrices and not classic \gls{RP} matrices. Second, since they consider
layers that realize binary embeddings of the input data, they only employ the sign activation function. Importantly, the
results presented in \citep{choromanska2016binary} show that using fully-random matrices, such as the Gaussian \gls{RP}
matrix, for projecting the input data yields better performing networks than using pseudo-random structured matrices.

Random weight matrices were also used in certain convolutional neural network architectures~\citep{saxe2011random}. In
particular, \citeauthor{saxe2011random} report convolutional networks with random weights that perform only slightly
worse than networks with learned parameters. This inspired us to investigate the prospects of using \gls{RP} schemes to
initialize weights in deep networks. We elaborate on this topic in Chapter~\ref{cha:rp_init}.

Apart from neural networks other models also have been successfully trained on randomly projected data. In particular,
\citet{paul2014random} evaluated \gls{SVM} classifiers and regression models on data projected with several \gls{RP}
schemes and achieved promising results on small- and medium-size datasets. Similarly to our results, they also found
Count Sketch to be one of the best performing \gls{RP} methods.

\subsection{Experiments on large-scale synthetic data}
\label{sec:rpsynth_eval}

To evaluate the performance of the fixed-weight \gls{RP} layer we begin with experiments on the synthetic datasets
described in Appendix~\ref{cha:datasets}. Each synthetic dataset variant consists of one million training examples
represented by one-million-dimensional feature vectors. First, we investigate the influence of the \gls{RP} layer
dimensionality on the network performance. We also analyze the computational cost of employing different types of
\gls{RP} layers. We then explore how the properties of the input data affect the training process. In particular, we
study the effects of the data sparsity and the effect of the fraction of features that are informative for learning. We
evaluate five \gls{RP} constructions presented in Section~\ref{sec:rp_schemes}, i.e., Gaussian, Achlioptas', Li's,
\gls{SRHT} and Count Sketch. We compare the effectiveness of \glspl{DNN} that employ fixed-weight \gls{RP} layers with
other baseline approaches to learning from sparse, high-dimensional data. Specifically, we experiment with replacing the
\acrlong{RP} with other fast dimensionality reduction methods, such as F-score, Chi-square, \gls{IG} and \acrlong{PCA}.
We also evaluate another baseline approach -- training efficient linear models directly on the high-dimensional feature
space. We discuss both the performance and the computational cost of the presented baseline approaches.

\subsubsection{Effects of the size of random projection layer}

We begin the evaluation with experiments in which we investigate the impact of the size of the \gls{RP} layer on the
network performance. These experiments were carried out on a synthetic dataset variant with density \mbox{$\rho =
10^{-5}$} and fraction of significant features $\psi = 0.2$ (Appendix~\ref{cha:datasets}).

We first generated the low-dimensional representations of the dataset using five evaluated \gls{RP} schemes. We reduced
the dataset dimensionality from one million to $k$ dimensions, where $k$ ranged between 10 and 1000. We then normalized
every feature in each projected dataset to zero mean and unit variance. We also experimented with scaling each feature
by its maximum absolute value or not applying any normalization of the projected data. However, these approaches
resulted in poorly performing models. For the details on \gls{RP} implementation and other implementation notes see
Section~\ref{sec:rp_impl_details}.

We trained deep networks on the projected datasets in two phases: unsupervised pretraining with \gls{CD} followed by
finetuning with error backpropagation. The networks were trained using mini-batch \gls{SGD} with momentum. Amplitudes of
the weights were limited with the L2 cost. During finetuning the learning rate was decreased according to a slow
exponential decay, while the momentum was slowly increased. We also used dropout to prevent overfitting. We employed
\glspl{ReLU} in the hidden layers and Gaussian units in the input layer. Validation sets constructed from the training
data were used to select the learning hyperparameters. For all input layer sizes, we used a network architecture with
two hidden layers, each consisting of $1000$ neurons. After pretraining, we added a logistic regression unit on top and
finetuned the networks to minimize the binary \gls{CE} cost.

\begin{figure}[!htb]
  \centering
  \includegraphics[width=\linewidth]{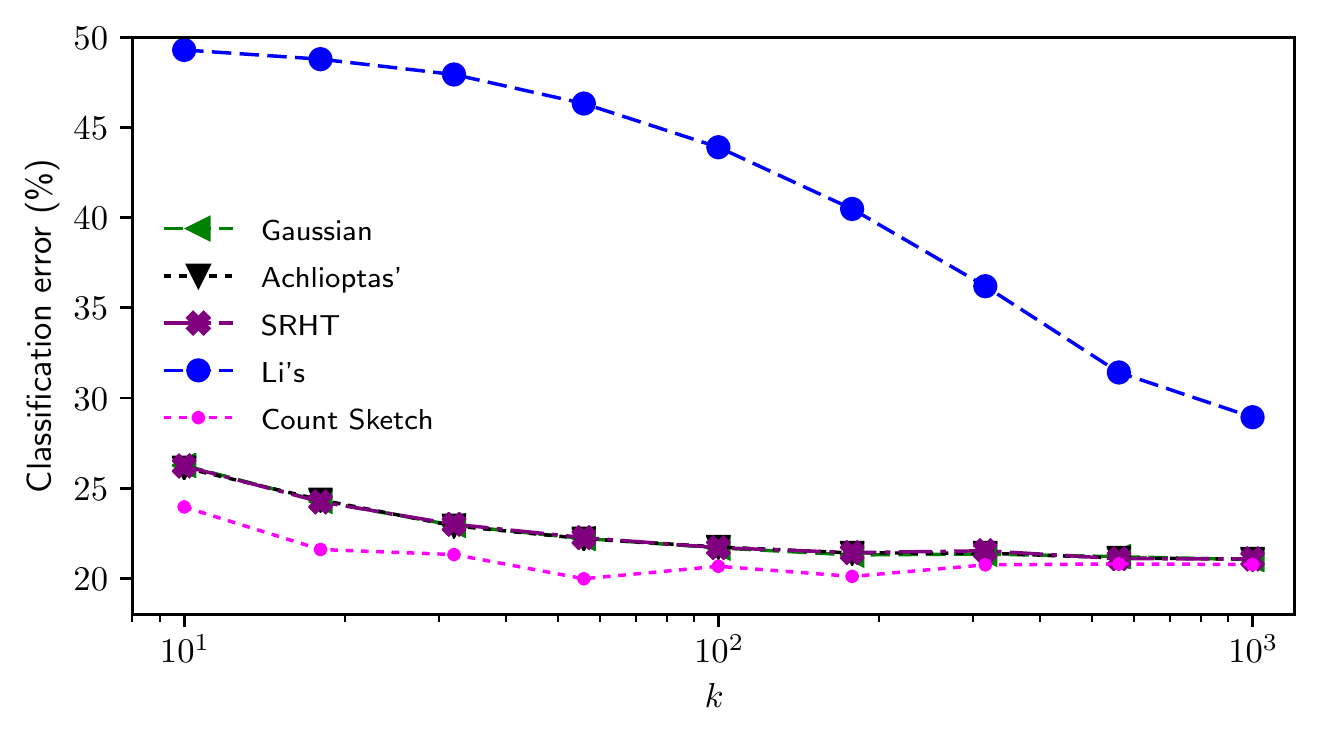}
  \caption{Classification error for different sizes of the random projection input layer~($k$). Gaussian, Achlioptas'
  and SRHT yielded almost equal results.}
  \label{fig:rpsynth_k}
\end{figure}

In Fig.~\ref{fig:rpsynth_k} we report the early stopping errors for different sizes of the \gls{RP} layer. Early
stopping error is the test error in the training epoch with the best performance on the validation set. In general,
increasing the value of $k$ improved the network performance. This agrees with the intuition derived from the
Johnson-Lindenstrauss lemma~\citep{johnson1984extensions} (see Section~\ref{sec:jl_lemma_ose}). If we assume that the
network classification error is correlated with the embedding error $\epsilon$, then for a constant number of examples
$n$, the network classification error should be correlated with $\frac{1}{\sqrt{k}}$. Therefore, the network error
should decrease when $k$ increases, as, in fact, can be seen in Fig.~\ref{fig:rpsynth_k}. For dense \gls{RP} schemes,
i.e., Gaussian, Achlioptas' and \gls{SRHT}, the best and almost equal results were achieved for the highest values of
$k$. Interestingly, Count Sketch performed slightly better than dense \glspl{RP}, especially for lower values of $k$.
For larger sizes of the \gls{RP} layer, its performance became comparable to dense \gls{RP} schemes. Li's projection was
significantly outperformed by the other four \gls{RP} schemes. This outcome cannot be attributed only to the high
sparsity of the projection matrix -- the Count Sketch projection matrix, which performed much better, had the same
number of non-zero elements as Li's projection matrix. While it has been argued that sparse projection matrices are not
suited for sparse data~\citep{ailon2006approximate}, these results demonstrate that matrix construction also plays an
important role. Note that, unlike Li's construction, columns in the Count Sketch projection matrix are fully orthogonal.
Moreover, orthogonal weight initialization has been shown to improve the performance of deep
networks~\citep{mishkin2015all}. We believe that this may be the reason behind the very good performance of the Count
Sketch \gls{RP} layer.

\begin{figure}[htb!]
  \centering
  \includegraphics[width=\linewidth]{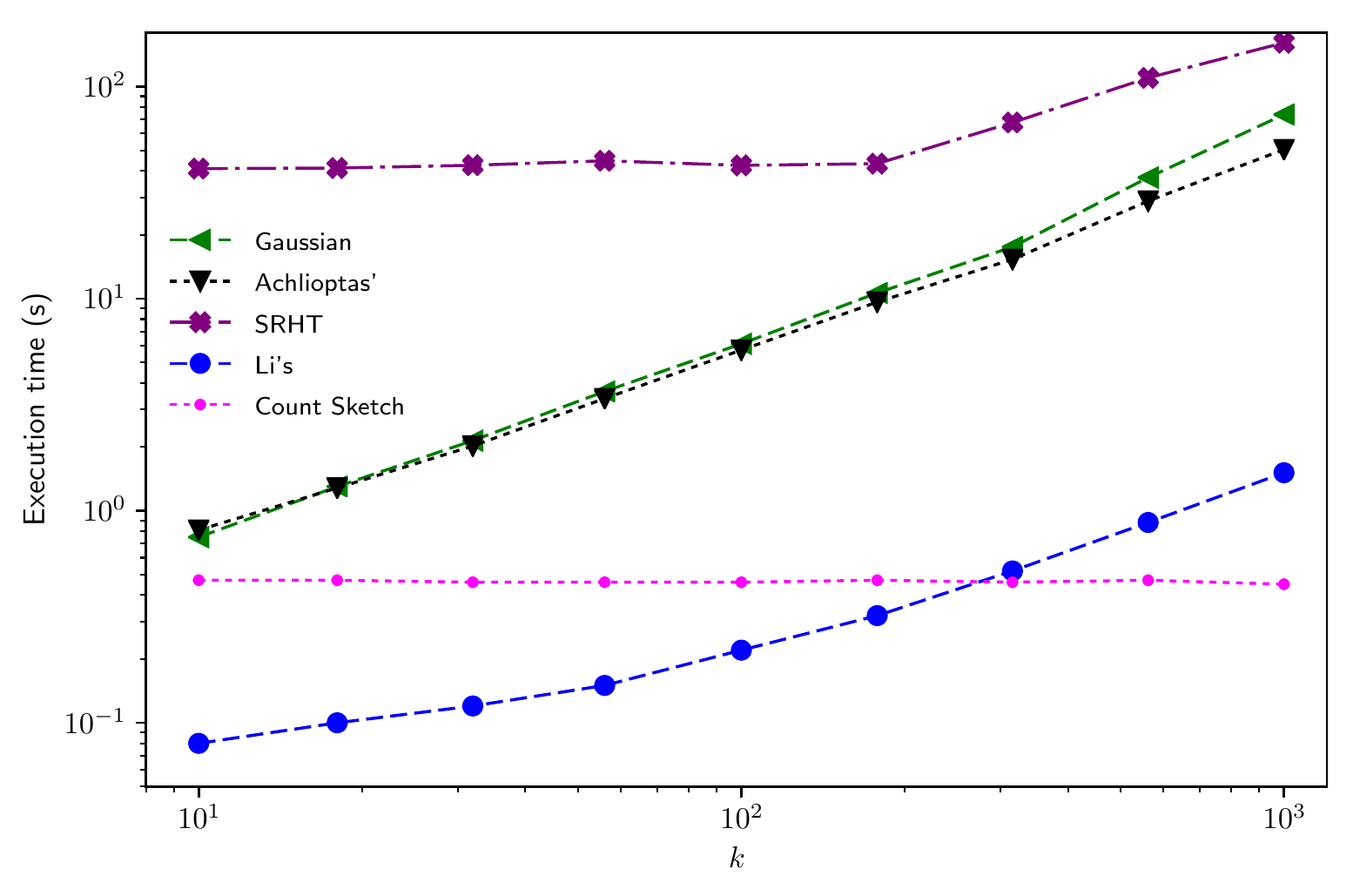}
  \caption{Average time of performing random projection on the million-dimensional synthetic dataset ($\rho = 10^{-5}$,
  $\psi = 0.2$) for different RP layer size~$k$.}
  \label{fig:rp_k_exec_time}
\end{figure}

In addition to network performance, we also investigated the computational cost of performing \gls{RP} for different
values of $k$. The experiments in this and the subsequent sections were carried out on a 64-bit machine with Intel Xeon
2.50GHz CPU (E5-2680 v3), 30MB cache and 128GB main memory. For every value of $k$, we ran the projection procedure five
times. The average \gls{RP} execution times are presented in Figure~\ref{fig:rp_k_exec_time}. For larger values of $k$,
sparse \gls{RP} schemes were significantly more efficient than dense schemes: for $k > 100$ they sped up the projection
more than ten-fold. The execution time of both Gaussian and Achlioptas' \glspl{RP} scaled linearly with the \gls{RP}
layer size. For the Count Sketch scheme the projection time was approximately constant in $k$, and for Li's and
\gls{SRHT} the dependence was more complex.

\setlength{\tabcolsep}{0pt}
\begin{figure}[htb!]
\begin{tabular}{cc}
  \includegraphics[width=0.5\linewidth]{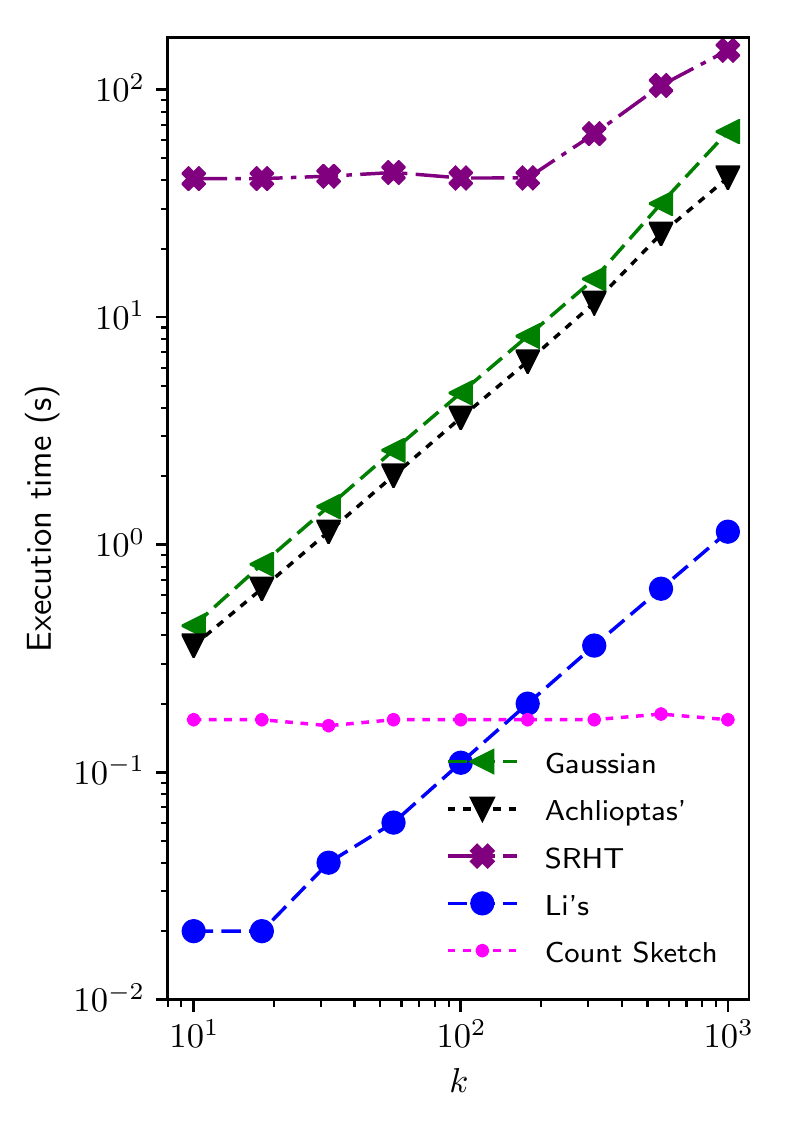} &
  \includegraphics[width=0.5\linewidth]{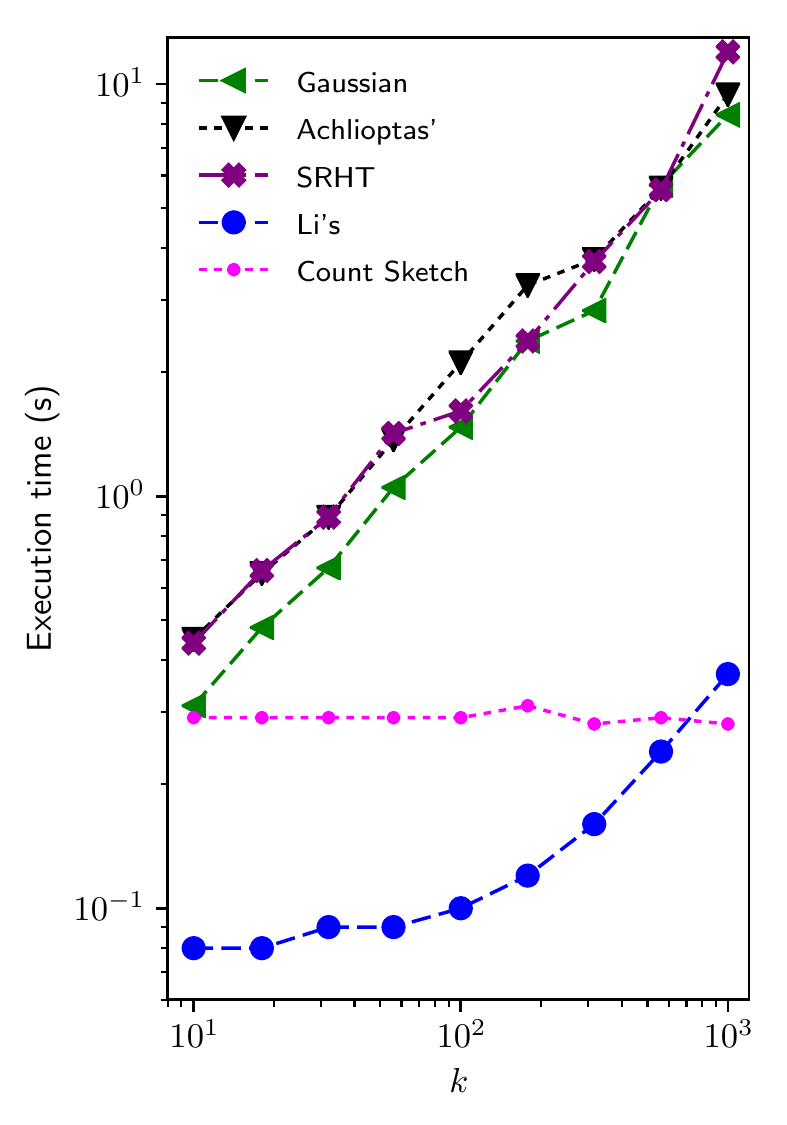} \\
  RP matrix generation & Matrix multiplication time \\
\end{tabular}
\caption{Average time of generating an RP matrix and performing the projection by matrix multiplication for the
synthetic dataset.}
\label{fig:rp_k_exec_time_gen_proj}
\end{figure}

To further explore these results, we separately measured the \gls{RP} matrix generation time and the projection time.
The results are presented in Figure~\ref{fig:rp_k_exec_time_gen_proj}. For most of the evaluated schemes, the
computational cost of creating the \gls{RP} matrix increases linearly with the size of the \gls{RP} layer. For Gaussian,
Achlioptas' and Li's matrices, this dependence is linear because the number of random numbers that need to be generated
is proportional to $k$. The computational time of generating the \gls{SRHT} projection matrix is also linear in $k$ --
this relation is, however, significantly offset by the high cost of performing the fast Walsh--Hadamard transform on a
matrix with, in our case, one million rows. The Count Sketch matrix, on the other hand, can be created in a constant
time. This is because the number of non-zero elements in the Count Sketch projection matrix is equal to the constant
data dimensionality $d$ and not to the projection dimensionality $k$. For Gaussian, Achlioptas' and \gls{SRHT}, the
projection time depended linearly on $k$. For Count Sketch \gls{RP} the dependence was approximately constant and for
Li's \gls{RP} it was nearly linear for higher values of $k$. This is a consequence of the type of the employed matrix
multiplication algorithm. For dense \gls{RP} schemes, the projection was realized with sparse-dense matrix
multiplication, which is linear in $k$. For sparse \gls{RP} matrices, we used a matrix multiplication procedure that
exploits the matrix sparsity. Specifically, we employed an implementation of the \gls{SMMP}
algorithm~\citep{bank1993sparse}. For a more detailed time complexity analysis see Section~\ref{sec:rp_impl_details}.

Note that we intentionally test \gls{RP} layers with sizes $k \le 1000$. There are several reasons why we do not
consider larger values of $k$. First of all, we observe that one thousand features extracted from the original
multi-million-dimensional feature space often captures most of the information that is useful for learning. This
hypothesis is supported by the good performance of deep networks with $k = 1000$ on several different real-world
datasets (see Section~\ref{sec:rp_real_world}). Second, restricting the \gls{RP} layer size limits the number of weights
in the \gls{DNN} which, in turn, decreases the training time. In our experiments on the synthetic datasets, network
training could be performed moderately fast, i.e., between $1$ and $3$ hours, depending on the \gls{RP} scheme. However,
for larger and more difficult real-world datasets, the network convergence may be significantly slower. In such cases,
the training may take several days. Finally, for large values of $k$, the available memory may become a limiting factor.
Randomly projecting even very high-dimensional data is computationally efficient, provided that the whole projection
matrix can be stored in the operating memory. However, for some datasets, this may be impossible, especially if the
projection matrix is dense. For example, projecting the $20$-million-dimensional \texttt{KDD2010-a} dataset to $k=2000$
dimensions requires nearly $160$GB of RAM for the projection matrix alone, which exceeds the available memory on most
modern machines. In such cases, the projection matrix has to be divided columnwise into smaller slices, which can be
stored and fetched on demand from the disk. Unfortunately, such out-of-core processing greatly increases the projection
time.

\subsubsection{Effects of data sparsity}

To evaluate the influence of the data sparsity on the performance of \glspl{DNN} with \gls{RP} layer we carried out
experiments on multiple synthetic dataset variants with density $\rho$ ranging from $10^{-6}$ to $1.7 \cdot 10^{-4}$ and
constant fraction of significant features $\psi = 0.2$. We chose these density values because they are representative of
typical high-dimensional real-world datasets, such as \texttt{url}, \texttt{KDD2010-a} or \texttt{webspam} (see
Table~\ref{tab:datasets} in Appendix~\ref{cha:datasets}).

\begin{figure}[htb!]
  \centering
  \includegraphics[width=\linewidth]{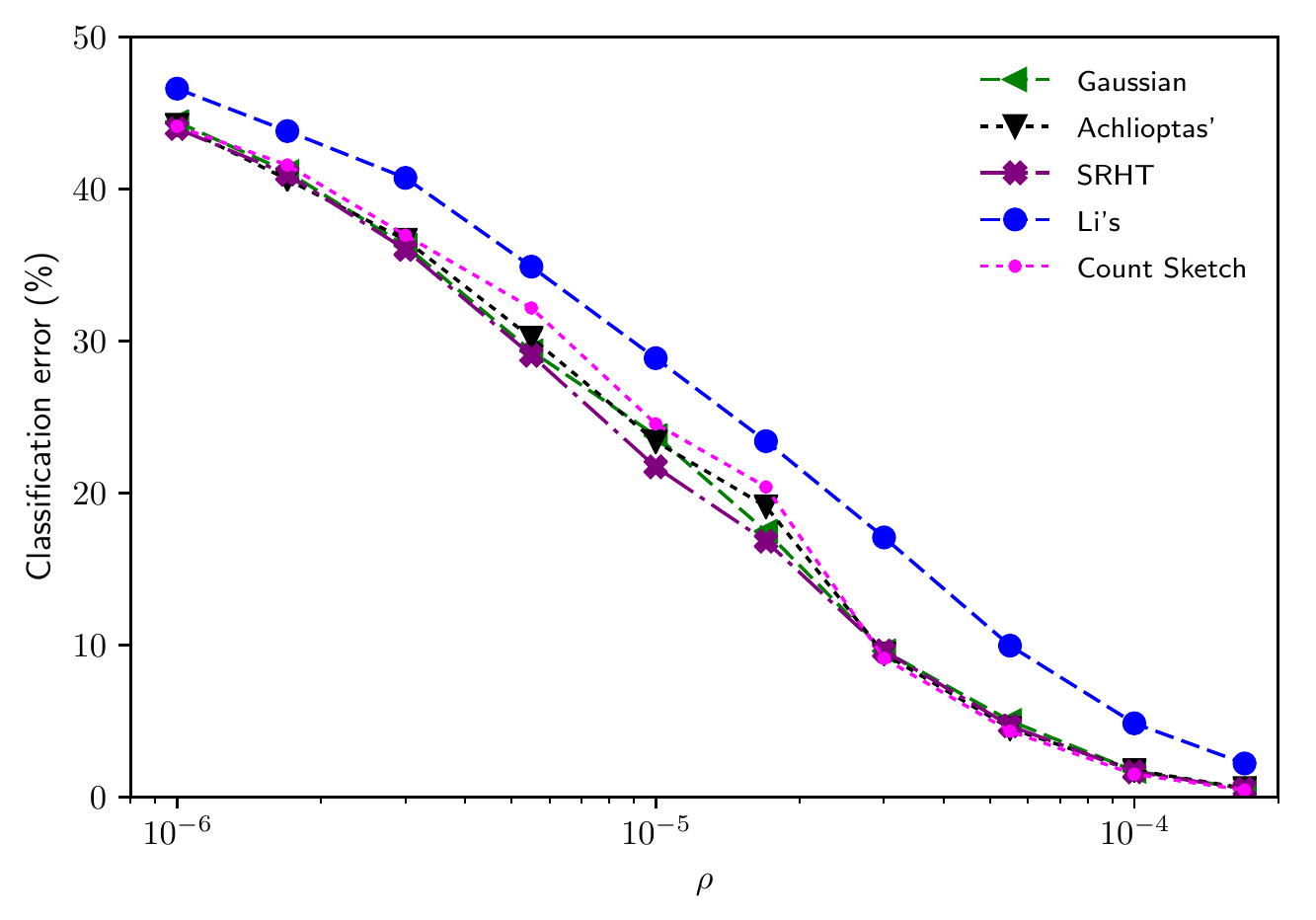}
  \caption{Classification error on the synthetic datasets with fixed significant feature fraction~\mbox{$\psi = 0.2$}
  and varying density level $\rho$.}
  \label{fig:rpsynth_rho}
\end{figure}

We generated $1000$-dimensional representations of the synthetic datasets and normalized each feature to zero mean and
unit variance. We then trained \glspl{DNN} on the lower-dimensional datasets following the training regime described in
the previous section. We used a network architecture with two hidden layers, each consisting of $3000$ neurons. In
Fig.~\ref{fig:rpsynth_rho} we report the early stopping errors for different density levels $\rho$. For all evaluated
\gls{RP} schemes, the accuracy of \glspl{DNN} depends nearly linearly on the logarithm of the dataset density. For most
density levels Gaussian, \gls{SRHT} and Achlioptas' projections yielded the best and quite similar results. The Count
Sketch projections performed slightly worse, especially for dataset variants with $\rho < 10^{-5}$. Similarly to the
previous experiments, for all evaluated density levels Li's projection was significantly outperformed by the other four
\gls{RP} schemes.

\subsubsection{Effects of the percentage of significant features}

Next, we carried out experiments to investigate the influence of the fraction of significant features on the network
performance. For these experiments, we employed synthetic dataset variants with fractions of significant features $\psi$
ranging from $0.01$ to $0.2$ and a constant density level $\rho = 10^{-4}$.

\begin{figure}[htb!]
  \centering
  \includegraphics[width=\linewidth]{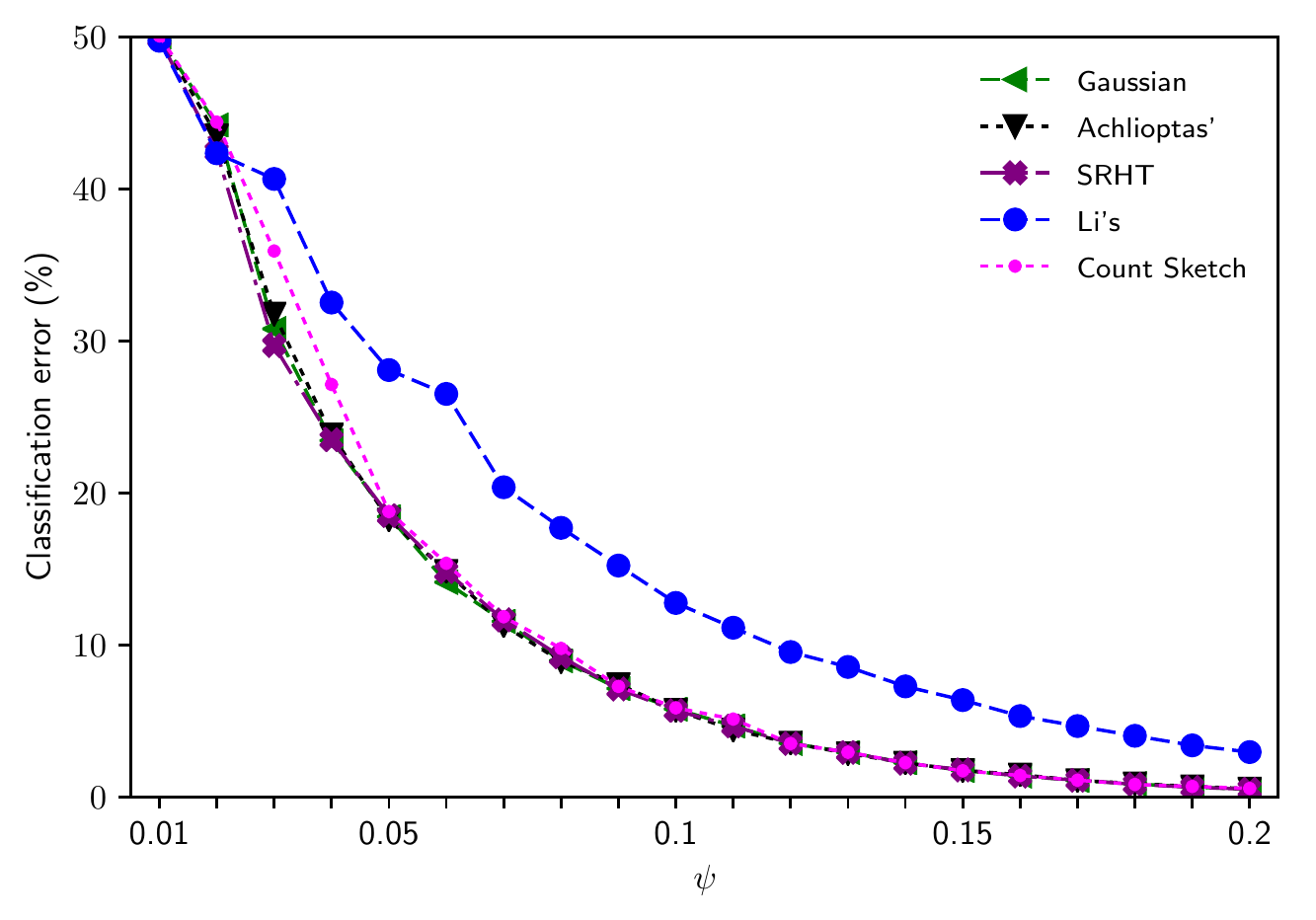}
  \caption{Classification error on the synthetic datasets with fixed density~\mbox{$\rho = 10^{-4}$} and varying
  fraction of significant features $\psi$.}
  \label{fig:rpsynth_psi}
\end{figure}

As in the previous section, we generated the $1000$-dimensional dataset representations and used them to train the
networks. We employed the same network architecture, i.e., two hidden layers with $3000$ units each. Early stopping
errors for different fractions of significant features are presented in Fig.~\ref{fig:rpsynth_psi}. The fraction of
significant features in the projected dataset had a strong impact on the performance of neural networks: networks
trained on dataset variants with larger $\psi$ performed much better, especially for lower values of $\psi$. For
example, for all \gls{RP} schemes except Li's, doubling the number of significant features from $5\%$ to $10\%$ reduced
the test error from approximately $18\%$ to less than $6\%$. For $\psi > 0.1$, all \gls{RP} schemes apart from Li's
yielded almost equal performance. For $\psi \le 0.1$, i.e., for dataset variants where each example contains on average
only $1 - 10$ significant features, dense \gls{RP} schemes performed better than sparse projections.

\subsubsection{Comparison with baseline approaches}

Training \glspl{DNN} on data whose dimensionality has been reduced with \gls{RP} is not the only viable approach to
learning models on sparse high-dimensional data. As we discussed in Section~\ref{sec:dr}, the dimensionality reduction
prior to network training can be performed with other efficient feature selection or feature extraction techniques.
Another even simpler approach is to train fast linear models directly on the high-dimensional feature space. In this
section we compare the performance and computational cost of such approaches.

For our tests we employ three synthetic dataset variants with density \mbox{$\rho \in \{3 \cdot 10^{-6}, 10^{-5}, 3
\cdot 10^{-5}\}$} and constant fraction of significant features $\psi = 0.2$. Additionally, we use the permutation
invariant version of the popular \texttt{MNIST} benchmark~\citep{lecun1998gradient}. While it is a relatively dense
(density $\approx 19\%$) and low-dimensional dataset, it is frequently used to evaluate neural networks and has
well-established reference results.

\paragraph{Dimensionality reduction methods for training DNNs.}

First, we explore the prospects of training \glspl{DNN} on sparse, unstructured, high-dimensional data whose
dimensionality has been reduced with techniques discussed in Section~\ref{sec:background_fs_sparse_high} and
Section~\ref{sec:background_fe_sparse_high}. Specifically, we evaluate three feature selection methods: \gls{IG},
Chi-square and F-score and \gls{PCA}-based feature extraction. We compare these methods to \acrlong{RP}. Similarly to
previous experiments, we test five \gls{RP} constructions, i.e., Gaussian, Achlioptas', Li's, \gls{SRHT} and Count
Sketch.

To compare the performance of deep networks trained on data with reduced dimensionality we first generated the
low-dimensional representations of the datasets using all evaluated methods. Similarly to the previous experiments, we
reduced the dimensionality of the synthetic datasets from one million to $1000$. For \texttt{MNIST} we reduced the
dimensionality from $784$ to $400$. For the experiments on the synthetic datasets, we used the same network architecture
and training regime as described in the previous sections. We followed a similar procedure for the \texttt{MNIST}
experiments. However, we used hidden layers with just $1000$ units each, and, after pretraining, we added a $10$-way
softmax layer instead of a sigmoid unit. This is the same network architecture as reported
in~\citep{sutskever2013importance}, except it has a smaller input layer ($400$ units instead of $784$). We finetuned
this network to minimize multinomial \gls{CE} cost.

\setlength{\tabcolsep}{8pt}
\begin{table*}[htb]
  \caption{Early stopping errors (\%) for different dimensionality reduction methods. For each dataset we highlight the
  best performing dimensionality reduction technique.}
  \label{tab:dr_accuracy_comparison}
  \centering
  \begin{tabular}{ccrrrrc} \toprule
    \begin{tabular}{@{}c@{}} Dimensionality \\ reduction method \end{tabular} & \phantom{abc} &
      \multicolumn{5}{c}{Dataset} \\ \cmidrule{3-7}
    && \multicolumn{3}{c}{\texttt{synthetic}, $\psi = 0.2$} & \phantom{abc} & \texttt{MNIST} \\ \cmidrule{3-5}
    && $\rho = 3 \cdot 10^{-6}$ & $\rho = 10^{-5}$ & $\rho = 3 \cdot 10^{-5}$ && \\ \midrule
    IG              && $50.04$ & $49.67$ & $50.05$ && $0.98$ \\
    F-score         && $50.00$ & $49.63$ & $48.82$ && $1.04$ \\
    Chi2            && $49.98$ & $49.61$ & $48.86$ && $1.03$ \\
    PCA             && $45.24$ & $22.80$ & $\mathbf{4.28}$  && $2.59$ \\ \midrule
    Gaussian RP     && $35.66$ & $\mathbf{21.10}$ & $7.94$  && $1.06$ \\
    Achlioptas' RP  && $35.64$ & $21.49$ & $7.62$  && $\mathbf{0.94}$ \\
    SRHT RP         && $35.64$ & $21.42$ & $7.84$  && $1.04$ \\
    Li's RP         && $39.47$ & $29.42$ & $15.37$ && $1.11$ \\
    Count Sketch RP && $\mathbf{35.14}$ & $22.19$ & $7.59$  && $1.34$ \\
    \bottomrule
  \end{tabular}
\end{table*}

\begin{figure}[!htb]
\begin{tabular}{ccc}
  \includegraphics[width=0.3\textwidth]{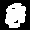} &
  \includegraphics[width=0.3\textwidth]{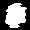} &
  \includegraphics[width=0.3\textwidth]{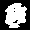} \\
  IG & F-score & Chi-square \\
\end{tabular}
\caption{Visualization of features that were chosen by different feature selection methods on the \texttt{MNIST}
dataset. Selected and discarded features are represented by white and black pixels, respectively.}
\label{fig:fs_mnist}
\end{figure}

In Table~\ref{tab:dr_accuracy_comparison} we report early stopping errors for all evaluated datasets and dimensionality
reduction methods. All feature selection methods, i.e., \gls{IG}, F-score and Chi-square performed very similarly on the
test datasets. On dense \texttt{MNIST}, they yielded results only slightly worse than networks trained on the original
$784$-dimensional data. \citet{srivastava2014dropout} achieved a test error of $0.92\%$ by employing a training regime
similar to ours, i.e., pretraining with \glspl{DBN} followed by finetuning with dropout. Intuitively, removing almost
half of the input dimensions (we discard $384$ features from the original $784$-dimensional data) should significantly
impede the network classification performance. In fact, the task of selecting meaningful features is quite simple for
the \texttt{MNIST} dataset -- the most informative features are located in the central area of the $28$-by-$28$-pixel
images. A large portion of features corresponding to pixels near the image edges is equal to zero for all train and test
examples. Therefore, all evaluated feature selection techniques had no difficulty identifying the most informative
features. In Fig.~\ref{fig:fs_mnist} we present a visualization of the $400$ features selected with \gls{IG}, F-score
and Chi-square.
\setlength{\tabcolsep}{7pt}

Evaluated feature selection methods did not perform well on the synthetic datasets. In fact, networks trained on
features chosen by these techniques performed not much better than a random binary classifier. The main reason for this
poor performance is the sparsity of the synthetic datasets. Let us consider the densest evaluated variant, i.e., $\rho =
3 \cdot 10^{-5}$. It has approximately $3 \cdot 10^{-5} \cdot 10^6 \cdot 10^6 = 3 \cdot 10^7$ non-zero values in the
training set. The reduced dataset is built from $1000$ selected features, and therefore contains on average
$\frac{10^3}{10^6} \cdot 3 \cdot 10^7 = 3 \cdot 10^4$ non-zero elements. This results in a situation where at least
$97\%$ of the training examples contains only zeros. More importantly, even by selecting only the significant features,
the feature selection methods inevitably discard $199,000 / 200,000 = 99.5\%$ of all significant features. This puts
them at a substantial disadvantage, compared to feature extraction techniques, which are able to combine information
from a bigger number of the original features.

On dense \texttt{MNIST}, all \gls{RP} techniques yielded comparable performance to the feature selection methods. The
only exceptions were the Count Sketch and Li's \glspl{RP}, which achieved slightly worse results. Their poorer
performance can be attributed to the low density of the projection matrices: $\frac{1}{\sqrt{784}} \approx 3.57\%$ for
Li's matrix and $\frac{1}{400} = 0.25\%$ for Count Sketch matrix. Projecting the data by multiplying the data matrix by
such sparse matrices can result in many informative features from the original space being lost in the transformed
feature space. \Gls{PCA} performed worse than the evaluated \gls{RP} schemes.

On sparse synthetic datasets, feature extraction methods outperformed feature selection techniques. In particular,
\gls{RP} methods achieved the best results on the two most sparse variants, i.e., with $\rho = 3 \cdot 10^{-6}$ and
$\rho = 10^{-5}$. While Gaussian, Achlioptas', \gls{SRHT} and Count Sketch \gls{RP} yielded comparable results, Li's
construction performed significantly worse on all synthetic dataset variants. The Count Sketch-based \gls{RP} yielded
the best results out of the evaluated \gls{RP} schemes on all synthetic datasets beating dense projections, such as
Gaussian or \gls{SRHT}. \Acrlong{PCA}'s performance strongly depended on the sparsity of the original dataset: for the
sparsest variant ($\rho = 3 \cdot 10^{-6}$) it yielded classification error nearly $10\%$ worse than most of the
\acrlong{RP} schemes, and for the densest variant with $\rho = 3 \cdot 10^{-5}$ it outperformed other competing
dimensionality reduction techniques.

\paragraph{Training linear models on unprojected feature space.}

As a reference to \glspl{DNN} trained on lower-dimensional data, we conducted experiments with linear classifiers
trained on the original data. Specifically, we evaluated \gls{LR} and \glspl{SVM}, implemented in the LIBLINEAR
package~\citep{fan2008liblinear}.

\gls{LR} and \gls{SVM} models were trained on the original feature space, with each feature normalized to $[-1, 1]$
(synthetic datasets) or $[0, 1]$ (\texttt{MNIST}) as recommended by \citet{hsu2003practical}. Following
\citet{yuan2010comparison}, we used solvers with L2 regularization, namely L2-regularized logistic regression and
L2-regularized L2-loss \acrlong{SVM}. For each solver and dataset, we carried out cross-validation experiments to find
the best value of the hyperparameter $C$\footnote{$C$ is the penalty parameter of the error term.}. We report \gls{LR}
and \gls{SVM} test errors in Table~\ref{tab:linear_accuracy_comparison}. For a comparison with the previous approach, we
also include the test errors of \glspl{DNN} trained on feature space reduced with Gaussian \gls{RP}.
\setlength{\tabcolsep}{7pt}
\begin{table*}[htb]
  \caption{Test errors (\%) for linear classifiers trained on the unprojected data. For a comparison, we also report the
  test errors for deep networks trained on data projected with Gaussian random projection.}
    \label{tab:linear_accuracy_comparison}
  \centering
  \begin{tabular}{ccrrrrc} \toprule
    Classifier & \phantom{abc} & \multicolumn{5}{c}{Dataset} \\ \cmidrule{3-7}
    && \multicolumn{3}{c}{\texttt{synthetic}, $\psi = 0.2$} & \phantom{abc} & \texttt{MNIST} \\ \cmidrule{3-5}
    && $\rho = 3 \cdot 10^{-6}$ & $\rho = 10^{-5}$ & $\rho = 3 \cdot 10^{-5}$ && \\ \midrule
    LR                && $44.90$ & $33.73$ & $25.72$ && $8.10$ \\
    SVM               && $44.90$ & $33.73$ & $25.72$ && $8.39$ \\ \midrule
    Gaussian RP + DNN && $35.66$ & $21.10$ & $7.94$  && $1.06$ \\
    \bottomrule
  \end{tabular}
\end{table*}

In our experiments, \gls{LR} and \gls{SVM} models yielded almost identical performance. The test error on the
\texttt{MNIST} dataset was similar to the $12.0\%$ error achieved by a linear classifier (1-layer neural network)
reported in~\citep{lecun1998gradient}. However, on all evaluated datasets, the accuracy of the linear models was
significantly lower than the accuracy of \glspl{DNN} with \gls{RP} layer. We believe that this is a result of model
underfitting that stems from training a linear model on highly nonlinear data.

\paragraph{Computational cost comparison.}

The time $T_\mathrm{total}$ required to train a neural network on a high-dimensional dataset is the sum of the
dimensionality reduction time $T_\mathrm{DR}$ and the time needed to train the network on the reduced feature space
$T_\mathrm{train}$. Here we are interested in comparing the computational efficiency of the evaluated dimensionality
reduction techniques, and thus we mostly focus on the dimensionality reduction time $T_\mathrm{DR}$ and the relation of
this time to the total training time $T_\mathrm{total}$.

To compare the dimensionality reduction execution times $T_\mathrm{DR}$, we ran each procedure five times on every
dataset. The averages of these times are presented in Table~\ref{tab:dr_time_comparison}. In our measurements of
$T_\mathrm{DR}$, we did not include the data loading time nor the time required to normalize the reduced features. The
total training time $T_\mathrm{total}$ includes the time of loading the data, reducing its dimensionality, data
normalization and training the network for $e$ epochs, where $e$ is the epoch with the best error on the validation set.
In Table~\ref{tab:dr_time_comparison}, we also report the percentage of the total training time spent for reducing the
data dimensionality $T_\mathrm{DR} / T_\mathrm{total}$.
\setlength{\tabcolsep}{3pt}
\begin{table*}[htb]
  \caption{Average time $T_{DR}$ of performing dimensionality reduction. In parenthesis we report $T_\mathrm{DR} /
  T_\mathrm{total} = T_\mathrm{DR} / (T_\mathrm{DR} + T_\mathrm{train})$ -- the fraction of time required to reduce the
  data dimensionality over the total time of training the network.}
  \label{tab:dr_time_comparison}
  \centering
  \begin{tabular}{crrrc} \toprule
    \begin{tabular}{@{}c@{}} Dimensionality \\ reduction \\ method \end{tabular} &
      \multicolumn{4}{c}{Dataset} \\ \cmidrule{2-5}
    & \multicolumn{3}{c}{\texttt{synthetic}, $\psi = 0.2$} & \texttt{MNIST} \\ \cmidrule{2-4}
    & $\rho = 3 \cdot 10^{-6}$ & $\rho = 10^{-5}$ & $\rho = 3 \cdot 10^{-5}$ & \\ \midrule
    IG              & $1.0$s~~~($0.1$\%)   & $1.3$s~~~($0.1$\%)   & $2.1$s~~~($0.1$\%)   & ~~~~$0.4$s~~~($0.3$\%) \\
    F-score         & $3.6$s~~~($0.1$\%)   & $3.8$s~~~($0.1$\%)   & $4.7$s~~~($0.1$\%)   & ~~~~$0.4$s~~~($0.2$\%) \\
    Chi2            & $1.0$s~~~($0.1$\%)   & $1.4$s~~~($0.1$\%)   & $2.2$s~~~($0.1$\%)   & ~~~~$0.4$s~~~($0.4$\%) \\
    PCA             & $3912.4$s~(3$3.1$\%) & $3865.4$s~($26.4$\%) & $4013.6$s~(3$9.8$\%) & $150.1$s~($35.6$\%) \\
    \midrule
    Gaussian RP     & $57.3$s~~~($0.5$\%)  & $63.3$s~~~($1.6$\%)  & $75.2$s~~~($2.7$\%)  & ~~~~$0.6$s~~~($0.3$\%) \\
    Achlioptas' RP  & $33.7$s~~~($1.2$\%)  & $40.5$s~~~($1.0$\%)  & $49.7$s~~~($1.3$\%)  & ~~~~$0.6$s~~~($0.4$\%) \\
    SRHT RP         & $112.1$s~~~($3.9$\%) & $109.9$s~~~($3.4$\%) & $128.5$s~~~($3.9$\%) & ~~~~$1.3$s~~~($0.4$\%) \\
    Li's RP         & $3.9$s~~~($0.2$\%)   & $4.4$s~~~($0.1$\%)   & $5.6$s~~~($0.2$\%)   & ~~~~$1.0$s~~~($0.6$\%) \\
    Count Sketch RP & $1.9$s~~~($0.1$\%)   & $2.5$s~~~($0.1$\%)   & $3.9$s~~~($0.2$\%)   & ~~~~$0.3$s~~~($0.1$\%) \\
    \bottomrule
  \end{tabular}
\end{table*}

For most of the evaluated dimensionality reduction methods, their execution time $T_\mathrm{DR}$ was much lower than the
total training time $T_\mathrm{total}$. This is especially true for sparse \gls{RP} methods and fast feature selection
techniques that, even for the densest synthetic dataset variants could be computed in less than several seconds.
\Acrlong{RP} schemes that employ dense projection matrices were approximately ten times slower compared to sparse
\gls{RP} schemes. However, dense \gls{RP} methods are still a viable option -- performing dimensionality reduction with
these schemes takes little time in relation to $T_\mathrm{total}$. \Acrlong{PCA} proved to be the least efficient method
in our evaluation. Dimensionality reduction with \gls{PCA} was almost $40$ times more expensive than dimensionality
reduction with the slowest dense \gls{RP} schemes -- depending on the dataset calculating \gls{PCA} took up to $40\%$ of
$T_\mathrm{total}$.

Note that the relative cost of performing the dimensionality reduction $T_\mathrm{DR} / T_\mathrm{total}$ can vary
significantly depending on the difficulty of the dataset. Our synthetic datasets are easier to learn than some
real-world datasets, and thus require shorter training (i.e., a smaller number of epochs). For harder datasets, the
networks may converge much slower, making the relative cost $T_\mathrm{DR} / T_\mathrm{total}$ less significant.

Finally, we compare the time of training \glspl{DNN} on reduced feature space with the time required to train linear
models on the original data. In Table~\ref{tab:linear_time_comparison}, we report the average times of training linear
classifiers. Due to the model simplicity and much lower number of learnable parameters, linear classifiers can be
trained several orders of magnitude faster than \glspl{DNN}.
\setlength{\tabcolsep}{9pt}
\begin{table*}[htb]
  \caption{Average time of training linear classifiers on the original data. For a comparison, we also report training
  times for deep networks trained on data projected with Gaussian random projection.}
    \label{tab:linear_time_comparison}
  \centering
  \begin{tabular}{crrrrc} \toprule
    Classifier & \multicolumn{5}{c}{Dataset} \\ \cmidrule{2-6}
    & \multicolumn{3}{c}{\texttt{synthetic}, $\psi = 0.2$} & \phantom{a} & \texttt{MNIST} \\ \cmidrule{2-4}
    & $\rho = 3 \cdot 10^{-6}$ & $\rho = 10^{-5}$ & $\rho = 3 \cdot 10^{-5}$ && \\ \midrule
    LR  & $2.3$s & $4.2$s & $10.1$s && ~~~~$23.9$s \\
    SVM & $2.0$s & $4.0$s & $10.0$s && ~~~~~~$9.8$s \\
    \midrule
    Gaussian RP + DNN & $\sim 12000.0$s & $\sim 4000.0$s & $\sim 2000.0$s && $\sim200.0$s \\
    \bottomrule
  \end{tabular}
\end{table*}

\paragraph{Conclusion.}

Our experiments show that the best approach to learning models on sparse, high-dimensional data is to train \glspl{DNN}
on data whose dimensionality has been reduced with \acrlong{RP}. Specifically, in our evaluation, networks with Count
Sketch, Gaussian, \gls{SRHT} and Achlioptas' \gls{RP} layers achieved the highest accuracy. Gaussian, \gls{SRHT} and
particularly Count Sketch \gls{RP} combined the best network performance with reasonable execution time. Replacing
\gls{RP} with feature selection techniques did not improve the network performance. Training \glspl{DNN} on data reduced
with \gls{PCA} yielded good results only on one relatively dense dataset variant. Unfortunately, computing \gls{PCA} for
large-scale datasets was computationally much more expensive than performing \acrlong{RP}. Linear classifiers trained on
the original data were computationally very efficient but produced models that performed poorly on nonlinear tasks.

\subsection{Experiments on large-scale real-world data}
\label{sec:rp_real_world}

To demonstrate the practical effectiveness of networks with \gls{RP} input we performed experiments on four real-world
classification datasets: \texttt{webspam}, \texttt{url}, \texttt{KDD2010-a} and \texttt{KDD2010-b}. The datasets are
described in Appendix~\ref{cha:datasets}. All four datasets are large binary classification tasks. For all datasets, we
randomly projected the data to~$1000$ dimensions and employed a network with two hidden layers, each one with~$3000$
neurons. Each network was pretrained and then finetuned to minimize the binary \gls{CE} cost.

\paragraph{Baseline algorithms.}

The state-of-the-art results in the classification of high-dimensional, unstructured, sparse data are currently achieved
with linear classifiers. In particular, the need to learn high-dimensional data with millions of examples has recently
led to a surge of interest in fast linear classifiers, such as support vector machines. Many algorithms have been
proposed to speed up the \gls{SVM} training. One of the first methods that could efficiently handle large-scale
optimization problems in \glspl{SVM} were LIBLINEAR and Pegasos. LIBLINEAR~\citep{fan2008liblinear} implements dual
coordinate descent method and focuses mainly on solving the dual problem, while Pegasos~\citep{shalev2011pegasos} uses
\gls{SGD} to solve the primal problem.

One line of research on large-scale \gls{SVM} classification focuses on the so-called \textit{out-of-core learning},
i.e., being able to work with data that does not fit in the operating memory. For example, \citet{yu2012large}, authors
of the LIBLINEAR library propose the \gls{BM} algorithm -- a simple approach that involves splitting the data into
smaller blocks, compressing and storing them on disk and then sequentially loading and processing each block. However,
despite being faster than LIBLINEAR, \gls{BM} does not produce models that perform significantly better.
\citet{chang2011selective} propose a modification to the \gls{BM} method, in which the set of informative examples from
the previous blocks persists in memory when a subsequent block is loaded. This approach, called \gls{SBM} outperforms
\gls{BM} both in speed and classification performance. \Acrlong{VW}~\citep{langford2007vowpal} is another out-of-core
learning implementation that uses a similar compression strategy to~\citep{yu2012large}. Unlike \gls{BM}, it employs a
variant of online gradient descent as the learning algorithm. Another line of research on out-of-core methods focuses on
parallelizing the training process. \citet{zhang2012efficient} employ the \gls{ADMM} for training linear classifiers in
a distributed environment. They use the L2-regularized L2-loss (squared hinge loss) \gls{SVM} as the classification
model. However, as pointed out by \citet{yuan2012recent}, their approach may suffer from slow convergence and the
difficulty of choosing the hyperparameters. Other works, in addition to ensuring good parallelization properties of the
developed algorithms, concentrate on encouraging the sparsity of trained models. For example, \citet{yuan2012scalable}
propose \gls{DADM} for L1-regularized L1-/L2-loss \glspl{SVM}. \Acrlong{DADM} can train sparse linear models and offers
competitive prediction performance. Based on the GLMNET method proposed by \citet{friedman2010regularization},
\citet{yuan2012improved} introduce a computationally efficient method called newGLMNET that solves the primal
L1-regularized L2-loss SVM problem.

For more complex datasets, where examples of different classes cannot be separated with a linear decision boundary,
nonlinear kernel \glspl{SVM} offer superior performance. Unfortunately, standard kernelized \glspl{SVM} do not scale
well to larger datasets and are computationally prohibitive for datasets with millions of examples and features.
Recently, the gap between linear and nonlinear \glspl{SVM} has been narrowed by novel approaches that employ additive
kernels. Specifically, \citet{wu2012power} proposed the \gls{PmSVM} algorithm that employs a polynomial approximation of
the gradient to speed up the training with coordinate descent. \citet{yang2012practical} further improved the running
time of \gls{PmSVM} by introducing look-up tables. This modification, called \gls{PmSVM-LUT}, although several-fold
slower than the state-of-the-art linear \gls{SVM} solvers improved the classification performance on several large-scale
datasets. Another method that attempts to bridge the scalability gap between linear and kernel \glspl{SVM} is the
\gls{AMM}. Introduced in~\citep{wang2011trading}, \gls{AMM} uses online learning with stochastic gradient descent to
solve a modified \gls{SVM} problem.

\paragraph{Results.}

In Table~\ref{tab:real_world_rp} we compare the performance of \glspl{DNN} with \gls{RP} layer with the performance of
the baseline algorithms. Error rates of the baseline algorithms were taken from: \citep{yang2012practical} for
LIBLINEAR, \citep{wang2011trading} for Pegasos, \citep{yuan2012scalable} and \gls{VW}
documentation\footnote{\url{https://github.com/JohnLangford/vowpal_wabbit/wiki/Malicious-URL-example}} for
\gls{VW}, \citep{chang2011selective} for \gls{SBM}, \citep{yang2012practical} for \gls{PmSVM-LUT},
\citep{yuan2012scalable} for \gls{DADM}, \citep{yuan2012improved} for newGLMNET, \citep{zhang2012efficient} for
\gls{ADMM} and \citep{wang2011trading} for \gls{AMM}.
\setlength{\tabcolsep}{10pt}
\begin{table*}[htb]
  \caption{Classification errors (\%) on large-scale real-world datasets. For each dataset we highlight the
  result of the best performing method.}
  \label{tab:real_world_rp}
  \centering
  \begin{tabular}{rccccc} \toprule
    Method & \phantom{a} & \multicolumn{4}{c}{Dataset} \\ \cmidrule{3-6}
    && \texttt{webspam} & \texttt{url} & \texttt{KDD2010-a} & \texttt{KDD2010-b} \\ \midrule
    Gaussian RP     && $0.38$ & $1.03$ & $10.86$ & $10.51$ \\
    Achlioptas' RP  && $0.40$ & $1.12$ & $10.88$ & $10.49$ \\
    Li's RP         && $0.36$ & $3.75$ & $11.95$ & $10.98$ \\
    SRHT RP         && $0.40$ & $1.01$ & $10.86$ & $10.49$ \\
    Count Sketch RP && $\mathbf{0.32}$ & $\mathbf{0.96}$ & $11.49$ & $10.54$ \\ \midrule
    LIBLINEAR       && $7.31$ & $1.55$ & $11.44$ & $11.06$ \\
    Pegasos         && $7.28$ & $1.50$ & -     & -     \\
    SBM             && $0.45$ & -      & -     & $10.33$ \\
    VW              && $1.52$ & $1.64$ & -     & $11.09$ \\
    ADMM            && $0.42$ & -      & -     & $10.01$ \\
    DADM            && $0.40$ & -      & -     & $10.43$ \\
    newGLMNET       && $0.36$ & -      & -     & $13.40$ \\
    PmSVM-LUT       && $5.72$ & $1.23$ & $\mathbf{10.39}$ & $\mathbf{9.99}$  \\
    AMM             && $4.50$ & $1.34$ & -     & -     \\
    \bottomrule
  \end{tabular}
\end{table*}

Gaussian, Achlioptas', \gls{SRHT} and Count Sketch projections performed similarly well on the real-world datasets,
while Li's method generally performed worse. This agrees with the results from the previous tests on synthetic datasets.
Overall, networks with randomly projected input significantly improved the current state-of-the-art results on the
\texttt{url} dataset and achieved competitive performance on \texttt{webspam} and \texttt{KDD2010} datasets.
Specifically, on \texttt{webspam}, \glspl{DNN} with Count Sketch \gls{RP} layer slightly outperformed the
state-of-the-art results obtained by \gls{DADM} and newGLMNET. The nonlinear \gls{PmSVM-LUT} achieved better results on
the \texttt{KDD2010} datasets but was significantly outperformed by \glspl{DNN} with \gls{RP} layer on the two denser
benchmarks.

\subsection{Experiments on bag-of-words data}
\label{sec:rp_bow}

In addition to the experiments described in the previous sections, we also conducted experiments with networks trained
on randomly projected \gls{BOW} data. Note that some of the datasets evaluated in the previous section also contain
\gls{BOW} features. For instance, in the \texttt{url} dataset the lexical features that constitute over half of the
$3.2$ million features are \gls{BOW} representations of tokenized \gls{URL} addresses~\citep{ma2009identifying}.
However, in this section, we focus on training \glspl{DNN} on purely \gls{BOW} representation. For the evaluation we
employ deep autoencoder networks trained on two text datasets: \mbox{20-newsgroups} (\texttt{TNG}) and \texttt{RCV1}
(for details on the dataset construction see Appendix~\ref{cha:datasets}). Unlike previous experiments, in which all
models were trained for the classification task, here we train the networks for document retrieval.

When training autoencoders, text data is often represented by \gls{BOW} vectors over a dictionary of $D$ most frequent
words in the training set. The dictionary size $D$ is commonly on the order of a few thousand
words~\citep{salakhutdinov2009learning,salakhutdinov2009semantic}. Our aim, therefore, is to employ \gls{RP} to train
networks on much larger dictionaries. By extending the vocabulary with rarely occurring words we hope to allow the
network to learn a more informative representation of the original text data.

\paragraph{Reference networks.}

As a baseline we use deep autoencoders trained on $2000$-word \gls{BOW} representations of \texttt{TNG} and
\texttt{RCV1}. We employ network architectures similar
to~\citep{grzegorczyk2016encouraging,salakhutdinov2009semantic,salakhutdinov2009learning}. Specifically, for
\texttt{TNG} we started by pretraining \mbox{2000-500-250-125-32} \glspl{DBN} with binary units in the hidden layers and
Gaussian units in the output layer. For \texttt{RCV1} we used a similar network with \mbox{2000-500-500-128}
architecture. For the input layer in both networks we used the constrained Poisson
model~\citep{salakhutdinov2009semantic}, i.e., an \gls{RBM} variant suitable for modeling word count data.
\Acrlongpl{DBN} were pretrained with \gls{CD}\textsubscript{1} and used to initialize deep autoencoders. The
autoencoders were then finetuned with error backpropagation to minimize the \gls{CE} cost. The hyperparameters were
selected with experiments on the validation sets.

We use the $32$-dimensional (for \texttt{TNG}) and $128$-dimensional (for \texttt{RCV1}) codes inferred with the trained
autoencoders for the document retrieval task. Following \citet{salakhutdinov2009semantic}, we perform document retrieval
by choosing a query from the test set and ranking the remaining test documents according to the cosine similarity
between their codes and the code of the query document. We do this for every document from the test set. Then, after
averaging the results, we calculate the precision-recall curve. To determine the relevance of a retrieved document to a
given query document we either directly compare their class labels (\texttt{TNG}) or use the fraction of equal labels
(\texttt{RCV1}).

\paragraph{Deep autoencoders with random projection layer.}

To evaluate the performance of deep autoencoders with \gls{RP} layer we first generated several \gls{BOW}
representations of \texttt{TNG} and \texttt{RCV1} using larger dictionary sizes: $D \in \{4000, 6000, 8000, 10000\}$. We
then randomly projected these representations to $2000$ dimensions and normalized each feature to zero mean and unit
variance. However, we did not use the raw \gls{BOW} data, i.e., word-counts for the words from the dictionary. Instead,
we first converted the \gls{BOW} representation to \gls{TF-IDF} representation. Initially, we also experimented with
training autoencoders with an \gls{RP} layer on raw \gls{BOW} data. However, using \gls{TF-IDF} values yielded
significantly better-performing networks.

\gls{TF-IDF} is one of the most widely used measure of word importance in information
retrieval~\citep{salton1991developments,salton1988term}. Suppose we have a corpus $\mathbb{C} = \{d_1, \ldots, d_N\}$ of
$N$ documents. We denote $f(t, d)$ as the number of occurrences of term $t$ in document $d$. The \gls{TF-IDF} score is
then defined for term $t$ in document $d$ as a product of two statistics, term frequency $\mathrm{TF}(t, d)$ and inverse
document frequency $\mathrm{IDF}(t; \mathbb{C})$:
\begin{equation}
\mathrm{TF}\text{-}\mathrm{IDF}(t, d; \mathbb{C}) = \mathrm{TF}(t, d) \cdot \mathrm{IDF}(t; \mathbb{C}).
\end{equation}
The term frequency can be defined in various ways, with the simplest: \mbox{$\mathrm{TF}(t, d) = f(t, d)$}. Here,
however, we use logarithmically scaled word frequency:
\begin{equation}
\mathrm{TF}(t, d) = 1 + \log(1 + f(t, d)),
\end{equation}
as in our experiments it produced the best performing networks. The inverse document frequency of a term measures its
rarity in the corpus $\mathbb{C}$:
\begin{equation}
\mathrm{IDF}(t; \mathbb{C}) = \log\left(\frac{N}{\lvert\{d \in \mathbb{C}: f(t, d) > 0\}\rvert}\right).
\end{equation}

We trained deep autoencoder networks on the projected \gls{TF-IDF} data using a network architecture and training
settings similar to those in the baseline network. However, we employed Gaussian units instead of constrained Poisson
units in the first layer. For the hidden layers, instead of binary units, we used rectifier linear units since they
yielded significantly better performing models. We used similar training regime as in the reference network, that is, we
pretrained the \glspl{DBN} with \gls{CD}\textsubscript{1} and unfolded them to initialize the autoencoders. The
autoencoders were then finetuned with error backpropagation to minimize the mean square error cost. For each dataset
representation, we selected the learning hyperparameters with experiments on the validation sets. We used the \gls{AUC}
to compare the performance of the trained models.

\paragraph{Results.}

\setlength{\tabcolsep}{5pt}
\begin{table*}[htb]
  \caption{Area under the precision-recall curve for deep autoencoders with RP layer trained on different \texttt{TNG}
  representations. The baseline network trained on \mbox{$2000$-dimensional} bag-of-words data achieves AUC of $0.373$.}
  \label{tab:tng_rp_auc}
  \centering
  \begin{tabular}{rccccc} \toprule
    RP layer type & \phantom{a} & \multicolumn{4}{c}{Dictionary size $D$} \\ \cmidrule{3-6}
                 && $4000$ & $6000$ & $8000$ & $10,000$ \\ \midrule
    Gaussian     && 0.341 & \textbf{0.366} & 0.336 & 0.333 \\
    Achlioptas'  && 0.321 & \textbf{0.348} & 0.326 & 0.337 \\
    Li's         && 0.307 & \textbf{0.356} & 0.321 & 0.350 \\
    SRHT         && 0.344 & 0.326 & \textbf{0.352} & 0.327 \\
    Count Sketch && \textbf{0.327} & 0.314 & 0.311 & 0.313 \\
    \bottomrule
  \end{tabular}
\end{table*}

\setlength{\tabcolsep}{5pt}
\begin{table*}[htb]
  \caption{Area under the precision-recall curve for deep autoencoders with RP layer trained on different \texttt{RCV1}
  representations. The baseline network trained on \mbox{$2000$-dimensional} bag-of-words data achieves AUC of $0.315$.}
  \label{tab:rcv1_rp_auc}
  \centering
  \begin{tabular}{rccccc} \toprule
    RP layer type & \phantom{a} & \multicolumn{4}{c}{Dictionary size $D$} \\ \cmidrule{3-6}
                 && $4000$ & $6000$ & $8000$ & $10,000$ \\ \midrule
    Gaussian     && \textbf{0.306} & 0.300 & 0.296 & 0.295 \\
    Achlioptas'  && \textbf{0.312} & 0.299 & 0.296 & 0.290 \\
    Li's         && \textbf{0.304} & 0.299 & 0.292 & 0.289 \\
    SRHT         && \textbf{0.305} & 0.303 & 0.294 & 0.293 \\
    Count Sketch && 0.254 & 0.260 & \textbf{0.280} & 0.278 \\
    \bottomrule
  \end{tabular}
\end{table*}

\begin{figure}[!htb]
\begin{tabular}{cc}
  \includegraphics[width=0.5\linewidth]{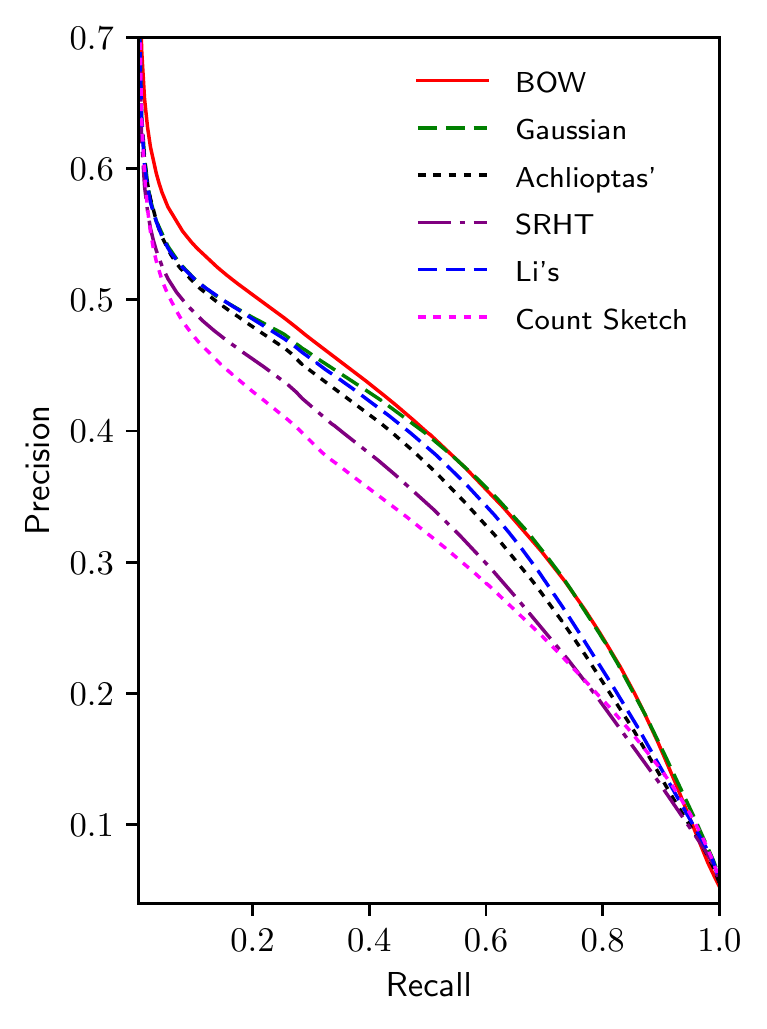} &
  \includegraphics[width=0.5\linewidth]{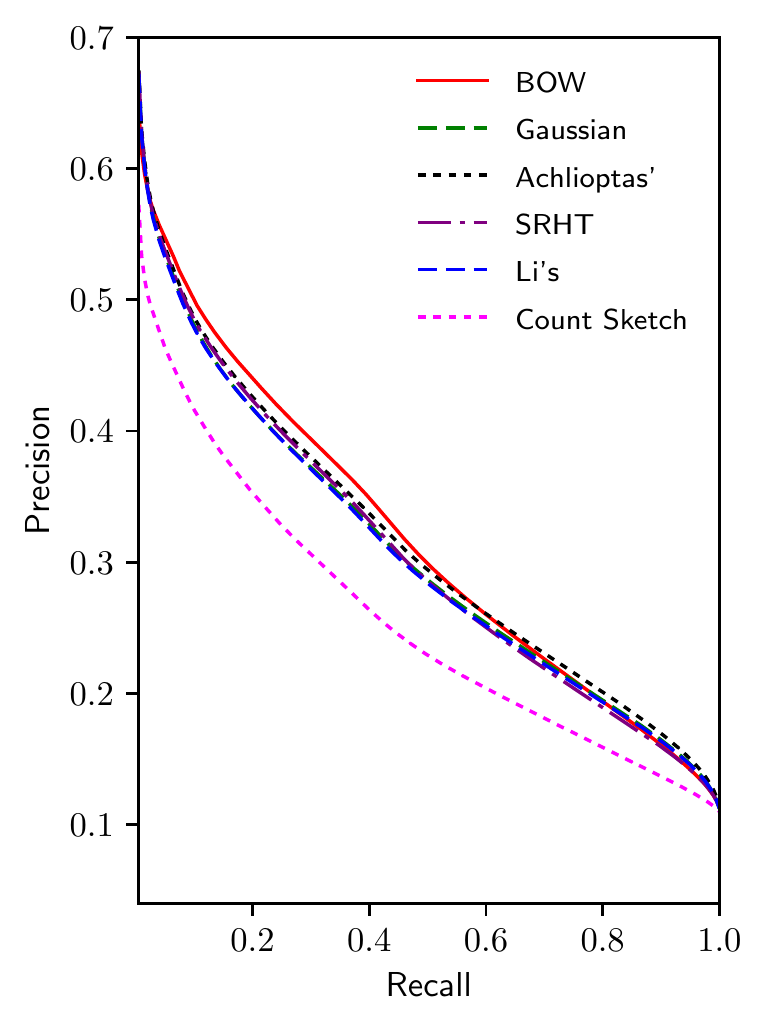} \\
  \texttt{TNG} & \texttt{RCV1} \\
\end{tabular}
\caption{Precision-recall curves for deep autoencoders trained on unprojected BOW data (solid red line) and randomly
  projected BOW representations (dashed lines).}
\label{fig:rp_auc}
\end{figure}

In Table~\ref{tab:tng_rp_auc} and Table~\ref{tab:rcv1_rp_auc} we report the \gls{AUC} values for autoencoders trained on
different dataset representations. The baseline autoencoders trained on unprojected $2000$-dimensional \gls{BOW} data
achieves \gls{AUC} of $0.373$ and $0.315$, for \texttt{TNG} and \texttt{RCV1}, respectively. Even the best performing
autoencoders with \gls{RP} input layer yield results worse than these reference networks. In general, increasing the
dictionary size $D$ does not lead to higher \gls{AUC} of the trained models. For most \gls{RP} schemes, the best
document retrieval performance is achieved with $D=4000$ or $D=6000$. This result suggests that broadening the input
dictionary to include rarely occurring words and then projecting this representation to $2000$ dimensions does not
produce a more informative data representation. In Figure~\ref{fig:rp_auc} we present the precision-recall curves for
autoencoders trained on randomly projected \gls{BOW} representations constructed over $6000$ and $4000$ most common
words, for \texttt{TNG} and \texttt{RCV1}, respectively. The best results are achieved by autoencoders with Gaussian and
Achlioptas' \gls{RP} layers. However, they are significantly worse than the baseline, especially for lower recall
values.

\setlength{\tabcolsep}{0pt}

We believe that the poor performance of networks trained on randomly projected \gls{BOW} data is a consequence of two
facts. First, as the autoencoders were trained on projected, real-valued data, their input units were Gaussian. The
reference networks, on the other hand, used the constrained Poisson model, which is tailored specifically for word count
data. Importantly, employing the constrained Poisson model makes the learning much more stable, by properly dealing with
documents of different lengths~\citep{salakhutdinov2009semantic}. Second, the $2000$-word dictionary used by the
reference autoencoders already captured most of the useful information from the text. This is especially significant for
the smaller \texttt{TNG} dataset and to a lesser degree for \texttt{RCV1}. Extending the vocabulary with rarely
occurring words added little information to the network. In fact, using too large dictionary size resulted in decreased
network performance, as can be seen in Table~\ref{tab:tng_rp_auc} and Table~\ref{tab:rcv1_rp_auc}: for most \gls{RP}
schemes increasing $D$ above $6000$ did not improve the network performance. To further investigate the influence of
enriching the input data representation we also experimented with concatenating the basic $2000$-word \gls{BOW}
representation with bigram and trigram features. While such concatenated representations (reaching more than $10^5$
features) carried more information from the input text, they yielded much worse performing networks. We believe that
adding more features to the input representation did not improve the network performance because in all tests the
\gls{RP} layer projected the data to a fixed number of dimensions ($2000$). We hypothesize that with a fixed
projection dimensionality adding less and less informative features to the input representation causes the projected
features to become increasingly noisy. This is because each feature in the projected space is a linear combination of
all input features, including less informative ones.

\section{Learnable random projection layer}
\label{sec:lrp_layer}

In the previous section, we employed \gls{RP} to enable training deep networks on sparse, high-dimensional, unstructured
datasets. We showed that training a network on randomly projected input can be viewed as prepending the network with a
linear layer, whose weights are initialized to elements of an \gls{RP} matrix. However, in order to reduce the
computational cost of the training, we did not adjust the weights in such fixed-weight \gls{RP} layers. In this section,
we show that finetuning the weights in \gls{RP} layers is feasible in practical applications and can improve the network
performance. This holds under two conditions: (i) the \gls{RP} scheme that is used to initialize the weights in the
\gls{RP} layer is sparse, and (ii) only the non-zero weights are updated. We discuss important network architecture
considerations as well as training regime settings that enable us to efficiently train networks with the finetuned
\gls{RP} layer.

While the idea of finetuning weights in the \gls{RP} layer may seem straightforward, there are several technical
difficulties that make its implementation challenging. They stem primarily from the high computational cost of
performing the weight updates and normalizing the layer outputs. As we discussed in the previous section, for
high-dimensional datasets performing even a single update of a dense \gls{RP} layer is computationally prohibitive.
Fortunately, we can reduce the number of weights to a manageable size by choosing a sparse variant of the \gls{RP}
layer. In this work, we propose to construct finetuned \gls{RP} layers using two sparse \acrlong{RP} schemes discussed
in Chapter~\ref{cha:rp}, i.e., Li's and Count Sketch-based projections. Compared to dense \gls{RP} matrices, Li's and
Count Sketch constructions reduce the total number of weights by a factor of $\sqrt{d}$ and $k$, respectively, where $d$
is the number of input units, and $k$ is the number of output units. This is usually sufficient to make the training
feasible.

\begin{figure}[htb!]
  \centering
  \includegraphics[width=.7\linewidth]{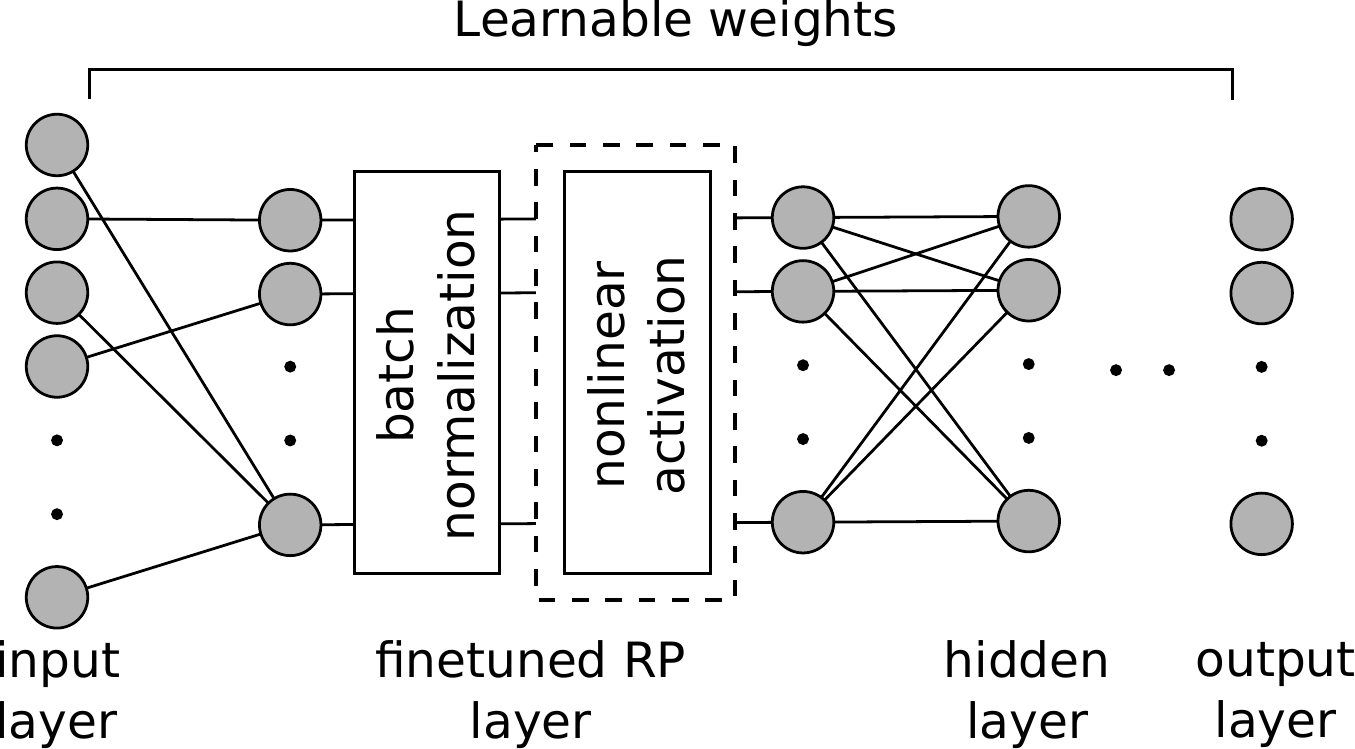}
  \caption{Neural network with finetunedr random projection layer. Weights in the finetuned random projection layer are
  initialized to a sparse RP matrix. Only the weights that are initially non-zero are part of the model. The output of
  the projection is batch normalized and optionally transformed with a nonlinear activation function.}
  \label{fig:lrp_network}
\end{figure}

To ensure that the number of model parameters does not increase during training, we update only these elements in the
finetuned \gls{RP} matrix that are initially non-zero. However, since the number of output units in an \gls{RP} layer is
relatively small, we do learn the biases for these units. This construction can be interpreted as an input layer with
sparse connectivity (Fig.~\ref{fig:lrp_network}). To further improve the training performance we can restrict the weight
updates in the \gls{RP} layer to a fraction of the training mini-batches. We found that even with the sparse \gls{RP}
layers this approximation is necessary for our largest benchmark datasets. Importantly, to reduce the bias introduced by
skipping some of the weight updates, the updates are performed for randomly selected mini-batches. Since for large
datasets \gls{SGD} typically uses incremental gradient~\citep{bertsekas2011incremental}\footnote{In the incremental
gradient method the mini-batches are processed in the same order during every epoch.}, this can be realized by simply
performing the update with a fixed probability for each mini-batch. We denote this probability as $\eta$. Equivalently,
$\eta$ can be interpreted as the expected fraction of mini-batches on which the \gls{RP} layer is finetuned.

When training networks with fixed-weight \gls{RP} layers, we normalize the \gls{RP} layer activations to zero mean and
unit variance using moments calculated on the training set. Since the weights in this \gls{RP} layer variant do not
change during training, this operation can be performed only once before the training. This is no longer true in
networks with finetuned \acrlong{RP} where the \gls{RP} weights change over time. In this case, to ensure proper
normalization of the \gls{RP} layer outputs we propose to either normalize the input data (and thus indirectly the
\gls{RP} layer outputs) or to insert a \acrlong{BN} layer~\citep{ioffe2015batch} between the \acrlong{RP} and the
activation function. Note, however, that in the first approach the feature-wise normalization of the input data is
limited to scaling (e.g., by the reciprocal of the maximum absolute value for each feature), since shifting the mean
would destroy the sparsity of the data. Compared to the one-time normalization in the fixed-weight \gls{RP} layer,
\acrlong{BN} is computationally more expensive, since the normalization has to be performed for each training
mini-batch. Furthermore, \gls{BN} introduces additional $2k$ learnable parameters needed for the scaling and shifting of
the $k$ normalized activations.

We found that networks with finetuned \gls{RP} layer are best trained end-to-end, starting from randomly initialized
weights in the layers succeeding the \gls{RP} layer. Initially, we also considered different training regimes. For
example, we experimented with networks that were first trained without changing the \gls{RP} layer and then finetuned
end-to-end with backpropagation. However, this training regime yielded inferior results.

\subsection{Experiments on large-scale data}
\label{sec:lrp_experiments}

We evaluated the performance of finetuned \gls{RP} layers on several large-scale datasets: a variant of the synthetic
dataset with the density \mbox{$\rho = 10^{-5}$} and the fraction of significant features $\psi = 0.2$, \texttt{webspam}
dataset and \texttt{url} dataset. Additionally, we report results on a toy benchmark -- \texttt{MNIST}.

We begin by evaluating the influence of different normalization schemes on the performance of networks with finetuned
\gls{RP} layers. We follow up by looking into the problem of choosing the optimal fraction of \gls{RP} layer updates
$\eta$. This gives us an insight into the balance between the \gls{RP} layer quality and the network training time.
Finally, we study whether adding a nonlinear activation to the \gls{RP} layer improves the network performance.

\subsubsection{Effects of normalization}

We experimented with two normalization schemes:
\begin{itemize}
  \item scaling each feature in the input dataset by the reciprocal of its maximum absolute value over the training set
  (further called MaxAbs scaling),
  \item \acrlong{BN} of the projected data.
\end{itemize}
In the comparison, we also include results obtained without any normalization of the input data or the \gls{RP} layer
outputs.

We trained \glspl{MLP} with the same network architectures and using the same training settings as in
Section~\ref{sec:nn_with_rp_layer}. Specifically, we employed a \mbox{784-400-1000-1000-10} architecture for
\texttt{MNIST} and a \mbox{d-1000-3000-3000-1} architecture for the large-scale datasets, where $d$ is the dataset
dimensionality. We used linear activation function in the finetuned \gls{RP} layer and \glspl{ReLU} in the subsequent
hidden layers. We trained the networks using mini-batch \gls{SGD} with momentum. To prevent overfitting we used dropout,
and we additionally limited the magnitudes of the weights with L2 cost. During training, we gradually decreased the
learning rate following an exponential decay, while simultaneously increasing the momentum. We chose the hyperparameter
values with experiments on the validation sets. In the experiments on \texttt{MNIST} and synthetic dataset, we performed
the weight update in the \gls{RP} layer for all mini-batches. For \texttt{webspam} and \texttt{url} we used $\eta =
0.5$, which made training our models feasible.

\setlength{\tabcolsep}{8pt}
\begin{table*}[htb]
  \caption{Test errors (\%) for networks with Li's finetuned random projection layer trained using different
  normalization schemes.}
  \label{tab:lrp_normalization_sparse}
  \centering
  \begin{tabular}{ccccc} \toprule
    Network architecture & \multicolumn{4}{c}{Dataset} \\ \cmidrule{2-5}
    & \texttt{MNIST} & \begin{tabular}{@{}c@{}} synthetic \\ $\rho = 10^{-5}$ \\ $\psi = 0.2$ \end{tabular} &
      \texttt{webspam} & \texttt{url} \\ \midrule
    No normalization    & $1.04$ & $\mathbf{26.25}$ & $0.46$ & $3.39$ \\
    MaxAbs scaling      & $\mathbf{0.97}$ & $38.21$ & $0.88$ & $3.41$\\
    Batch Normalization & $1.10$ & $26.55$ & $\mathbf{0.35}$ & $\mathbf{3.30}$ \\ \midrule
    Fixed-weight RP layer & $1.11$ & $29.42$ & $0.36$ & $3.75$ \\
    \bottomrule
  \end{tabular}
\end{table*}

\setlength{\tabcolsep}{8pt}
\begin{table*}[htb]
  \caption{Test errors (\%) for networks with Count Sketch finetuned random projection layer trained using different
  normalization schemes.}
  \label{tab:lrp_normalization_count}
  \centering
  \begin{tabular}{ccccc} \toprule
    Network architecture & \multicolumn{4}{c}{Dataset} \\ \cmidrule{2-5}
    & \texttt{MNIST} & \begin{tabular}{@{}c@{}} synthetic \\ $\rho = 10^{-5}$ \\ $\psi = 0.2$ \end{tabular} &
      \texttt{webspam} & \texttt{url} \\ \midrule
    No normalization    & $\mathbf{0.97}$ & $20.36$ & $0.47$ & $0.87$ \\
    MaxAbs Scaling      & $1.10$ & $40.87$ & $0.99$ & $0.90$ \\
    Batch Normalization & $1.22$ & $\mathbf{20.16}$ & $\mathbf{0.25}$ & $\mathbf{0.75}$ \\ \midrule
    Fixed-weight RP layer & $1.34$ & $22.19$ & $0.32$ & $0.96$ \\
    \bottomrule
  \end{tabular}
\end{table*}

We report the early stopping errors for different normalization schemes in Table~\ref{tab:lrp_normalization_sparse} and
Table~\ref{tab:lrp_normalization_count}. Training networks with the finetuned \gls{RP} layer on unnormalized data
yielded surprisingly good results: compared to the reference networks with fixed-weight \gls{RP} layers, these networks
performed better on all datasets except \texttt{webspam}. Our initial experiments also showed that training networks
with the finetuned \gls{RP} layer on unnormalized data is greatly facilitated by learning the biases. Interestingly,
scaling the features by reciprocal of their maximum absolute values yielded the worst results, especially for
large-scale datasets. For \texttt{webspam} and synthetic dataset, networks trained on data scaled in this way performed
even worse than the reference networks with fixed-weight RP layer. Only networks that used \gls{BN} consistently
outperformed the reference networks. Importantly, in experiments on the real-world datasets, i.e., \texttt{webspam} and
\texttt{url}, networks with \gls{BN} performed best. Therefore, in our subsequent experiments, we focus only on networks
with batch normalized \gls{RP} layer.

\subsubsection{Adjusting the fraction of RP layer updates}

The fraction of mini-batches that are used for learning the weights in the \gls{RP} layer, $\eta$, lets us control the
computational cost of its training. For smaller datasets, where performing the weight update for every mini-batch is
possible, $\eta = 1$ should yield the best performing network. However, for large-scale datasets, training the \gls{RP}
layer on every input mini-batch may be too costly. Therefore, when the available training time is limited, choosing the
$\eta$ value becomes an important training decision. While increasing the $\eta$ value makes the \gls{RP} layer learn
faster, it also increases the computational time.

\begin{figure}[!htb]
  \includegraphics[width=1.0\textwidth]{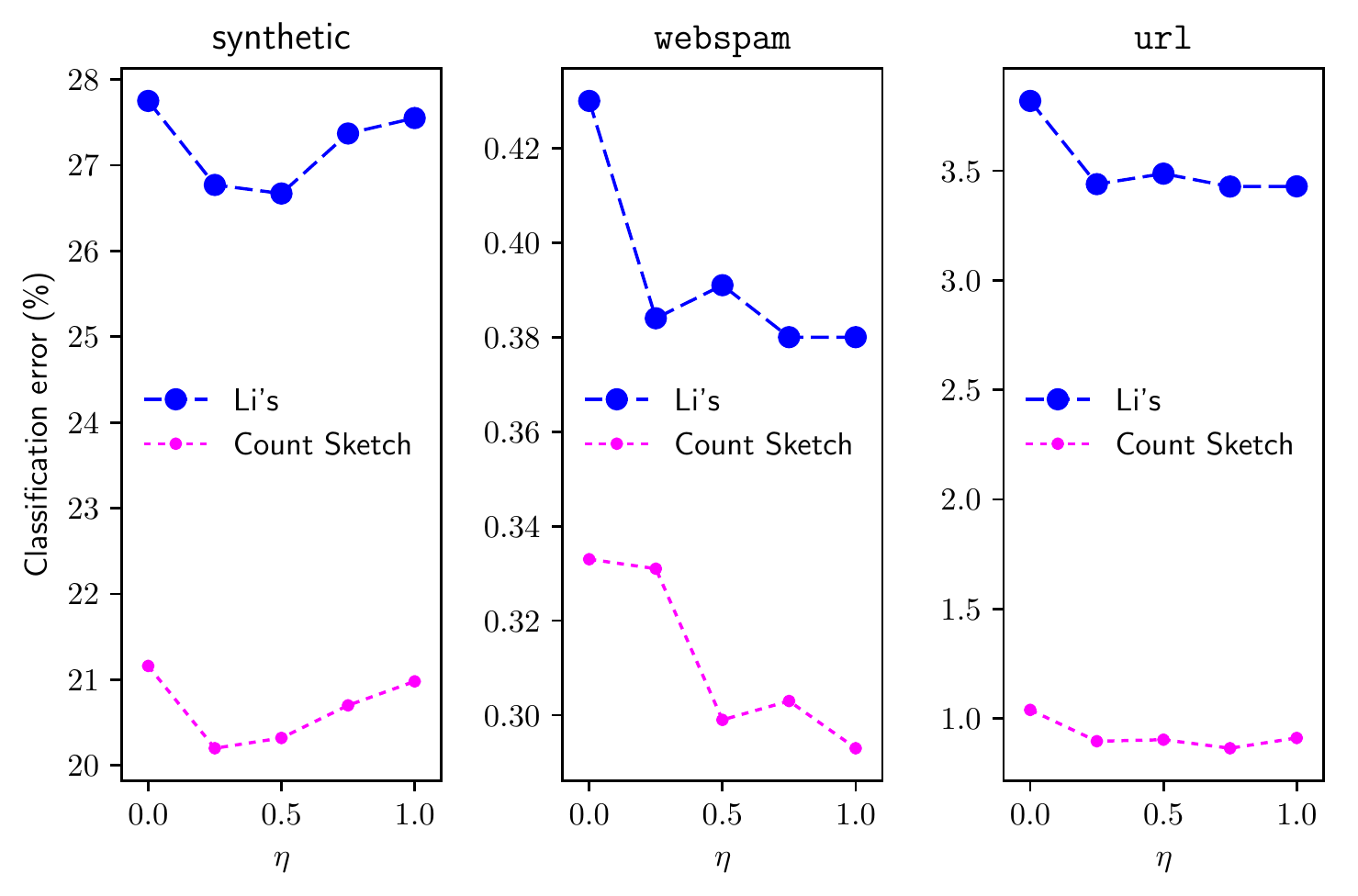}
  \caption{Influence of $\eta$ on the performance of networks with finetuned random projection layer. $\eta$ is the
  fraction of mini-batches used to train the random projection layer.}
  \label{fig:lrp_eta}
\end{figure}

To shed light on the balance between the \gls{RP} layer quality and the training time, we conducted a series of
experiments with varying $\eta$. Importantly, we limited each network training time to 24 hours. Consequently, using
larger $\eta$ values resulted in lower number of training epochs. We experimented on three large-scale datasets:
synthetic dataset with $\rho = 10^{-5}$ and $\psi = 0.2$, \texttt{webspam} and \texttt{url}. We used the same network
architectures and training settings as described in the previous section. In Figure~\ref{fig:lrp_eta} we present how the
performance of networks with finetuned \gls{RP} layer depended on $\eta$. For each dataset and $\eta$ value, we report
the network's early stopping error.

\setlength{\tabcolsep}{0.0em} % for the horizontal padding

The balance between the \gls{RP} layer quality and the training time varied between the datasets. For the synthetic
dataset, the best performing networks were trained with \mbox{$\eta \in \{0.25, 0.5\}$}, while for the real-world
datasets $\eta$ values close to 1 yielded the best results. Importantly, all experiments with small $\eta$ values show
that networks can significantly benefit from finetuning the \gls{RP} layer weights, even if these weights are rarely
updated. Nevertheless, the optimal $\eta$ depends on factors, such as the properties of the dataset, its dimensionality,
the network architecture and the available training time. Therefore, for practical applications we recommend starting
with relatively small $\eta$ and, if the training time is acceptable, test higher $\eta$ values. Assuming that the
number of training epochs is sufficient and that the network is properly regularized, the performance should improve
with increasing $\eta$.

\subsubsection{Activation function in finetuned random projection layer}

So far we only experimented with \gls{RP} layers without an activation function (or, equivalently, with the linear
activation function). In this section, we investigate the prospects of applying the sigmoid, \gls{LReLU} and \gls{ReLU}
nonlinearities after batch normalization of the \gls{RP} layer output. Note that if we disregard the large computational
cost stemming from high data dimensionality $d$ and the additional \gls{BN} layer, our previously used network
architectures could technically be simplified from \mbox{d-1000-3000(ReLU)-3000(ReLU)-1} to
\mbox{d-3000(ReLU)-3000(ReLU)-1}. This is because the first two layers effectively realize two consecutive linear
transformations of the input data and can be replaced with just one linear layer. Therefore, inserting a nonlinear
activation function after the \gls{RP} layer should increase the network ability to represent complex nonlinear data.

\setlength{\tabcolsep}{18pt}
\begin{table*}[htb]
  \caption{Test errors (\%) for networks with Li's finetuned random projection layer using different activation
  functions.}
    \label{tab:lrp_activation_sparse}
  \centering
  \begin{tabular}{ccccc} \toprule
    Activation function & \multicolumn{4}{c}{Dataset} \\ \cmidrule{2-5}
    & \texttt{MNIST} & \begin{tabular}{@{}c@{}} synthetic \\ $\rho = 10^{-5}$ \\ $\psi = 0.2$ \end{tabular} &
      \texttt{webspam} & \texttt{url} \\ \midrule
    Linear              & \textbf{1.10} & \textbf{26.55} & \textbf{0.35} & \textbf{3.30} \\
    Sigmoid             & 1.12 & 50.05 & 0.50 & 5.80 \\
    LReLU               & 1.28 & 30.57 & 0.48 & 3.86 \\
    ReLU                & 1.25 & 30.59 & 0.38 & 3.78 \\
    \midrule
    Reference RP layer  & 1.11 & 27.49 & 0.36 & 3.75 \\
    \bottomrule
  \end{tabular}
\end{table*}

\setlength{\tabcolsep}{18pt}
\begin{table*}[htb]
  \caption{Test errors (\%) for networks with Count Sketch finetuned random projection layer using different activation
  functions.}
    \label{tab:lrp_activation_count}
  \centering
  \begin{tabular}{ccccc} \toprule
    Activation function & \multicolumn{4}{c}{Dataset} \\ \cmidrule{2-5}
    & \texttt{MNIST} & \begin{tabular}{@{}c@{}} synthetic \\ $\rho = 10^{-5}$ \\ $\psi = 0.2$ \end{tabular} &
      \texttt{webspam} & \texttt{url} \\ \midrule
    Linear              & \textbf{1.22} & \textbf{20.16} & \textbf{0.25} & \textbf{0.75} \\
    Sigmoid             & 1.23 & 28.11 & 0.36 & 1.05 \\
    LReLU               & 1.51 & 27.15 & 0.34 & 0.82 \\
    ReLU                & 1.41 & 27.16 & 0.33 & 0.81 \\
    \midrule
    Reference RP layer  & 1.34 & 20.42 & 0.32 & 0.96 \\
    \bottomrule
  \end{tabular}
\end{table*}

For our experiments, we used the same network architecture as in the previous experiments. We report the early stopping
errors for different activation functions in Table~\ref{tab:lrp_activation_sparse} and
Table~\ref{tab:lrp_activation_count}. Compared to networks with the fixed-weight \gls{RP} layer, networks with linear
finetuned \gls{RP} layer performed better on all datasets. Importantly, linear finetuned \gls{RP} layer further improved
the state-of-the-art results on \texttt{webspam} and \texttt{url} datasets. Surprisingly, introducing a nonlinearity
after the \gls{RP} layer decreased the network performance. In fact, networks with a nonlinearity after the finetuned
\gls{RP} layer performed very similarly to, or were outperformed by, networks with the fixed-weight \gls{RP} layer. We
hypothesize that this poor performance is a consequence of the small size of the \gls{RP} layer output\footnote{Because
of the computational cost, in our main experiments we limited the output of the finetuned \gls{RP} layer to 1000
dimensions.} and the sparse connectivity in the \gls{RP} layer. Particularly, when a sparse \gls{RP} layer processes a
sparse training example, the total input to the nonlinearity is also sparse. If we now apply an element-wise nonlinear
transformation, we lose some information about the projected example. Specifically, in the case of \gls{ReLU}
activation, we effectively zero-out, on average, half of the non-zero elements in the sparse input. We believe that this
loss of information causes the decrease in network performance. If our hypothesis is correct, \acrlong{RP} followed by a
nonlinear activation function should perform better with larger output dimensionality.

\begin{figure}[!htb]
  \includegraphics[width=1.0\textwidth]{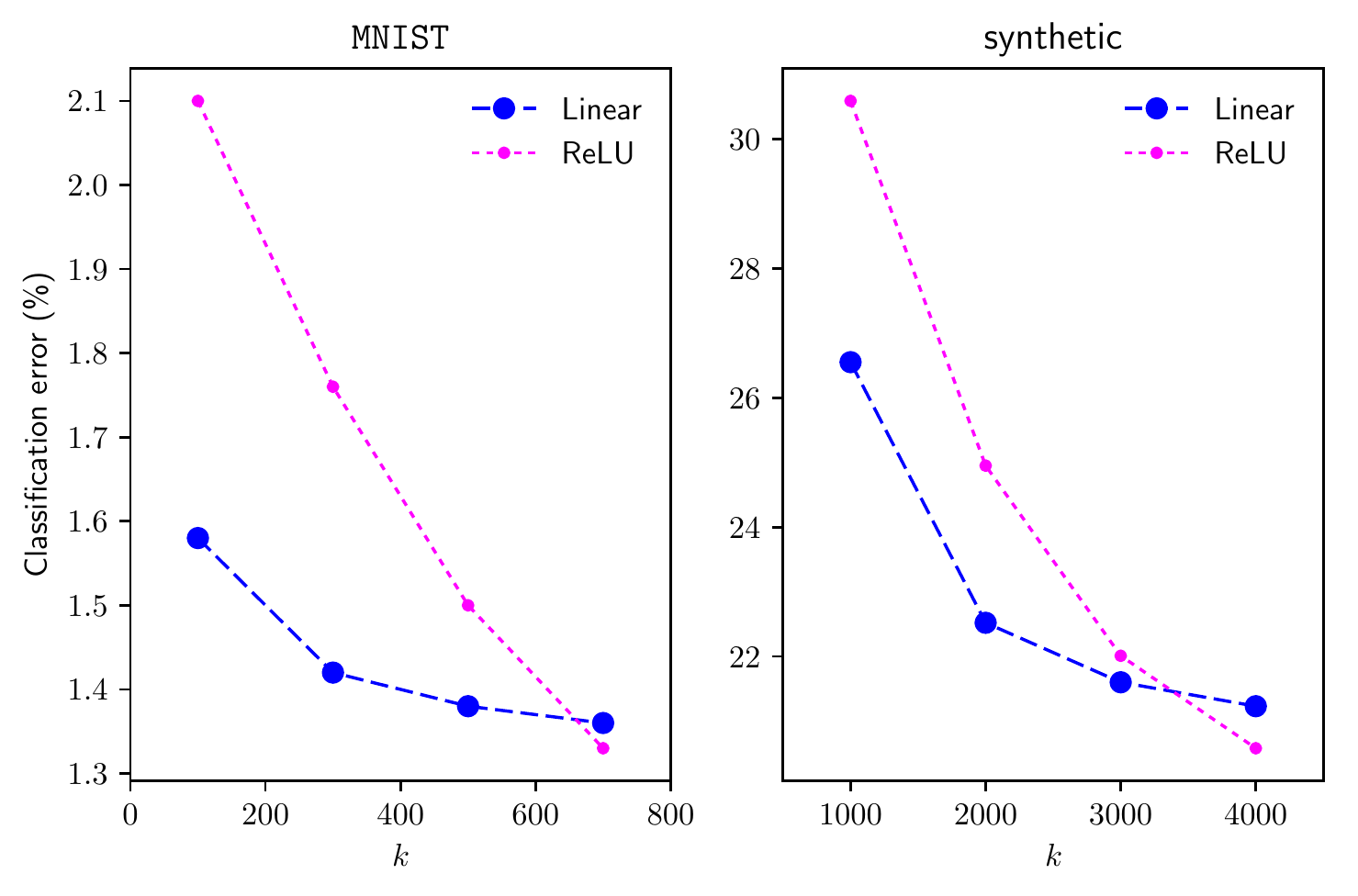}
  \caption{Performance of networks with different activation functions in the finetuned random projection layer for
  varying number of outputs, $k$, in the RP layer.}
  \label{fig:lrp_avk}
\end{figure}

To verify this hypothesis, we performed additional experiments with larger \gls{RP} layers. In particular, we
experimented with Li's \gls{RP} layer on \texttt{MNIST} and the synthetic dataset. Training large \gls{RP} layers on
\texttt{webspam} and \texttt{url} was not possible due to the computational cost. To reduce the training time, we
employed smaller network architectures compared to the previous experiments. Specifically, in experiments on
\texttt{MNIST} we trained networks with \mbox{784-$k$-300-10} architecture, for \mbox{$k \in \{100, 300, 500, 700\}$},
and in experiments on the synthetic datasets we used \mbox{$10^6$-$k$-3000-1} architecture, for \mbox{$k \in \{1000,
2000, 3000, 4000\}$}. We experimented with linear and \gls{ReLU} activation function after \gls{RP}. We employed the
same training settings as in the previous experiments. For each $k$ and each activation function we selected the
learning hyperparameters with experiments on the validation sets. Figure~\ref{fig:lrp_avk} presents the early stopping
errors for networks with different activation functions in the \gls{RP} layer and varying \gls{RP} layer size.
\setlength{\tabcolsep}{0.0em}

Our results suggest that introducing the \gls{ReLU} activation function after the \gls{RP} layer can improve the network
performance, provided that the dimensionality of the \gls{RP} layer $k$ is sufficiently high. In our experiments on the
synthetic dataset, it was necessary to use 4000 units in the \gls{RP} layer to make \gls{ReLU} viable. However, such a
large \gls{RP} layer greatly increases the overall computational cost of training. Therefore, for practical applications
involving large, high-dimensional data we recommend using networks with linear finetuned \gls{RP} layers.

\section{Implementation notes}
\label{sec:rp_impl_details}

In this section, we present important implementation details of algorithms used in this chapter. In particular, we focus
on the technicalities of implementing efficient \acrlong{RP} of sparse large-scale data.

\subsection*{Deep neural networks}

We implemented a library with all essential deep learning algorithms and models that we use in this chapter. The
implementation is designed to run on \gls{GPGPU} and is written in C++ and
CUDA\footnote{\url{http://www.nvidia.com/object/cuda_home_new.html}}. It supports \glspl{RBM}, \glspl{DBN}, \glspl{MLP}
and deep autoencoders along with their training algorithms: error backpropagation for feedforward networks and \gls{CD}
for energy-based models. As most deep learning algorithms map well to the level 3 BLAS (Basic Linear Algebra
Subprograms) operations, we implemented them with NVIDIA CUDA BLAS library
(cuBLAS)\footnote{\url{http://docs.nvidia.com/cuda/cublas/index.html}}. To efficiently produce pseudo-random numbers
from various distributions we use the NVIDIA CUDA Random Number Generation library
(cuRAND)\footnote{\url{http://docs.nvidia.com/cuda/curand/host-api-overview.html}}. Our library functionality is exposed
in Python via bindings based on Boost
Python\footnote{\url{http://www.boost.org/doc/libs/1_64_0/libs/python/doc/html/index.html}}. The bindings make the
library easy to use and hide the low-level memory management of the underlying objects. For a more elaborate description
of our GPU-accelerated library and its features see~\citep{dlcuda2015}. As shown in~\citep{dlcuda2015}, our
implementation offers a significant speedup compared to NumPy/OpenBLAS and MATLAB implementations running on a
multi-core system.

All of our experiments with \glspl{DNN} were conducted on NVIDIA Tesla K40 XL GPUs, which are based on the Kepler
microarchitecture.

\subsection*{Random projection methods}

We implemented the five \gls{RP} methods from Section~\ref{sec:rp_schemes} in Python with extensive use of
NumPy\footnote{\url{http://www.numpy.org/}}. NumPy is a Python library for scientific computing that provides us with
simple handling of large multidimensional arrays, an abundance of useful linear algebra procedures and random number
generation capabilities. For sparse matrix construction and manipulation and efficient sparse matrix multiplication
procedures we use the \texttt{scipy.sparse} module, which is a part of the SciPy
library\footnote{\url{https://www.scipy.org/}}.

\paragraph{Data projection.}

To randomly project an $n \times d$ data matrix $\mathbf{A}$ with a given \gls{RP} scheme we first explicitly construct
its $d \times k$ projection matrix $\mathbf{P}$ and then compute the matrix product $\mathbf{AP}$. This lets us
implement all projection schemes in a uniform framework. Note that by enforcing the explicit construction of
$\mathbf{P}$ we only slightly hinder the \gls{SRHT} projection -- during construction of \mbox{$\mathbf{P}_{SRHT} =
\frac{1}{\sqrt{k}}\mathbf{DHS}$} we are still able to benefit from the speedup provided by the fast Fourier transform
when computing $\mathbf{HS}$ (see Section~\ref{sec:rp_srht}). Because all datasets in our evaluations are represented by
sparse matrices in the \gls{CSR} format, we use either sparse-dense or sparse-sparse matrix multiplication algorithm,
depending on the format of the projection matrix. When the projection matrix is dense, i.e., for Gaussian, Achlioptas'
and \gls{SRHT} schemes we use the \texttt{csr\_matvec}
function\footnote{\url{https://github.com/scipy/scipy/blob/v0.19.1/scipy/sparse/sparsetools/csr.h}}, which runs in
\mbox{$\mathcal{O}(k(\mathrm{nnz}(\mathbf{A})+n))$} operations. When $\mathbf{P}$ is sparse, i.e., for Li's and Count
Sketch schemes, we employ the \gls{SMMP} algorithm~\citep{bank1993sparse}, which has the computational complexity of
\mbox{$\mathcal{O}(nK^2 + \max\{n,d\})$}, where $K$ is the maximum number of non-zero elements in rows of $\mathbf{A}$
and columns of $\mathbf{P}$. Despite the quadratic dependence on $K$, in practice, \gls{SMMP} is much more efficient
than \texttt{csr\_matvec}. While in the worst case the multiplication may require $\mathcal{O}(nd^2)$ operations, $K$ is
typically much lower than $d$ and grows very slowly with $k$. Of course, for higher densities of $\mathbf{A}$ the
maximum number of non-zero elements in rows of $\mathbf{A}$ may decide the value of $K$. However, for most sparse
real-world datasets this value does not exceed several hundreds (see Table~\ref{tab:datasets_nnz}).
\setlength{\tabcolsep}{10pt}
\begin{table*}[htb]
  \caption{Maximum and average number of non-zero elements in rows of the training data matrix for large-scale
  datasets.}
  \label{tab:datasets_nnz}
  \centering
  \begin{tabular}{rrrr} \toprule
    Dataset        & \phantom{a} & \multicolumn{2}{c}{Number of non-zero elements per example} \\ \cmidrule{3-4}
                                && Average & Maximum \\ \midrule
    \texttt{url}                && 71      & 414 \\
    \texttt{webspam}            && 63      & 46,947 \\
    \texttt{KDD2010-a}          && 15      & 85 \\
    \texttt{KDD2010-b}          && 19      & 75 \\
    synthetic, $\rho = 10^{-4}$ && 100     & 148 \\
    \bottomrule
  \end{tabular}
\end{table*}
Even for \texttt{webspam}, where the maximum number of non-zero elements in rows is high (and therefore determines $K$),
the multiplication time of \gls{SMMP} is low ($\sim$ 20 seconds for Count Sketch and $\sim$ 4 minutes for Li's matrix;
projection to $k=1000$ dimensions) compared to \texttt{csr\_matvec} ($\sim$ 30 minutes for the Gaussian matrix). For
most datasets, however, the value of $K$ is determined by the maximum number of non-zero elements in columns of
$\mathbf{P}$. For Count Sketch, the value of $K$ can be estimated using the analogy of the Count Sketch matrix
construction to the balls-into-bins problem~\citep{raab1998balls}. If we treat columns of $\mathbf{P}$ as bins and the
non-zero elements as balls, we can calculate with high probability (defined as probability that tends to 1 when the
number of columns grows) the maximum bin load, which is equivalent to $K$. Specifically, when \mbox{$d \ge k \log{k}$}
the maximum load of any bin is $\Theta(\frac{d}{k})$, i.e., on the order of the mean value. Therefore, the time of
multiplying the data matrix by the Count Sketch projection matrix decreases with the growth of $k$ and for higher values
of $k$ is lower-bounded by the maximum non-zero element count in the rows of the data matrix. For Li's scheme, the value
of $K$ grows with $d$ and $k$, however, this dependence is not linear. Obviously, $K$ depends on $d$ as each column of
$\mathbf{P}$ contains, on average, $\sqrt{d}$ non-zero elements. $K$'s dependence on $k$ is a consequence of the fact
that increasing the number of columns also increases the probability of adding a column with a higher number of non-zero
elements.

In Table~\ref{tab:rp_layer_complexity} we report a summary of the time complexity of constructing different projection
matrices and using them to perform \acrlong{RP}.
\setlength{\tabcolsep}{8pt}
\begin{table*}[htb]
  \caption{Time complexity of constructing a random projection matrix and performing the projection via matrix
  multiplication $\mathbf{AP}$, where $\mathbf{A}$ is a $n \times d$ data matrix and $\mathbf{P}$ is a $d \times k$
  projection matrix. $\mathrm{nnz}(\mathbf{A})$ denotes the number of non-zero elements in matrix $\mathbf{A}$. $K$ is
  the maximum number of non-zero elements in rows of $\mathbf{A}$ and columns of $\mathbf{P}$.}
  \label{tab:rp_layer_complexity}
  \centering
  \begin{tabular}{rccc} \toprule
    Random projection scheme & \phantom{} & Matrix construction & Matrix multiplication \\ \midrule
    Gaussian     && $\mathcal{O}(dk)$             & $\mathcal{O}(k(\mathrm{nnz}(\mathbf{A}) + n))$ \\
    Achlioptas'  && $\mathcal{O}(dk)$             & $\mathcal{O}(k(\mathrm{nnz}(\mathbf{A}) + n))$ \\
    Li's         && $\mathcal{O}(d^\frac{1}{2}k)$ & $\mathcal{O}(nK^2 + max\{n, d\})$ \\
    SRHT         && $\mathcal{O}(d(k + \log{d}))$ & $\mathcal{O}(k(\mathrm{nnz}(\mathbf{A}) + n))$ \\
    Count Sketch && $\mathcal{O}(d)$              & $\mathcal{O}(nK^2 + max\{n, d\})$ \\
    \bottomrule
  \end{tabular}
\end{table*}

\paragraph{Matrix slicing.}

With high data and projection dimensionality, even the projection matrix alone may not fit into the available memory.
For example, in our experiments on the real-world datasets projecting the \texttt{KDD2010-b} data matrix to $1000$
dimensions using a dense \gls{RP} scheme requires constructing a projection matrix that needs nearly $120$GB of memory,
assuming a $4$-byte floating point representation. To perform the projection efficiently, additional space is also
needed for the input data and the projected result. To alleviate the memory consumption problem we employed two types of
data matrix slicing: horizontal slicing of the input data and vertical slicing of the projection matrix. With these two
techniques, we can control the amount of memory that is needed to perform the projection. Of course, these techniques
come with an additional cost of disk IO operations.

\begin{figure}[htb!]
  \begin{tabular}{m{0.6\linewidth}cm{0.3\linewidth}}
    \includegraphics[width=\linewidth]{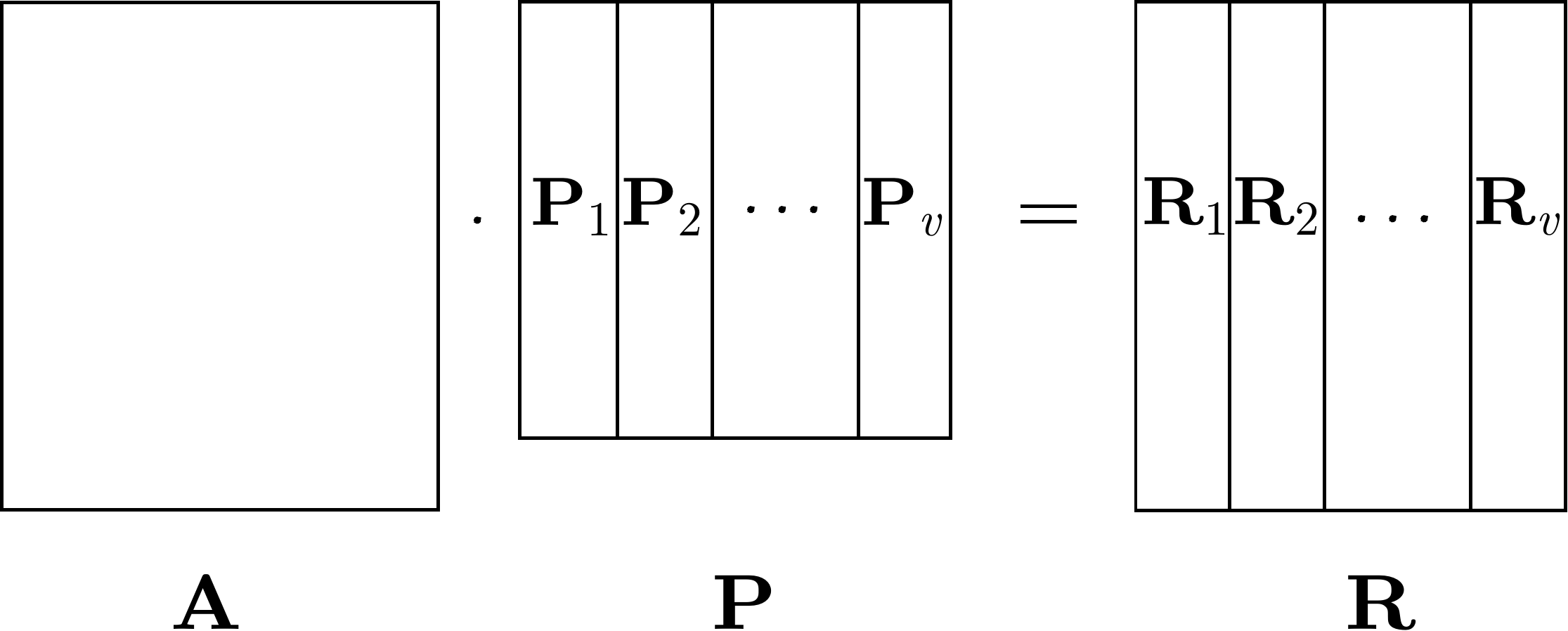} & \phantom{a} &
    \begin{math}
      \begin{aligned}
        \mathbf{R}_i &= \mathbf{AP}_i, \text{ } i \in [1, v] \\
        \mathbf{R}            &= [\mathbf{R}_1, \ldots, \mathbf{R}_v]
      \end{aligned}
    \end{math}
  \end{tabular}
  \caption{Random projection procedure with columnwise slicing of the projection matrix. $\mathbf{A}$ is the dataset
matrix, $\mathbf{P}$ is the projection matrix and $\mathbf{R}$ is the result matrix. The number of projection matrix
slices is $v$. $[\mathbf{X}, \mathbf{Y}]$ operation denotes the columnwise concatenation of matrices $\mathbf{X}$ and
$\mathbf{Y}$.}
  \label{fig:p_slicing}
\end{figure}

To randomly project a data matrix $\mathbf{A}$ using an \gls{RP} matrix $\mathbf{P}$ we do not need to create the whole
projection matrix at once. Instead, we can generate $v$ vertical slices $\mathbf{P}_i$ of the projection matrix
$\mathbf{P}$ one by one and use them to calculate slices $\mathbf{R}_i$ of the result matrix $\mathbf{R}$. This
procedure is illustrated in Figure~\ref{fig:p_slicing}. Importantly, at any point during the projection, we only need to
store $\mathbf{A}$, $\mathbf{P}_i$ and $\mathbf{R}_i$ in memory.

We employ slicing of $\mathbf{P}$ for the Gaussian, Achlioptas', Li's and \gls{SRHT} \gls{RP} matrices. Slicing of the
Count Sketch matrix $\mathbf{P}_{CS}$ is not necessary since even for high dataset dimensionality $d$ and projection
dimensionality $k$ it only contains $d$ non-zero elements. Therefore, it takes just up to a few hundred megabytes of
memory. Columnwise slices of Gaussian, Achlioptas' and Li's projection matrices can be generated using the same random
number distributions as the one that is used to create their respective unsliced projection matrices. Of course, when
creating slices $\mathbf{P}_i$ we must appropriately scale the non-zero elements depending on the number of slices $v$.
Slices of the \gls{SRHT} matrix $\mathbf{P}_{SRHT}$ can also be easily generated. $\mathbf{P}_{SRHT}$ construction
starts with a sparse matrix whose generation uses a random number distribution similar to Li's matrix. This matrix is
then deterministically transformed with the Walsh-Hadamard transform and has a random half of its rows multiplied by
$-1$ (see Section~\ref{sec:rp_schemes}). Therefore, $\mathbf{P}_{SRHT}$ slices are defined by the initial Li's-like
matrix slices, which, as discussed above, can be easily generated.

\begin{figure}[htb!]
  \begin{tabular}{m{0.62\linewidth}m{0.02\linewidth}m{0.3\linewidth}}
    \includegraphics[width=\linewidth]{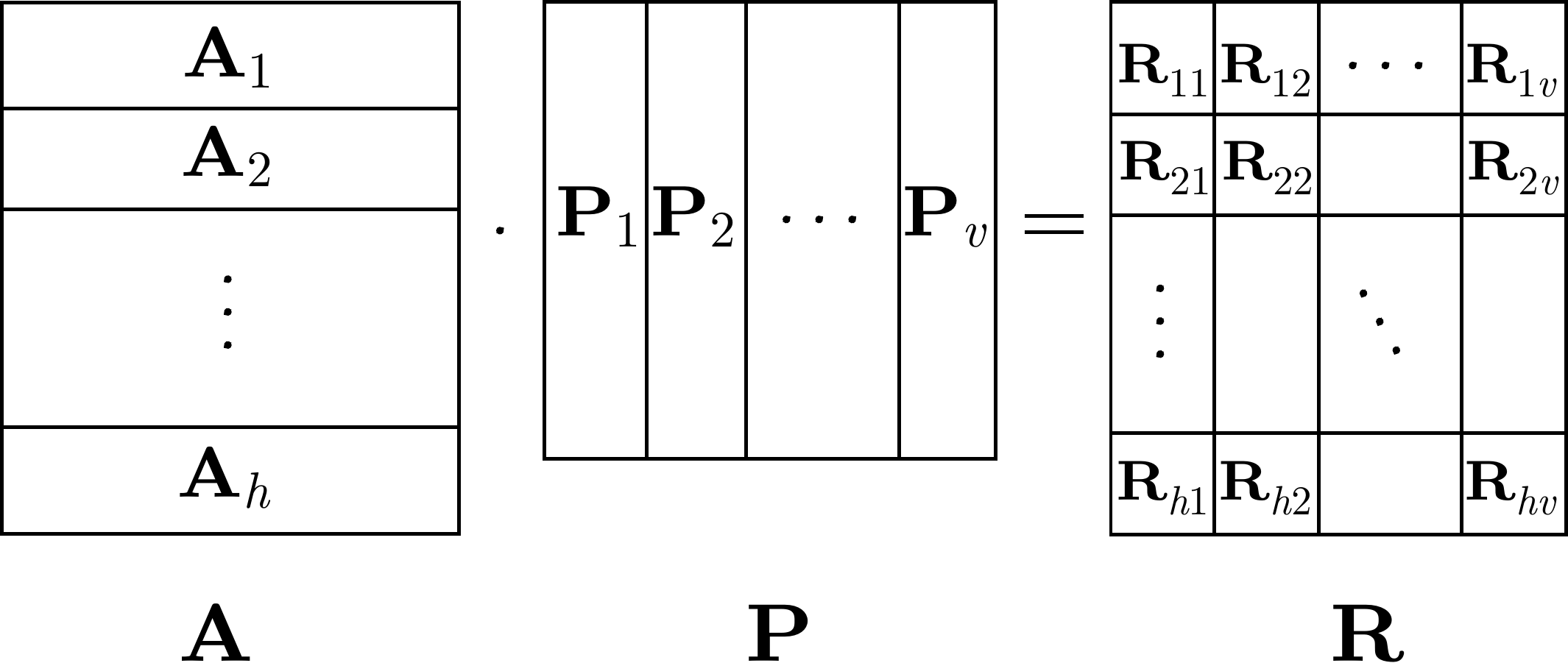} &&
    \begin{math}
      \begin{aligned}
        \mathbf{R}_{ij} &= \mathbf{A}_i\mathbf{P}_j \text{, } i \in [1, h], 
          \text{ } j \in [1, v]\\
        \mathbf{R}             &= 
          \begin{bmatrix}
            \mathbf{R}_{11} & \mathbf{R}_{12} & \cdots & \mathbf{R}_{1v} \\
            \mathbf{R}_{21} & \mathbf{R}_{22} & & \mathbf{R}_{2v} \\
            \vdots & & \ddots & \vdots \\
            \mathbf{R}_{h1} & \mathbf{R}_{h2} & \cdots & \mathbf{R}_{hv}
          \end{bmatrix}
      \end{aligned}
    \end{math}
  \end{tabular}
  \caption{Random projection procedure with rowwise slicing of the data matrix and columnwise slicing of the projection
matrix. $\mathbf{A}$ is the dataset matrix, $\mathbf{P}$ is the projection matrix and $\mathbf{R}$ is the result matrix.
$v$ and $h$ are the numbers of projection and dataset matrix slices, respectively.}
  \label{fig:dp_slicing}
\end{figure}

When the data matrix $\mathbf{A}$ takes a significant amount or RAM or does not fit into the memory, it may also be
necessary to split it rowwise into smaller parts $\mathbf{A}_i$. This is the case, for example, for the \texttt{webspam}
dataset. In Figure~\ref{fig:dp_slicing} we illustrate how to perform dataset slicing in addition to the projection
matrix slicing. Both the number of projection matrix slices $v$ and the number of dataset slices $h$ can be adjusted to
control the amount of required memory.
\setlength{\tabcolsep}{0pt}

\begin{algorithm}
\caption{Pseudocode for the random projection procedure with dataset and projection matrix slicing. $\mathbf{A}$ is the
dataset matrix, $\mathbf{P}$ is the projection matrix, and $\mathbf{R}$ is the result matrix. $v$ and $h$ are the
numbers of projection and dataset matrix slices, respectively.}
\label{alg:rp_chunking}
\begin{algorithmic}[1]
    \STATE split $\mathbf{A}$ rowwise into $h$ slices, so that $\mathbf{A} = [\mathbf{A}_1, \ldots, \mathbf{A}_h]^T$
    \STATE split $\mathbf{P}$ columnwise into $v$ slices, so that $\mathbf{P} = [\mathbf{P}_1, \ldots, \mathbf{P}_v]$
    \FOR{$j = 1$ \TO $v$}
      \FOR{$i = 1$ \TO $h$}
        \STATE $\mathbf{R}_{ij} = \mathbf{A}_i\mathbf{P}_j$
        \STATE save $\mathbf{R}_{ij}$
      \ENDFOR
      \STATE read $\mathbf{R}_{kj}, \text{ for } k \in [1, h]$
      \STATE $\mathbf{R}_{\bullet j} = [\mathbf{R}_{1j}, \ldots, \mathbf{R}_{hj}]^T$
      \STATE save $\mathbf{R}_{\bullet j}$
    \ENDFOR
    \STATE read $\mathbf{R}_{\bullet k}, \text{ for } k \in [1, v]$
    \RETURN $\mathbf{R} = [\mathbf{R}_{\bullet 1}, \ldots, \mathbf{R}_{\bullet v}]$
\end{algorithmic}
\end{algorithm}

The pseudocode for \gls{RP} with matrix slicing is presented in Algorithm~\ref{alg:rp_chunking}. Since the partial
projection results $\mathbf{R}_{\bullet j}$ and the final result $\mathbf{R}$ are small compared to the projection and
dataset matrices, the most memory-demanding part of Algorithm~\ref{alg:rp_chunking} is the projection phase (line 5).
During this operation, the procedure needs to store only a single dataset slice $\mathbf{A}_{i}$, one projection matrix
slice $\mathbf{P}_{j}$ and the projection result $\mathbf{R}_{ij}$.

\subsection*{Dimensionality reduction methods}

For the \gls{PCA}, Chi-square and F-score dimensionality reduction procedures we used the implementations from the
\texttt{scikit-learn} Python library\footnote{\url{http://scikit-learn.org}}. \texttt{scikit-learn} natively supports
sparse matrices, is computationally efficient and has a small memory footprint. To compute \gls{PCA}, we employed the
truncated \gls{SVD} algorithm with the randomized solver by \citet{halko2011algorithm}. Importantly, in addition to
being computationally efficient, it can also work directly on sparse data. Since truncated \gls{SVD} does not center
features in the input data, we normalized each feature in the \texttt{MNIST} dataset. Normalization was not necessary
for the synthetic dataset variants, as their construction ensures that each feature is centered around zero. To select
features in the synthetic datasets with the Chi-square method we used data matrices with absolute feature values. This
is because the Chi-square implementation requires the entries of the data matrix to be non-negative. A common practice
is to rescale every feature to the $[0,1]$ interval. However, this operation is impossible for the synthetic datasets as
it would destroy their sparse structure and raise the memory footprint to almost $4$TB for each dataset (assuming a
$4$-byte representation of the floating-point numbers). We implemented the \gls{IG} feature selection by creating a
custom score function for the \texttt{SelectKBest}
class\footnote{\url{http://scikit-learn.org/stable/modules/generated/sklearn.feature_selection.SelectKBest.html}} in the
\texttt{scikit-learn} package.

\section{Conclusions}

In this chapter, we studied the viability of training \glspl{DNN} with the \gls{RP} layer with the goal of creating
models that can efficiently learn from sparse, high-dimensional data. Our results demonstrate that networks with
\gls{RP} layer can match or improve over the state-of-the-art classification results on data with millions of dimensions
and no spatial structure. This opens a path to applying neural networks in tasks where directly learning from the data
would be infeasible: experiments on the \texttt{KDD2010} datasets, for example, involved up to $30,000$-fold reduction
of the input dimensionality.

We studied two variants of the \gls{RP} layer: one with weights that are fixed during training and one where they are
finetuned with error backpropagation. Our experimental evaluation of \glspl{DNN} with fixed-weight \gls{RP} layer shows
that Gaussian, Achlioptas', \gls{SRHT} and Count Sketch projections perform well, while the Li's projection yields worse
results. This could be attributed to the sparsity of the projected data -- on the \texttt{MNIST} dataset, which is
dense, Li's method performed well. Note also that Achlioptas', Count Sketch, Li's and \gls{SRHT} are fast: the first
three do not employ dense projection matrices and the last one can be computed efficiently using a transform similar to
the fast Fourier transform. Taking this into account, \gls{SRHT} and Count Sketch projections combine the best network
performance with efficient data projection. We also experimented with using \gls{RP} for \gls{BOW} data. Specifically,
we experimented with training deep autoencoders similar to the ones described in~\citep{salakhutdinov2009semantic} on
randomly projected \gls{BOW} vectors. While this approach enabled us to train autoencoders on larger dictionaries, it
did not achieve performance comparable to the reference networks. This result can be a consequence of two facts. First,
the autoencoders with projected data require Gaussian input units. The reference networks employ the constrained Poisson
model, which is tailored to \gls{BOW} data. Second, the dictionary used by the reference models already captured most of
the word count in the text.

Our experiments with finetuned \gls{RP} layer suggest that adjusting the non-zero weights in a sparse \gls{RP} layer can
significantly improve the overall network performance. In particular, by using the finetuned Count Sketch \gls{RP} layer
we were able to train networks that achieved more than 30\% lower classification error on \texttt{webspam} and
\texttt{url} datasets, compared to the state-of-the-art methods. To make training of the \gls{RP} layer feasible we
employed several architectural optimizations and training regime modifications. First, instead of normalizing the input
data we applied \acrlong{BN} after the \gls{RP} layer. Second, we finetuned only these \gls{RP} weights that were
initially non-zero. Finally, we found that applying a nonlinear activation after the \acrlong{BN} is viable only when
the input data is projected to a high-dimensional space. In practice, the performance gain from this nonlinearity does
not justify the additional computational cost introduced by finetuning an \gls{RP} layer with a high-dimensional output.

%% file: rp_init.tex
\cleardoublepage
\chapter{Initializing deep networks with random projection matrices}
\label{cha:rp_init}

How should we initialize weights in deep neural networks? The answer to this question depends on the network
architecture, neuron connectivity and the activation function. Most often it involves a carefully scaled normal or
uniform distribution. But what if we used a less trivial random number distribution? In this chapter we investigate the
performance of deep neural networks with weights initialized using \gls{RP} matrices described in
Section~\ref{sec:rp_schemes}. In particular, we study five \gls{RP} weight initialization schemes: two dense, i.e.,
Gaussian and \gls{SRHT}, and three sparse, i.e., Achlioptas', Li's and Count Sketch. We focus mostly on rectifier
networks, as the \gls{ReLU} transfer function is currently the most popular choice for efficient training of deep
architectures~\citep{lecun2015deep}.

We begin by stressing the importance of weight initialization in deep networks. We briefly review the most popular
initialization schemes and related techniques. We then motivate the viability of using \gls{RP} matrices that satisfy
the Johnson--Lindenstrauss lemma as the initial weights in deep networks. We follow up by investigating \gls{RP}
initialization in \glspl{CNN} and in pretrained fully-connected networks. We show that using \gls{RP} matrices as
initial weights is a viable approach in \glspl{CNN}. Specifically, in our evaluation \glspl{CNN} initialized with
\gls{SRHT} matrices consistently outperformed the current state-of-the-art initialization scheme on several image
classification datasets. Most pretrained networks, however, did not benefit from \gls{RP} initialization.

Results from this chapter were presented in~\citep{wojcik2017random}. A Torch7 package implementing our \gls{RP}
initialization for \glspl{CNN} is available at \url{https://github.com/piotriwojcik/rpinit}.

\section{Weight initialization techniques}
\label{sec:weight_init}

For typical cost functions training a deep neural network can be viewed as a non-convex optimization problem. As the
loss function contains multiple local minima, the optimization process tends to converge to different solutions,
depending on the initial conditions. Therefore, initial network parameters, i.e., weights and biases can significantly
affect the speed of convergence and the quality of the found solution~\citep{larochelle2009exploring}. A bad
initialization scheme may even prevent the network from learning. The simplest example is an all-zero initialization. It
leads to a situation where every hidden unit in a given layer receives the same input signal ($\sum_{i} {x_{i}w_i} + b =
0$, because $\forall_i w_i=0$ and $b=0$), produces the same output and computes the same gradient during
backpropagation. Consequently, all neurons receive the same weight updates and learn the same function, wasting the
network capacity. To avoid this symmetry, the initial weights should not be equal. Note that the symmetry breaking is
not strictly necessary for the output layer, as every output unit receives different gradient signal, and thus learns
different weights~\citep{bengio2012practical}.

A simple and popular initialization approach is, therefore, to initialize the weight matrix $\mathbf{W}$ with random
values drawn from some probability distribution. In fact, a zero-mean normal distribution with a small variance:
\begin{equation}
  W_{ij} \sim \mathcal{N}(0, s^2),
\end{equation}
often works surprisingly well. Such initialization scheme with $s = 10^{-2}$ was used, for example, in the influential
ImageNet network by \citet{krizhevsky2012imagenet}. One advantage of using small initial weights is particularly clear
when training a network with sigmoid or hyperbolic tangent activation functions. The derivative of the sigmoid function
is:
\begin{equation}
  \sigma'(x) = \sigma(x)(1 - \sigma(x)),
\end{equation}
with maximum for $\sigma(x) = 0.5$. Since the weight updates in backpropagation are proportional to this derivative,
sigmoid units will learn the fastest for activations close to $0.5$, i.e., when the total unit input is close to zero
(\mbox{$\sigma(0) = 0.5$}). When the magnitude of the input rises, the derivative of the activation function quickly
becomes very small, making the weight updates inefficient. This leads to units entering a saturated state, which
effectively stops their learning. On the other hand, the gradient is proportional to the magnitudes of weights.
Therefore, setting the weights too low also leads to small gradient and slow learning process. This becomes an issue
especially when backpropagating through deeper networks, where the gradient diminishes exponentially with each layer.
This exponential decrease (or explosion, if the initial weights are too large) of the error signal as a function of the
distance from the output layer is often referred to as the vanishing (or exploding) gradient
problem~\citep{hochreiter2001gradient}.

The issues mentioned above prompted a vigorous research on network initialization. The importance of proper
initialization when using first-order optimization methods, such as \gls{SGD}, was emphasized
in~\citep{sutskever2013importance}. Therein, \citeauthor{sutskever2013importance} showed that \gls{SGD} with momentum
and carefully initialized weights can yield results comparable to training with higher-order methods, such as
Hessian-free optimization. A parallel line of research focused on making deep networks more robust to the choice of
initial weights. For example, \gls{ReLU}~\citep{nair2010rectified} is a significant step towards alleviating the
vanishing gradient problem. Training can also be facilitated with \acrlong{BN}~\citep{ioffe2015batch}, which normalizes
the distribution of activations in hidden layers. However, despite these advances, networks employed to obtain current
state-of-the-art results still use carefully designed weight initialization schemes~\citep{he2015delving}.

The topic of weight initialization is vast and includes a number of different approaches ranging from simply drawing the
weights from a well-designed probability distribution to more complex approaches, such as transfer learning. Below we
briefly review the most common weight initialization techniques.

\paragraph{Pre-deep learning weight initialization.}

Before the advent of modern deep neural networks, multiple weight initialization techniques were developed to increase
the backpropagation convergence rate and quality of solutions. These approaches include, e.g., genetic algorithms,
simulated annealing~\citep{masters1993practical}, linear algebraic methods~\citep{yam2000weight} and other
techniques~\citep{drago1992statistically,martens1996stochastically,nguyen1990improving}. However, these methods are
usually impractical in modern deep neural networks because of being either computationally too expensive or suitable
only for the sigmoid or hyperbolic tangent activation functions.

\paragraph{Sparse initialization.}

\citet{martens2010deep} proposed the Hessian-free optimization method for backpropagation networks. While this algorithm
outperforms the standard \gls{SGD}, it too benefits from a well-designed random initialization scheme. In particular,
the best results in~\citep{martens2010deep} were obtained with there-proposed \gls{SI} approach. \Acrlong{SI}
initializes units with sparse, randomly generated weight vectors. In particular, a fixed number of elements (15 in
Martens' experiments) is randomly chosen in each weight vector. These elements are initialized with random weights,
usually drawn from a Gaussian distribution, while the other elements are set to zero. \Acrlong{SI} was designed to
fulfill two goals: to prevent the saturation of network units and to make the units initially as different from each
other as possible. \citet{sutskever2013importance} confirmed the usefulness of \gls{SI} also in networks trained with
\gls{SGD}.

\paragraph{Random initialization with scaled variance.}

A significant improvement in weight initialization stemmed from an observation that by simply setting the weights to
small random numbers one does not take into account the variance of the layer activations and gradients. A better way to
initialize the weights is, therefore, to scale them in a way that normalizes these variances. Without such
normalization, the magnitudes of activations of final layers in deeper architectures could either be extremely large or
too small to produce a proper training gradient. \citet{lecun1998efficient} recommend scaling down the weights of
sigmoid units by the square root of the so-called \textit{fan-in}, i.e., the number of unit inputs. The weights were
then initialized with:
\begin{equation}\label{eq:lecun_init}
  W_{ij} \sim U\bigg[ -\frac{1}{\sqrt{f_\mathrm{in}}}, \frac{1}{\sqrt{f_\mathrm{in}}}\bigg],
\end{equation}
where $f_\mathrm{in}$ is the \textit{fan-in}, and $U[-x,x]$ is the uniform distribution over the $[-x,x]$ interval.
Glorot et al.~\citep{glorot2010understanding} in their initialization scheme for sigmoid and hyperbolic tangent units
additionally employed the \textit{fan-out}, i.e., the number of outputs:
\begin{equation}\label{eq:glorot_init_sigmoid}
  W_{ij}\text{(sigmoid)} \sim U\bigg[-\sqrt{\frac{6}{f_\mathrm{in} + f_\mathrm{out}}}, \sqrt{\frac{6}{f_\mathrm{in} +
  f_\mathrm{out}}}\bigg],
\end{equation}
\begin{equation}\label{eq:glorot_init_tanh}
  W_{ij}\text{(tanh)} \sim U\bigg[-4\sqrt{\frac{6}{f_\mathrm{in} + f_\mathrm{out}}}, 4\sqrt{\frac{6}{f_\mathrm{in} +
  f_\mathrm{out}}}\bigg].
\end{equation}
This initialization scheme, often called the \textit{Xavier} initialization, was one of the factors that made it
possible to move away from generative pretraining and successfully train deep networks from
scratch~\citep{glorot2011deep}. However, Glorot's initialization was not primarily designed for rectified linear units.
A weight initialization scheme suitable for \gls{ReLU} and Leaky ReLU, the so called \textit{He's} initialization was
proposed in~\citep{he2015delving}:
\begin{equation}\label{eq:hes_init}
  W_{ij} \sim \mathcal{N}\bigg(0, \frac{2}{f_\mathrm{in}}\bigg).
\end{equation}
Using this initialization scheme \citeauthor{he2015delving} reported to have successfully trained~\mbox{30-layer}
networks from scratch. He's method is currently the state-of-the-art initialization scheme for practical applications of
deep \gls{ReLU} networks.

Other works employing initialization schemes motivated by controlling the variance of network activations or the
variance of the weight gradients include, e.g.,~\citep{sussillo2014random,krahenbuhl2015data}.
\citet{sussillo2014random} proposed an initialization scheme, called the random walk initialization that focuses on
preventing the vanishing gradient problem by forcing the norms of backpropagated errors to be constant.
\citet{krahenbuhl2015data} normalized network activations in \glspl{CNN} by using activation statistics estimated from
the training data.

\paragraph{Unsupervised pretraining.}

In their seminal work~\citet{hinton2006reducing} demonstrated that \glspl{DNN} can be trained in two phases:
layer-by-layer unsupervised pretraining using \glspl{DBN}, followed by supervised finetuning with error backpropagation.
The resultant deep models significantly outperformed the state-of-the-art approaches on multiple machine learning
tasks~\citep{hinton2006reducing,hinton2012speech}. Although the introduction of \glspl{ReLU} and efficient
initialization schemes made the pretraining phase not necessary for most applications, many recommendations indicate
that pretraining typically helps~\citep{bengio2012practical,erhan2010does}.

Pretraining can be regarded not only as a type of weight initialization scheme but also as a regularizer that improves
generalization~\citep{larochelle2009exploring}. Another justification for unsupervised pretraining stems from the
imbalance between the available labeled and unlabeled data: data acquisition is relatively inexpensive, compared to
labeling. Therefore, unsupervised pretraining can often incorporate much bigger training sets than supervised
finetuning.

Comparatively less work has been published on initialization of network layers for generative pretraining employed
before supervised finetuning. Similarly to feed-forward architectures, weights before pretraining are typically densely
initialized with random numbers drawn from a zero-mean normal distribution with a small standard deviation:
\begin{equation}
  W_{ij} \sim \mathcal{N}(0, s^2).
\end{equation}
Specifically, \citet{hinton2012practical} recommends using $s = 10^{-2}$. An alternative approach is to initialize these
weights with Martens' sparse initialization scheme, which was originally proposed for backpropagation
networks~\citep{martens2010deep}. Specifically, \citet{grzegorczyk2015effects} showed that sparse initialization in
\glspl{DBN} with \gls{NReLU} hidden layers slightly improves the network performance.

\paragraph{Orthogonal initialization.}

\citet{saxe2013exact} showed that using random orthogonal weights instead of scaled random Gaussian numbers
(Eq.~\ref{eq:lecun_init}) yields a similar quality of initialization to unsupervised pretraining and makes training
deeper models possible. Although \citeauthor{saxe2013exact} derived these results for linear networks, they demonstrated
that orthogonal initialization leads to better gradient propagation also in deep nonlinear networks. This idea was
further extended by~\citet{mishkin2015all} who employed orthonormal initialization combined with batch normalization of
layer output similar to~\citep{ioffe2015batch} but performed only on the first batch.

\citet{grzegorczyk2016encouraging} showed that orthogonal initialization can also be achieved with unsupervised
pretraining. They proposed to explicitly encourage the orthogonality of features learned by an \gls{RBM} model during
the pretraining phase. Their goal was to increase the diversity of learned features. To this end,
\citet{grzegorczyk2016encouraging} modified the \gls{CD} algorithm in a way that penalizes parallel components of the
weight vectors. They showed that deep networks pretrained in this manner can be finetuned to higher levels of
performance, compared to standard pretraining.

\paragraph{Transfer learning.}

A popular trick to speed up the network training, used especially with modern \glspl{CNN}, is to employ features that
were already trained for a different task or on a different dataset~\citep{bengio2012deep}. This is realized by simply
copying (transferring) the weight matrices of several layers of the base network to the layers of the target network.
Remaining layers of the target networks are initialized in a standard manner. Then, either the whole target network or
only the randomly initialized layers are trained to solve the target task. This approach is motivated by the fact that
the initial layers in \glspl{CNN} usually recognize generic image features, while deeper layers are tuned for a specific
task. Therefore, early layers of a trained network should be useful also in other networks trained on similar data.
Transfer learning is especially beneficial in situations when the available dataset has a small number of examples.

\section{Random projection initialization}

Feeding data through a Gaussian initialized network with fully-connected layers realizes an operation similar to a
series of consecutive \glspl{RP} that roughly preserves the distances between the observations. To see this, let us
consider a single untrained neural network layer with $d$ input and $k$ output units, where the weight matrix $\vect{W}
\in \mathbb{R}^{d \times k}$ has been initialized, following a common practice, to small random numbers drawn from a
Gaussian distribution~$\mathcal{N}(0, 10^{-4})$~\citep{hinton2012practical,krizhevsky2012imagenet} and the biases have
been set to zero. In this layout, columns of matrix $\vect{W}$ represent latent features learned by the neurons in the
hidden layer. The layer receives an input vector $\vect{I} \in \mathbb{R}^{d}$ and computes the output $\vect{O} \in
\mathbb{R}^{k}$:
\begin{equation}\label{eq:nn_rp1}
  \vect{O}_{nn} = f(\vect{I}\vect{W}),
\end{equation}
where $f(\vect{x})$ denotes application of the activation function to each element of the vector $\vect{x}$. The usual
bias term was omitted here, as in this example all biases are equal to zero. Note that the computation realized by this
layer is similar to performing an \gls{RP} of the input vector using a projection matrix $\vect{R}$:
\begin{equation}\label{eq:nn_rp2}
  \vect{O}_\text{rp} = \vect{I}\vect{R}.
\end{equation}
In fact, matrices $\vect{W}$ and $\vect{R}$ are constructed in exactly the same manner: they both consist of random
numbers drawn from a Gaussian distribution with a small standard deviation. The only difference between
Eq.~\ref{eq:nn_rp1} and Eq.~\ref{eq:nn_rp2} is the presence of an element-wise activation function $f$. However, in our
settings, application of $f$ does not alter the output vector significantly. This is obviously true for linear units,
where the activation function is an identity. For nonlinear activations that are well approximated by an identity
function near zero, e.g., hyperbolic tangent, exponential linear~\citep{clevert2015fast} or softsign, applying $f$ also
has little impact on the result vector. This is because weights in $\vect{W}$ have small absolute values, and therefore
entries in the vector $\vect{I}\vect{W}$ are also close to zero. For activation functions that are nearly linear and
differentiable around zero, application of $f$ results in shifting the entries of $\vect{I}\vect{W}$ by $f(0)$ and
scaling them with a constant equal to $f'(0)$. This is the case, for example, for the popular sigmoid activation
function, which in our example adds $0.5$ to each entry and scales it by a factor $0.25$. The rectifier linear
activation modifies the input vector in a less trivial way: it zeros-out only the negative entries of
$\vect{I}\vect{W}$. However, the resultant vector $\vect{O}$ still contains, on average, half of the elements of
$\vect{O}_\text{rp}$.

Following the above example, a network consisting of $l$ hidden layers performs $l$ consecutive Gaussian \glspl{RP},
resulting in a $k$-dimensional data representation, where $k$ is the size of the last hidden layer. Since the Gaussian
\gls{RP} satisfies the Johnson-Lindenstrauss lemma~\citep{indyk1998approximate,dasgupta2003elementary}, unless $k$ is
small, feeding data through a Gaussian-initialized network yields an output that roughly preserves the structure in the
original data. (For a formal proof of the distance-preserving nature of Gaussian-initialized networks
see~\citep{giryes2015deep}.) This fact is supported by recent findings that random untrained weights can perform
surprisingly well in certain network architectures. In particular, in a large-scale evaluation of neural network
architectures on multiple object recognition tasks pretrained weights only slightly outperformed completely random
filters~\citep{pinto2009high,pinto2010evaluation}. Similarly, \citet{saxe2011random} reported surprisingly high
performance of \glspl{CNN} with random weights on \texttt{NORB} and \mbox{\texttt{CIFAR-10}} datasets. We argue that
these results may, to some extent, be attributed to the structure-preserving nature of embeddings realized by untrained
neural networks.

The argument used in the above example assumed that weights are initialized with Gaussian \gls{RP} matrices. However, it
also holds for other \gls{RP} schemes. This motivates us to explore the prospects of initializing the weights in neural
networks with more intricate random matrices. In particular, we investigate initialization of deep networks with
\gls{RP} matrices described in Section~\ref{sec:rp_schemes}, i.e., Achlioptas', Li's, \gls{SRHT} and Count Sketch. As
the Gaussian \gls{RP} initialization with proper normalization is equivalent to the reference initialization scheme we
omit it in this chapter.

\subsubsection{Random projection initialization in CNNs}

Convolutional neural networks are made from three types of layers: convolutional layers, pooling layers and
fully-connected layers. Following~\citep{huang2016deep} we only initialize weights in the convolutional layers. In all
reported experiments biases are initialized to zeros.

Consider a convolution kernel matrix of size: $c_\mathrm{out} \times c_\mathrm{in} \times k \times k$, where
$c_\mathrm{in}$ and $c_\mathrm{out}$ are the number of input and output planes, respectively, and $k$ is the kernel
size. In \gls{RP}-based initialization we use $n = c_\mathrm{in} \times k \times k$ as the data dimensionality and
$c_{out}$ as the projection dimensionality. Therefore, before training, the convolutional layer can be seen as
performing an \gls{RP} of the input volume into a $c_{out}$-dimensional space. We normalize the kernel matrices to the
same standard deviation as in He's initialization, i.e., to~$\sqrt{2/n}$. The only exception is the Count Sketch
initialization, where we do not normalize the weights to a fixed standard deviation, but instead, multiply them with a
scale factor~$\gamma$ chosen with validation experiments. This difference is due to the sparsity of the Count Sketch
projection matrix -- each of its rows is a randomly chosen standard basis vector multiplied by either $1$ or $-1$.
Normalizing such matrix to a fixed standard deviation alters only its non-zero entries, causing their absolute values to
grow to huge numbers. This can impede the learning process or cause the gradients to explode. In \gls{SRHT}
initialization we set the sparsity parameter to $q = n^{-1}\log^2N$. For the number of projected examples~$N$ we use the
number of training images that are fed through the network during a single epoch. We have also experimented with using a
transposed initialization scheme and not employing weight scaling. However, these tests yielded worse results.

\subsubsection{Random projection initialization in pretrained networks}

In addition to \glspl{CNN}, we also investigate initialization in fully-connected networks with generative pretraining.
For the pretraining phase we employ stacked \glspl{RBM}~\citep{hinton2006reducing}. \Acrlong{RP} initialization is
performed before the pretraining phase, and thus serves as the starting point for the \gls{CD} algorithm. When
initializing the \gls{RBM} weights with an \gls{RP} matrix we use the number of visible units as the data dimensionality
and the number of hidden units as the projection dimensionality. Similarly to \gls{CNN} initialization, we leave the
biases set to zero. A standard advice when training an \gls{RBM} is to start with small
weights~\citep{hinton2012practical}. Therefore, after initialization we normalize the weight matrices to zero mean and a
small standard deviation $s = 0.01$. The Count Sketch initialization is, again, an exception: because of its sparsity,
we scale the weights by a constant factor chosen with validation experiments. In \gls{SRHT} initialization we set the
sparsity parameter~$q$ in the same way as in \gls{CNN} initialization. We also tested the network performance without
weight normalization. However, these experiments yielded significantly worse results, and we do not report them here.

\section{Experiments}

We evaluated the \gls{RP} initialization schemes on several popular image and text datasets, namely \texttt{MNIST},
\texttt{NORB}, \mbox{\texttt{CIFAR-10}}, \mbox{\texttt{CIFAR-100}}, \texttt{SVHN}, \texttt{TNG} and \texttt{RCV1}. For
details of these datasets and their preprocessing see Appendix~\ref{cha:datasets}. We carried out experiments on two
important machine learning tasks, namely image classification and document retrieval.

We begin by evaluating \gls{RP} initialization in deep \acrlongpl{CNN}. We then follow up with a similar evaluation for
networks pretrained using stacked \glspl{RBM}.

\subsection{Image classification with convolutional neural networks}

For the evaluation of \gls{RP} initialization in \glspl{CNN}, we employed three datasets, namely
\mbox{\texttt{CIFAR-10}}, \mbox{\texttt{CIFAR-100}} and \texttt{SVHN}. The evaluation was carried out using
\glspl{ResNet} with stochastic depth~\citep{huang2016deep}. Note, however, that \gls{RP} initialization proposed herein
is not tailored to this specific network architecture, but can be used in any rectifier network. Our goal in this
evaluation is not to improve the performance beyond the current state-of-the-art results on \texttt{CIFAR} or
\texttt{SVHN}, but instead to find out whether modern \gls{CNN} architectures can benefit from \gls{RP} initialization.
We use \glspl{ResNet} with stochastic depth as an example of such modern architecture. We used the best-performing
architectures and hyperparameter sets from~\citep{huang2016deep}, i.e., 110-layer \gls{ResNet} for the \texttt{CIFAR}
experiments and a 152-layer \gls{ResNet} for the \texttt{SVHN} experiments. With the reference He's
initialization~\citep{he2015delving} these models achieved state-of-the-art performance on \mbox{\texttt{CIFAR-10}} and
\mbox{\texttt{CIFAR-100}} with standard data augmentation and the second best published result on \texttt{SVHN}.

To account for the random nature of weight initialization, we trained ten network instances for each initialization
method, using different random number generator seeds. Afterward, we carried out statistical tests to assess the
confidence of \gls{RP} initialization schemes outperforming the He's initialization. Specifically, for each network
instance, we averaged the test errors from the last~$100$ (for \texttt{CIFAR}) or~$10$ (for \texttt{SVHN}) epochs and
compared the averages obtained with the He's initialization against averages obtained using \gls{RP} initialization. For
this comparison, we employed the Wilcoxon-Mann-Whitney two-sample rank-sum test. Note that comparing neural network
performance with a statistical test is not a common practice. However, a simple comparison of the early-stopping errors
would be inconclusive in this evaluation. By performing a statistical test we are able to more reliably assess the
performance of the evaluated weight initialization schemes. All experiments were carried out using the implementation of
\glspl{ResNet} with stochastic depth by Yu Sun\footnote{Available at
\url{https://github.com/yueatsprograms/Stochastic_Depth}}.

\setlength{\tabcolsep}{0.0em} % for the horizontal padding
\renewcommand{\arraystretch}{0}
\begin{figure}[!htb]
\begin{tabular}{ccc}

  \texttt{CIFAR-10} & \texttt{CIFAR-100} & \texttt{SVHN} \\
  \vspace{0.5em} \\

  \includegraphics[width=0.333\textwidth]{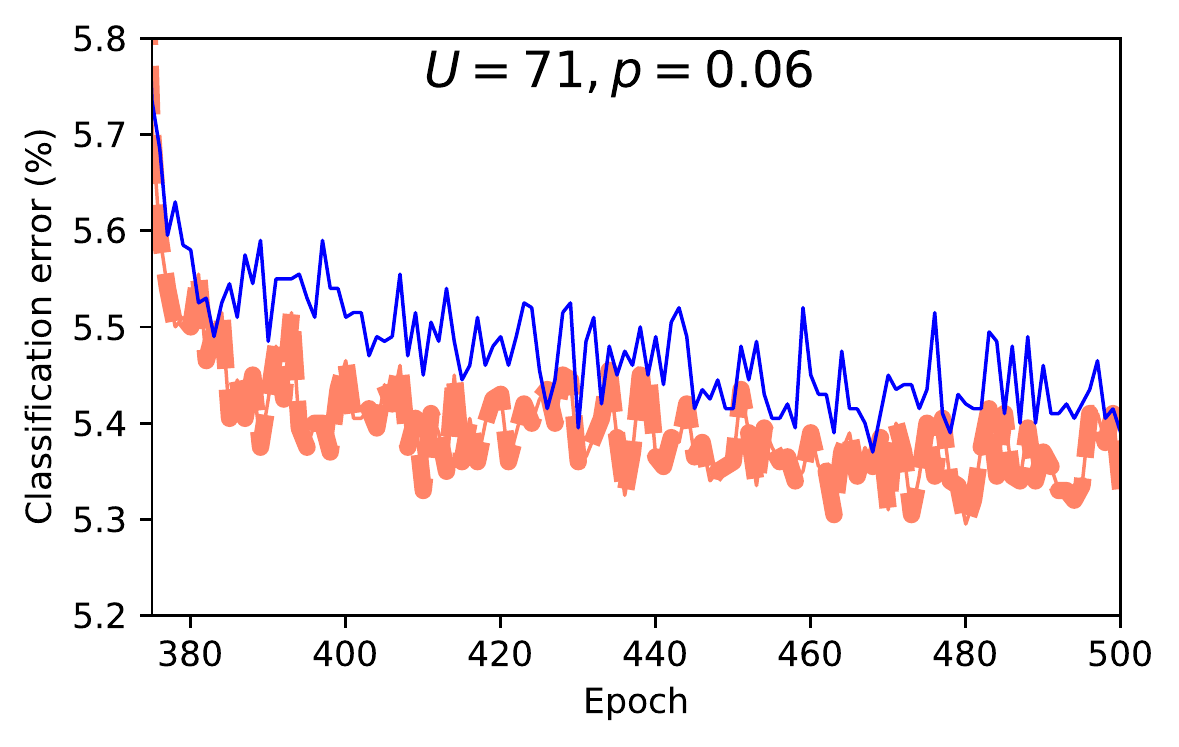} &
  \includegraphics[width=0.333\textwidth]{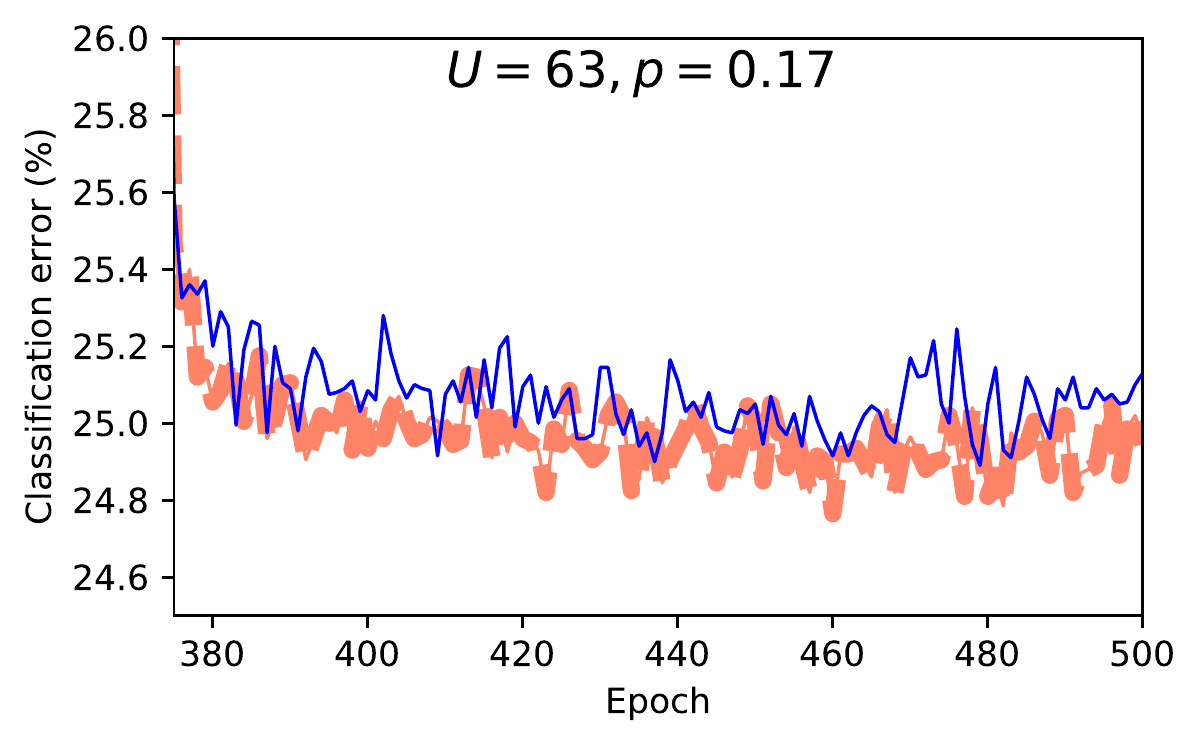} &
  \includegraphics[width=0.333\textwidth]{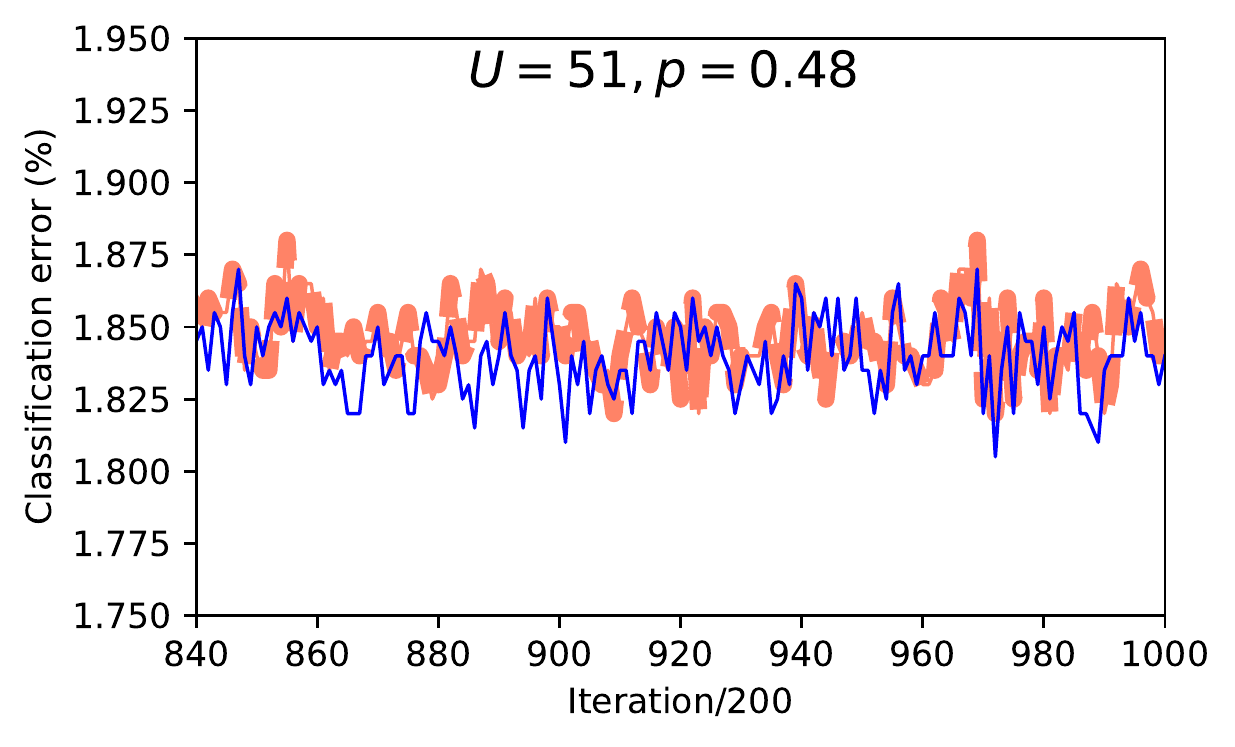} \\
  \multicolumn{3}{c}{Achlioptas' initialization}\\
  \vspace{1.22em} \\

  \includegraphics[width=0.333\textwidth]{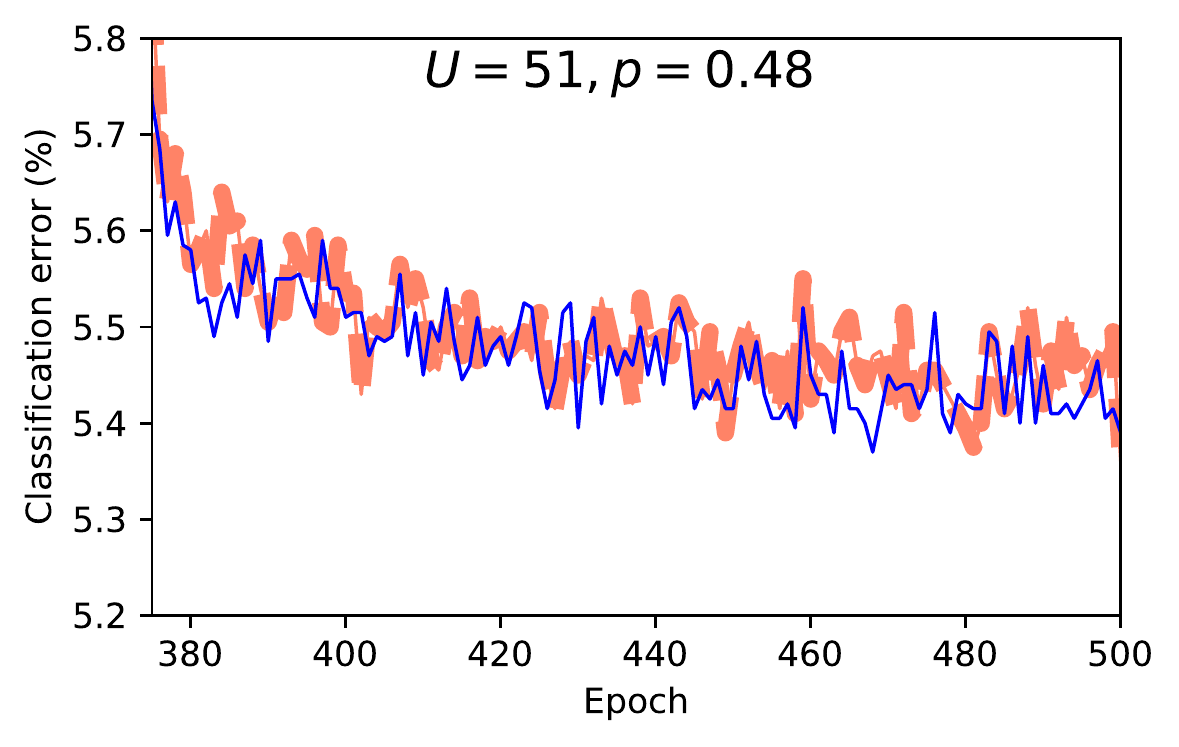} &
  \includegraphics[width=0.333\textwidth]{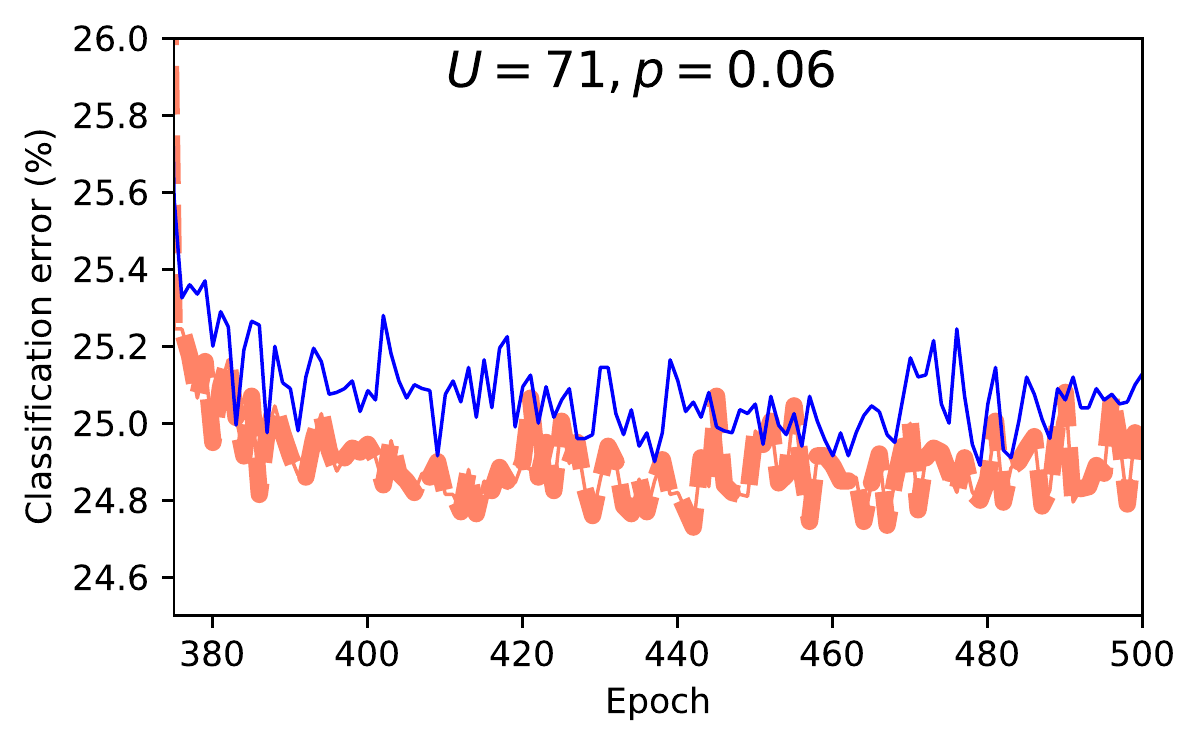} &
  \includegraphics[width=0.333\textwidth]{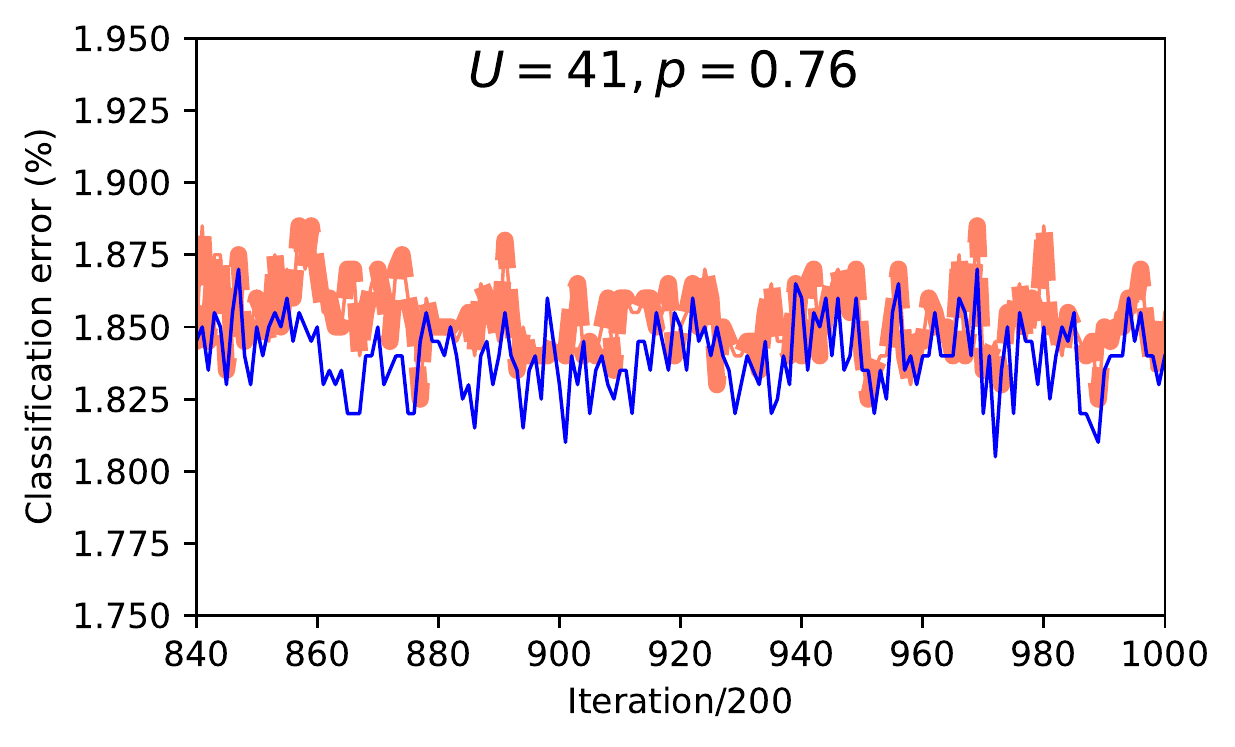} \\
  \multicolumn{3}{c}{Li's initialization}\\
  \vspace{1.22em} \\

  \includegraphics[width=0.333\textwidth]{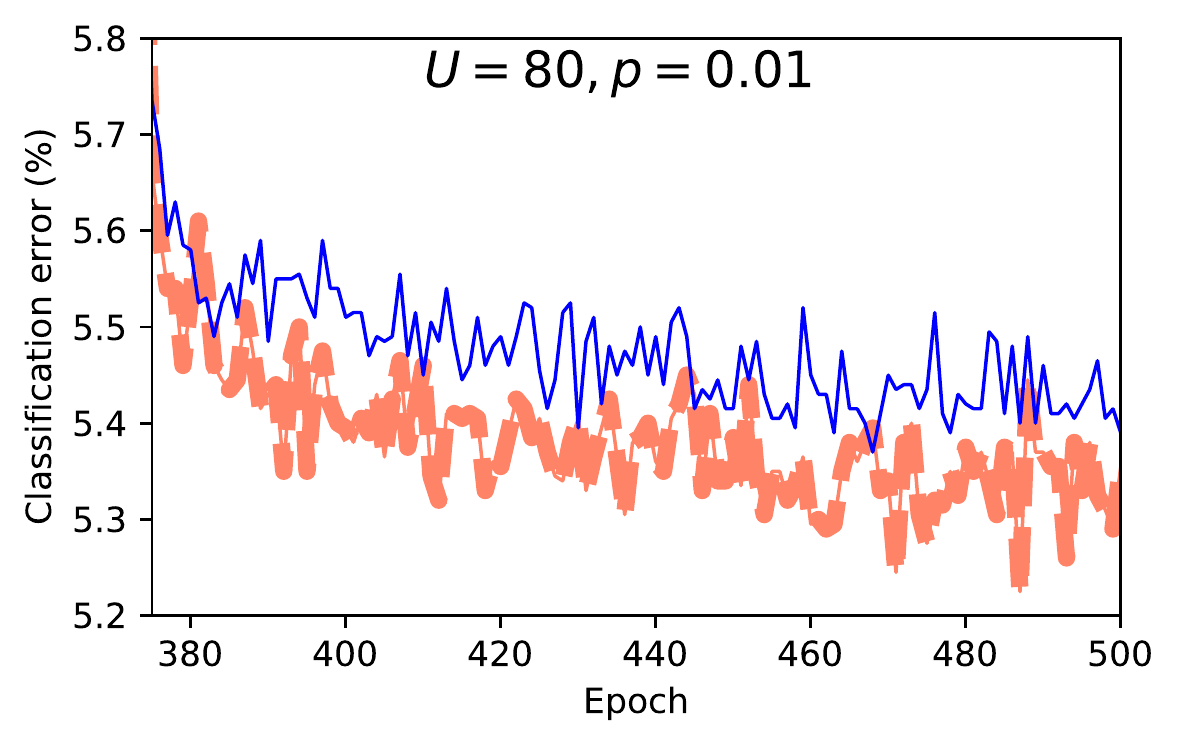} &
  \includegraphics[width=0.333\textwidth]{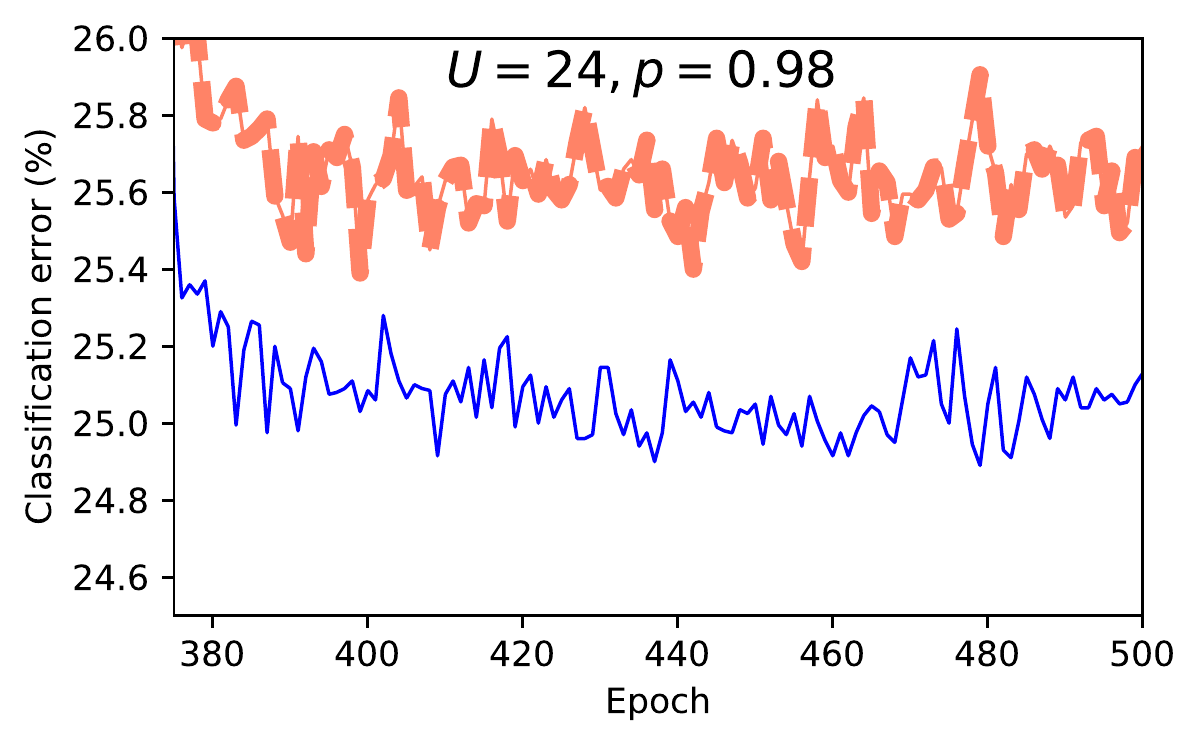} &
  \includegraphics[width=0.333\textwidth]{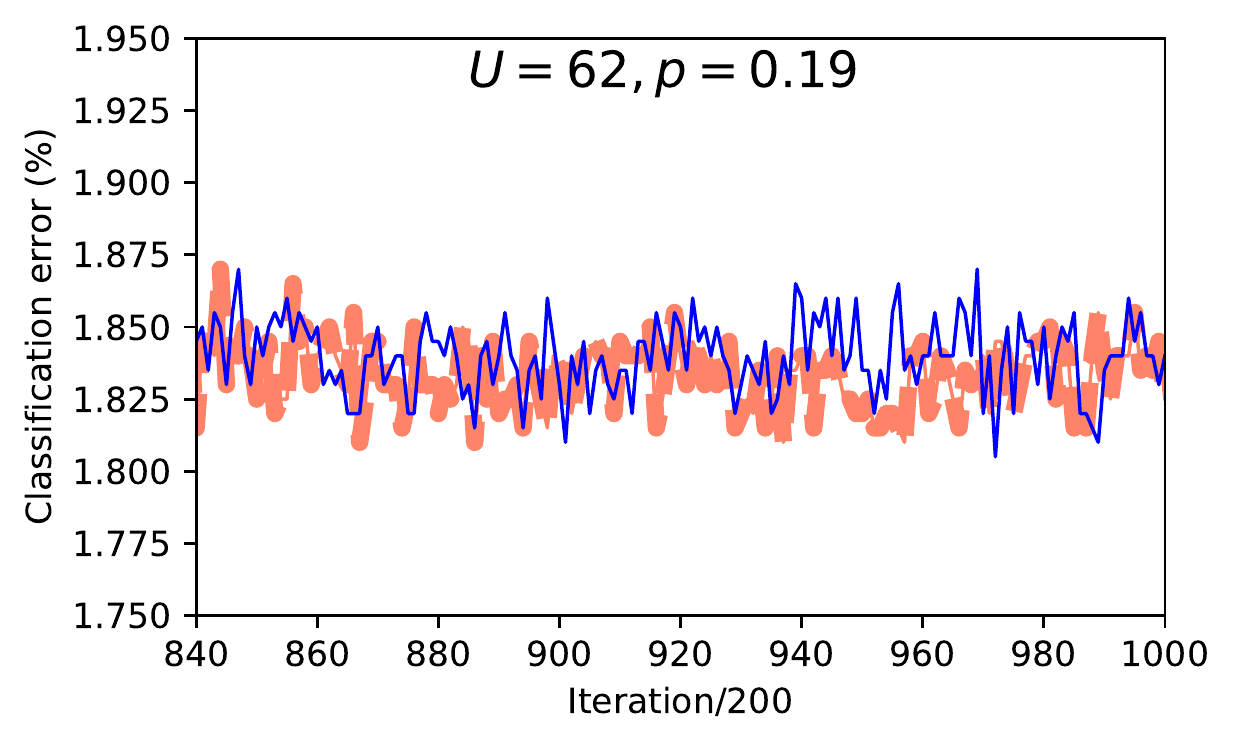} \\
  \multicolumn{3}{c}{Count Sketch initialization}\\
  \vspace{1.22em} \\

  \includegraphics[width=0.333\textwidth]{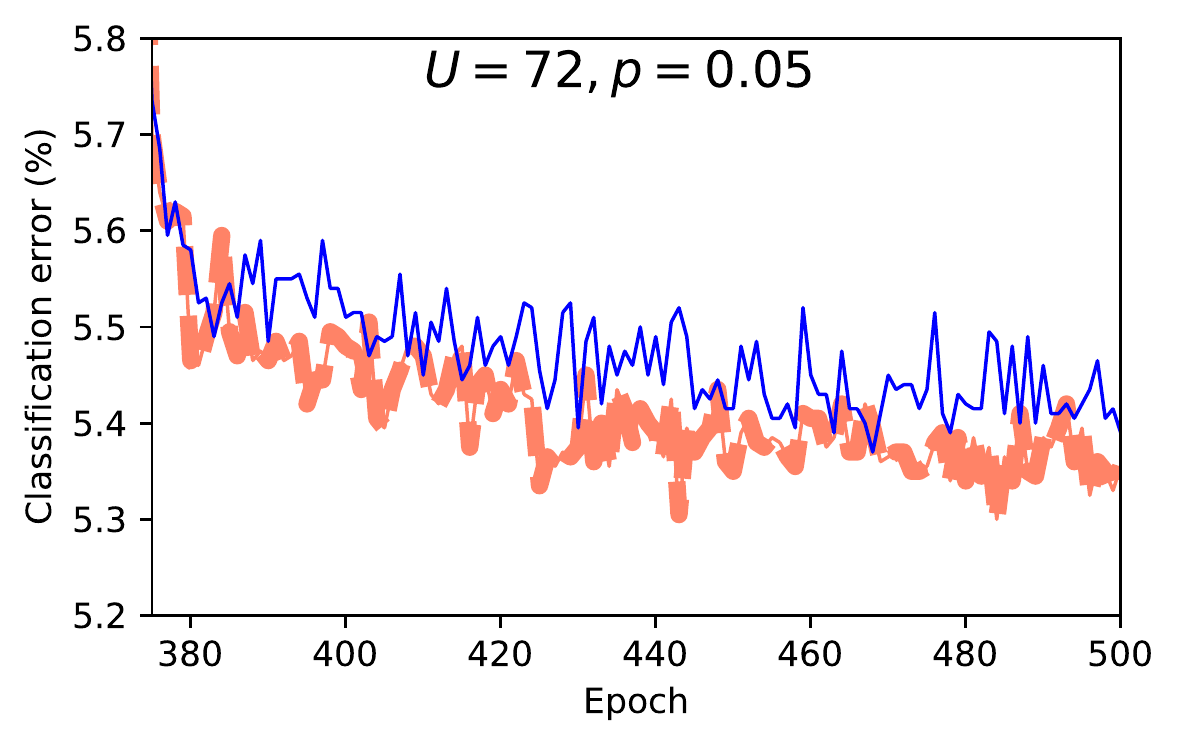} &
  \includegraphics[width=0.333\textwidth]{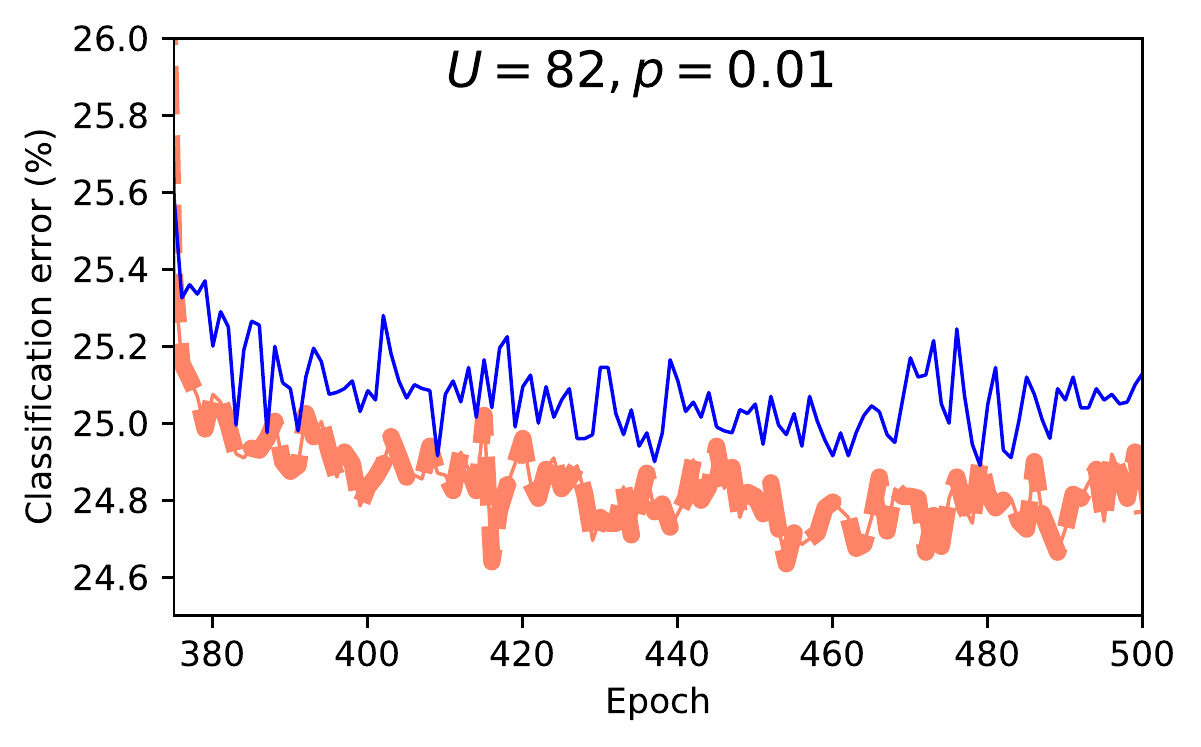} &
  \includegraphics[width=0.333\textwidth]{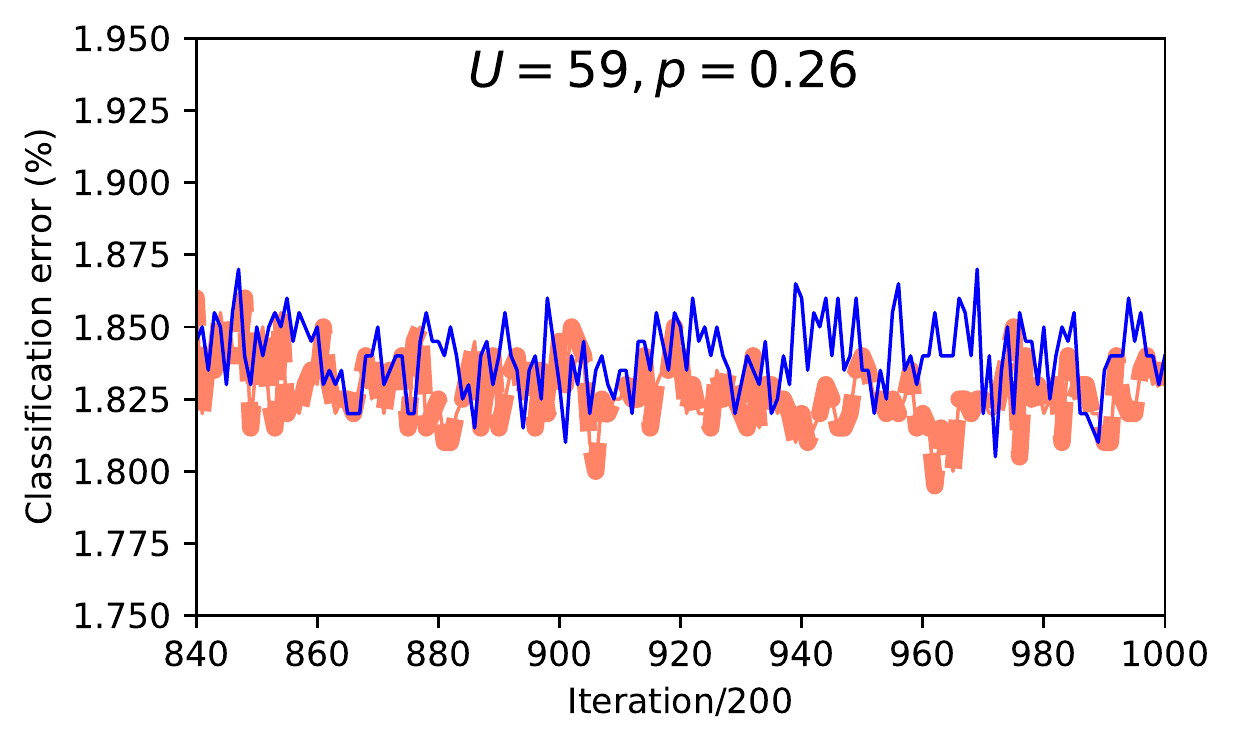} \\
  \multicolumn{3}{c}{SRHT initialization} \\
  \vspace{1em}
\end{tabular}
\caption{RP initialization in convolutional neural networks. Plots show median test error on \mbox{\texttt{CIFAR-10}},
\mbox{\texttt{CIFAR-100}} and \texttt{SVHN} for different RP initializations (dashed orange line) and the reference He's
initialization (solid blue line).}
\label{tab:rp_init_cifar}
\end{figure}

Performance of the evaluated initialization schemes on the \texttt{CIFAR} and \texttt{SVHN} datasets is reported in
Fig.~\ref{tab:rp_init_cifar}. The plots report median accuracy on the test sets, the significance level of the
hypothesis that the evaluated \gls{RP} initialization outperforms He's initialization and the value of the $U$ statistic
in the Wilcoxon-Mann-Whitney test. Following~\citep{huang2016deep}, we report the test error in each epoch for
\mbox{\texttt{CIFAR}} and after every~$200$ iterations for the \texttt{SVHN} experiments. We present the test errors
after setting the learning rate to the smallest value.

The \gls{SRHT} initialization outperformed the reference initialization on all datasets. Note that this is the only
dense \gls{RP} initialization scheme among the four evaluated methods; the other three schemes, i.e., Li's, Achlioptas'
and Count Sketch are sparse. The second best-performing \gls{RP} initialization scheme, i.e, Achlioptas', yielded good
but not statistically significant results on \texttt{CIFAR} and average results on \texttt{SVHN}. With one-third
non-zero entries, it is the second most dense initialization scheme evaluated in this work. Li's initialization achieved
a similar performance level: competitive on \texttt{CIFAR} but slightly worse than the reference on \texttt{SVHN}. The
Count Sketch scheme was the best performing \gls{RP} initialization method on \mbox{\texttt{CIFAR-10}} while yielding
results comparable to reference on \texttt{SVHN} and significantly worse than reference on \mbox{\texttt{CIFAR-100}}.
This method proved to be very sensitive to the scaling factor~$\gamma$, which led to inconsistent performance. Depending
on the dataset, it performed best with $\gamma \in \{0.1, 0.3\}$. The values of $\gamma$ higher than $1.0$ caused the
network to not converge, while values smaller than $0.1$ led to poor performance.

Overall our results suggest that the sparsity of the initialization scheme plays the deciding role in \glspl{CNN}
initialization: \glspl{CNN} perform best when their weights are densely initialized, as is the case for He's and
\gls{SRHT} initializations. The second important factor is the orthogonality of the initialized weights: the slightly
closer to orthogonal \gls{SRHT} initialization performs better than He's initialization. This finding is also supported
by experiments with orthogonal initialization in deep \glspl{CNN} by \citet{mishkin2015all}.

\subsection{Image classification with pretrained networks}

To evaluate \gls{RP} initialization in pretrained networks, we experimented on \texttt{MNIST} and Jittered-Cluttered
\texttt{NORB} datasets. For the \texttt{MNIST} experiments, we employed one of the network architectures
from~\citep{srivastava2013improving}, namely a binary input layer followed by two hidden layers with 1000 \gls{ReLU}
units and a~10-way softmax. For the \texttt{NORB} experiments, we used the best performing network architecture reported
in~\citep{nair2010rectified}, i.e., two hidden layers with 4000 and 2000 \gls{ReLU} units, respectively, followed by
a~6-way softmax. Inputs in this network were modeled with Gaussian units. Evaluated networks were pretrained with the
\gls{CD}\textsubscript{1} algorithm and finetuned for 500 epochs with error backpropagation. We trained the networks
using mini-batch \gls{SGD} with momentum. To avoid overfitting during pretraining we used L2 weight decay in all layers.
During finetuning we regularized the networks with dropout and decreased the learning rate according to a slow
exponential decay while slowly increasing the momentum value. Learning hyperparameters and the scaling factor for the
Count Sketch initialization were selected with experiments on the validation sets.

\begin{figure}[!htb]
\begin{tabular}{cc}
  \texttt{MNIST} & \texttt{NORB} \\
  \vspace{0.5em} \\

  \includegraphics[width=0.49\textwidth]{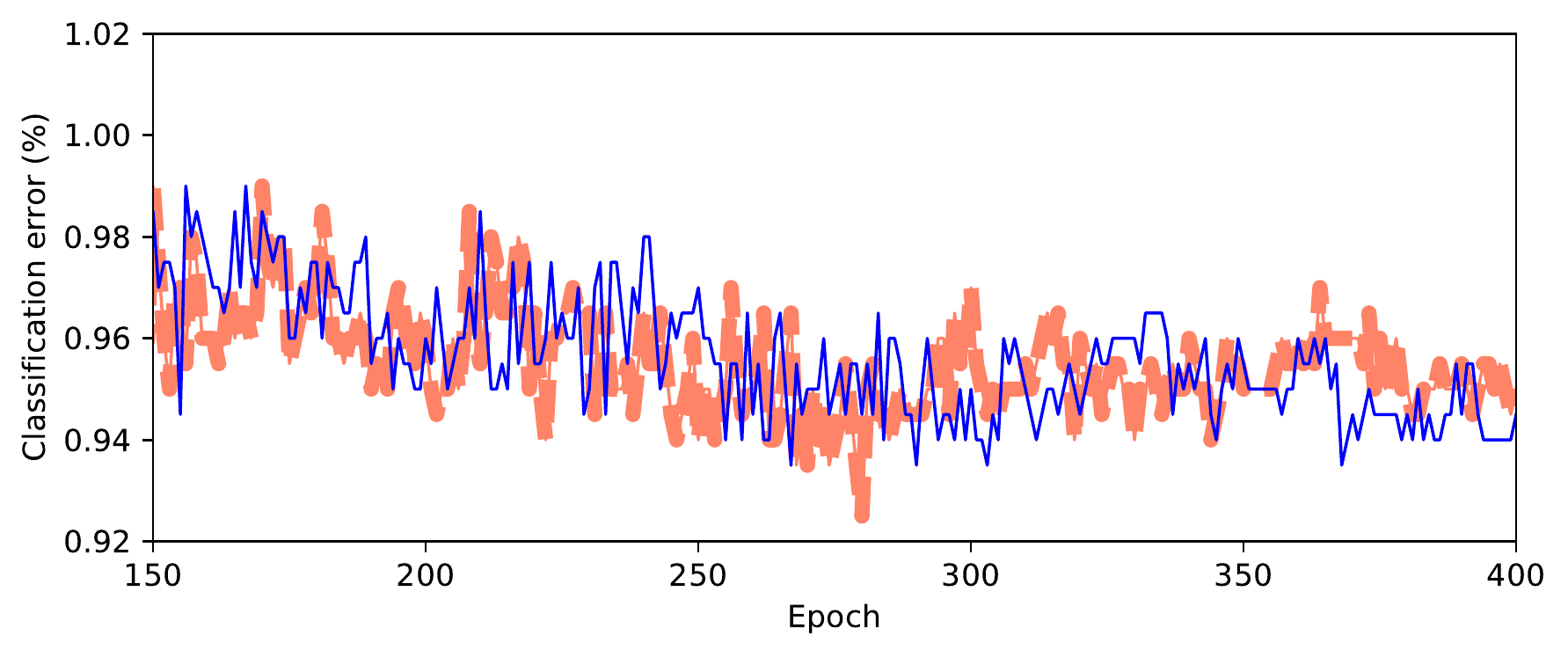} &
  \includegraphics[width=0.49\textwidth]{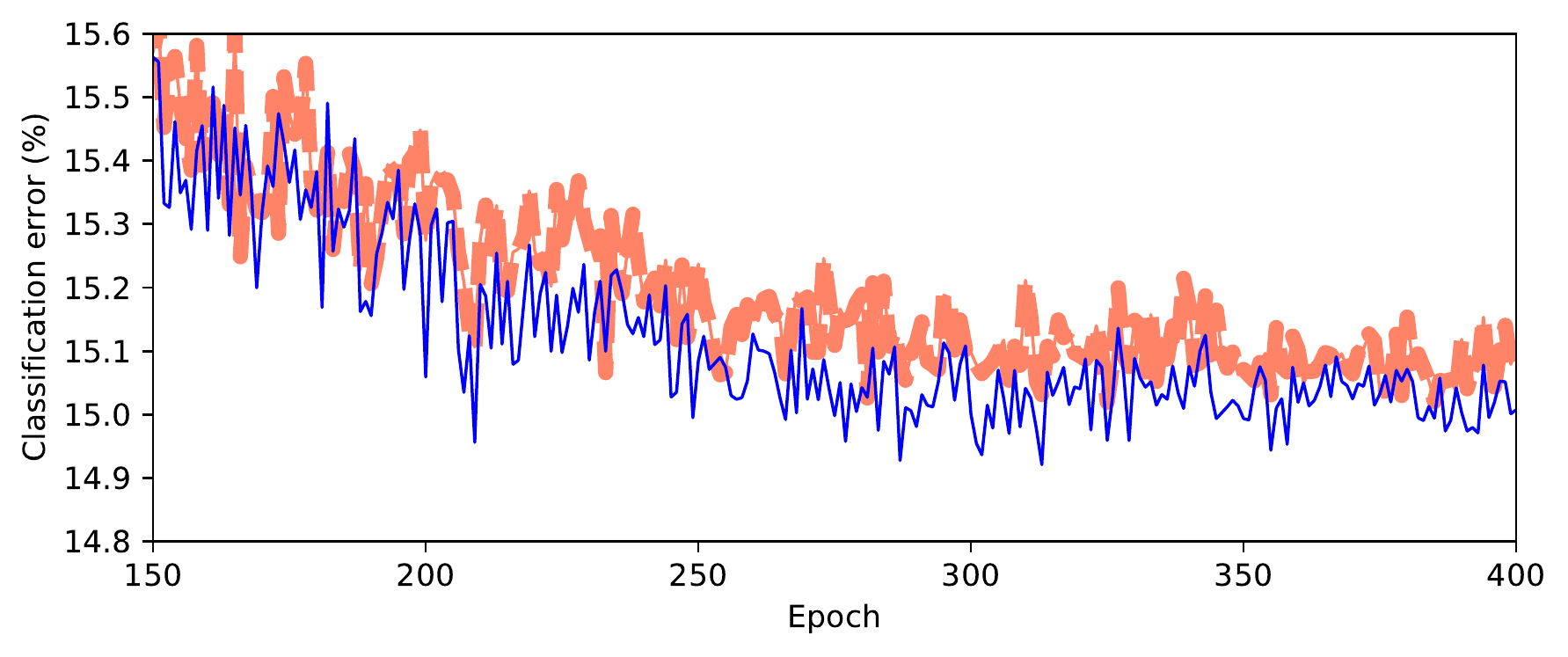} \\
  \multicolumn{2}{c}{Achlioptas' initialization} \\
  \vspace{1.22em} \\

  \includegraphics[width=0.49\textwidth]{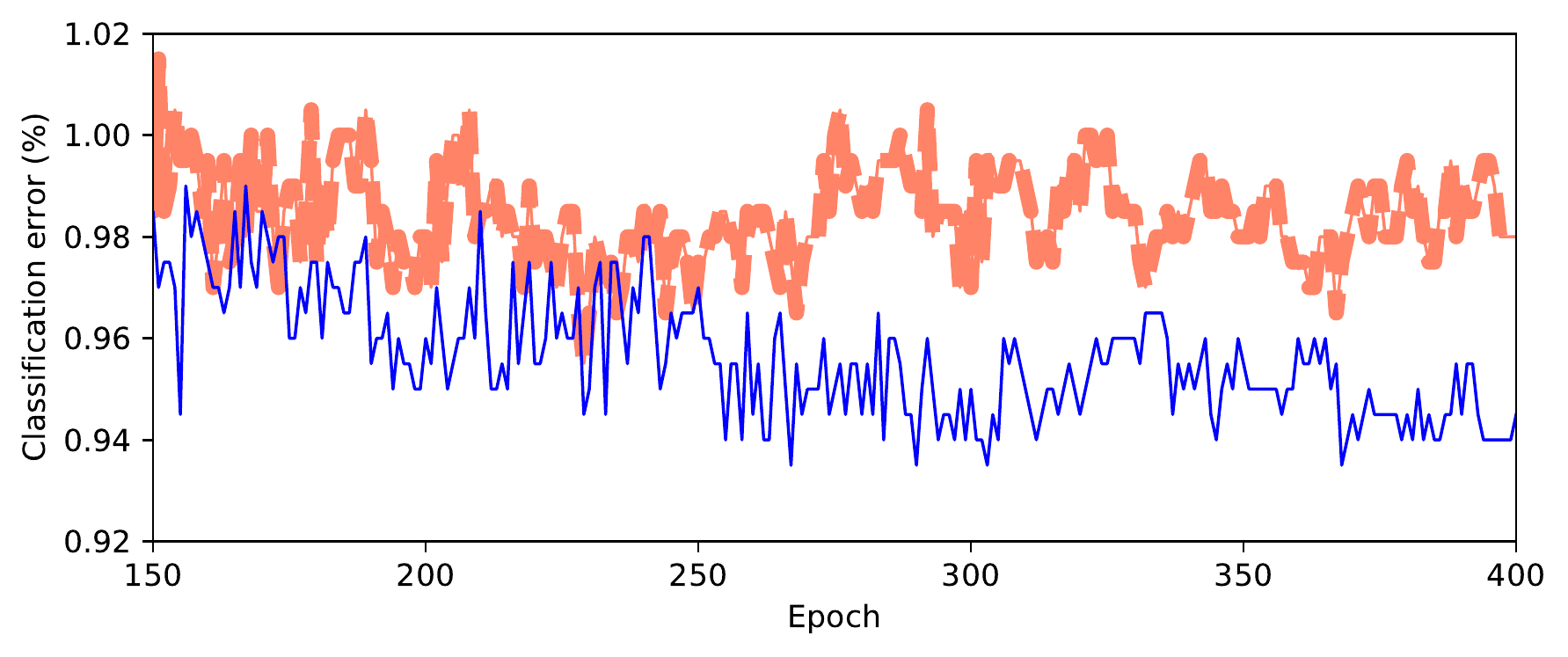} &
  \includegraphics[width=0.49\textwidth]{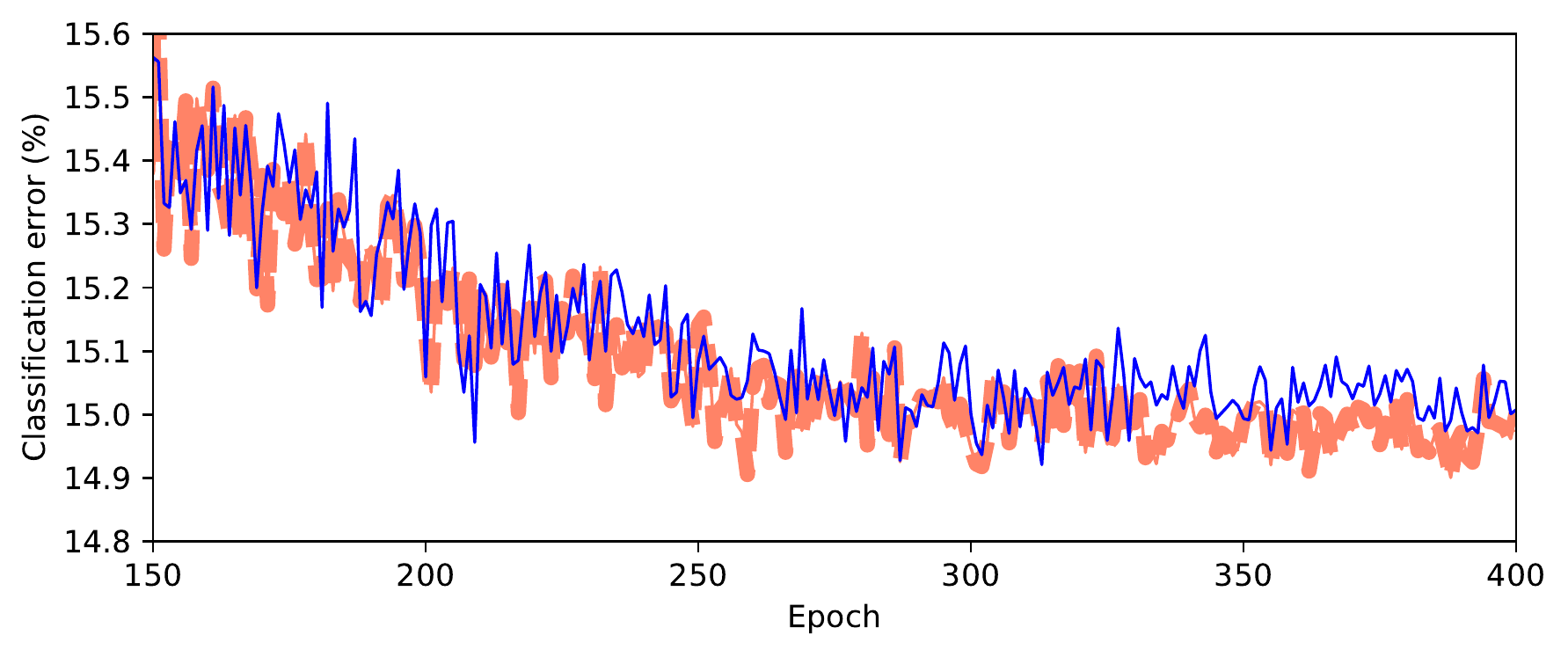} \\
  \multicolumn{2}{c}{Li's initialization}\\
  \vspace{1.22em} \\

  \includegraphics[width=0.49\textwidth]{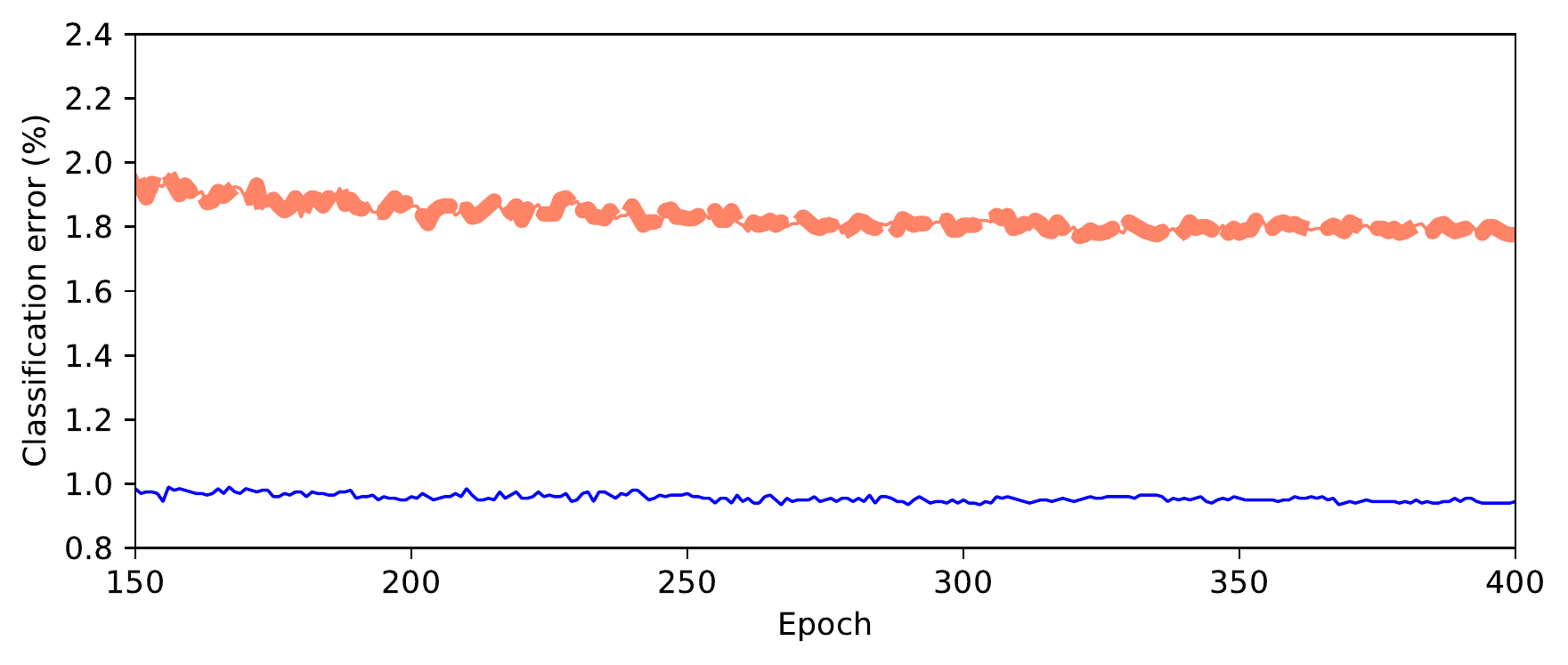} &
  \includegraphics[width=0.49\textwidth]{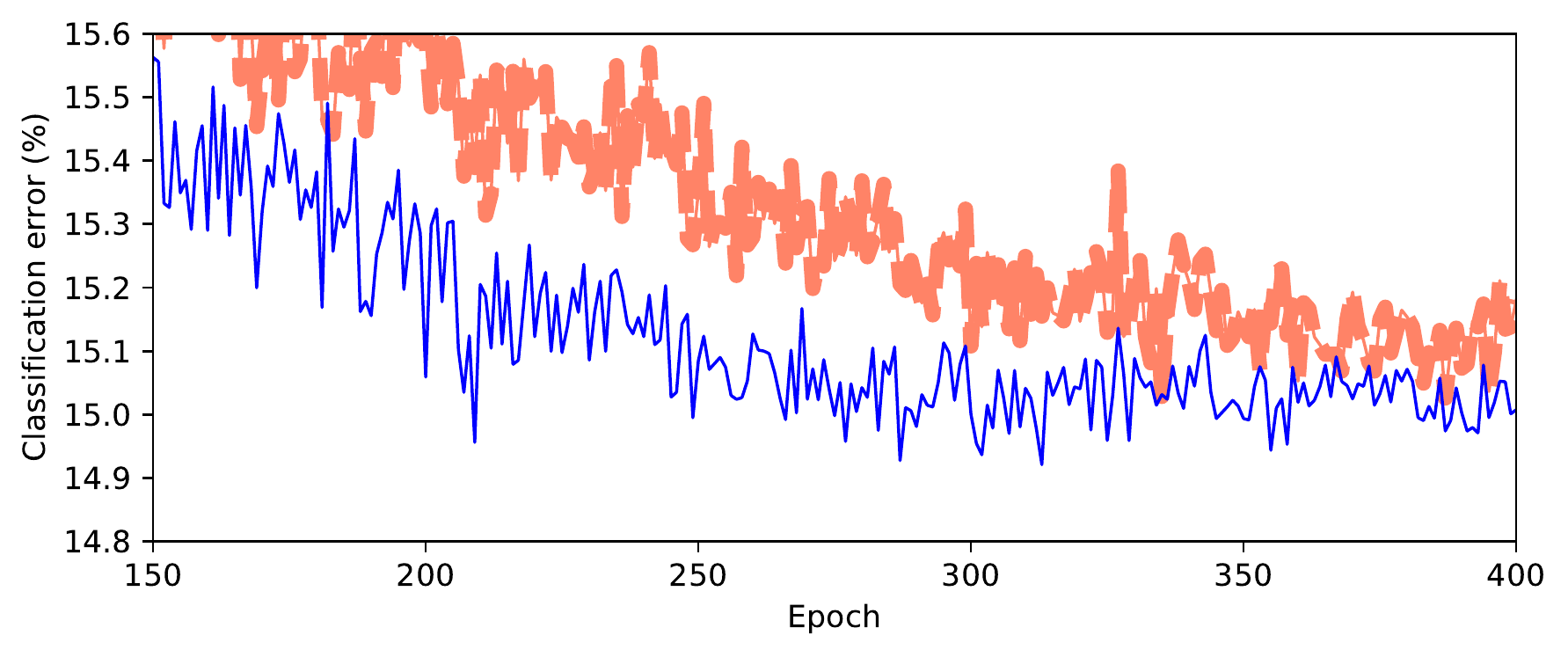} \\
  \multicolumn{2}{c}{Count Sketch initialization}\\
  \vspace{1.22em} \\

  \includegraphics[width=0.49\textwidth]{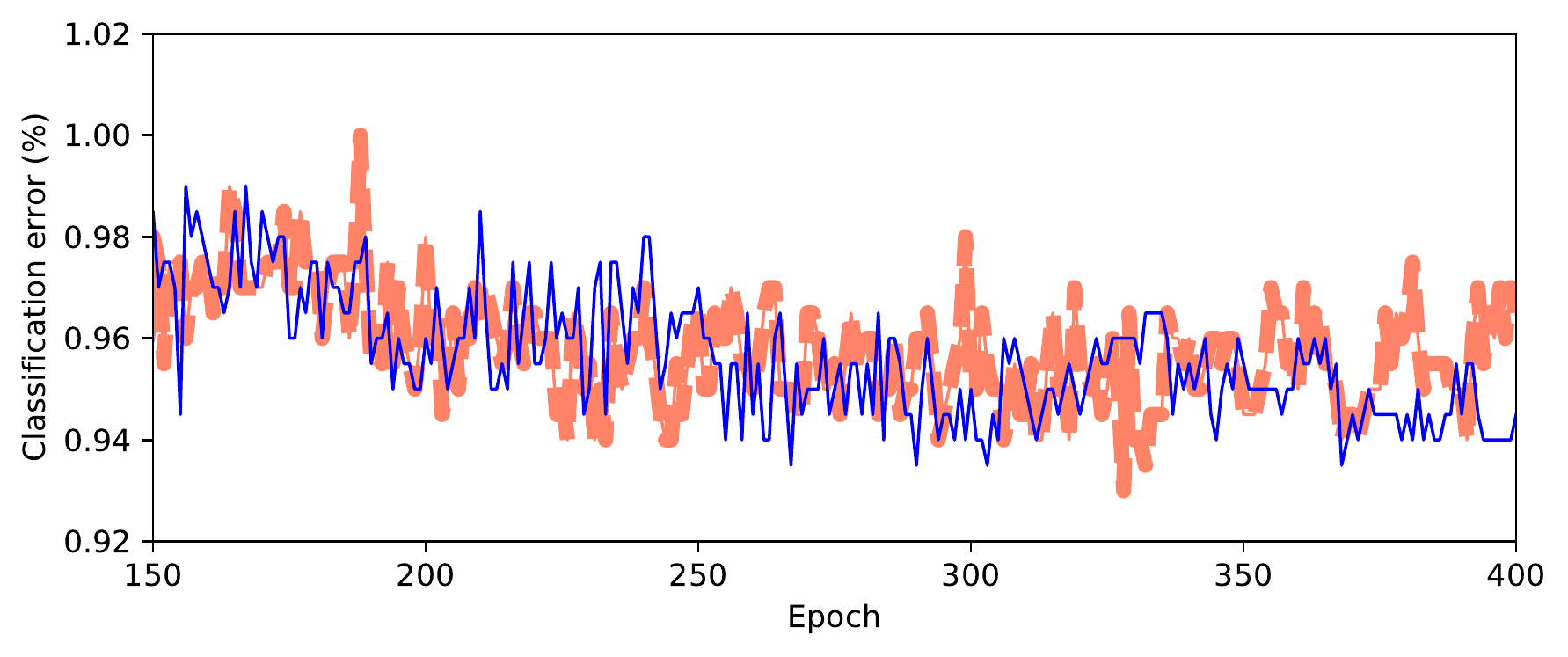} &
  \includegraphics[width=0.49\textwidth]{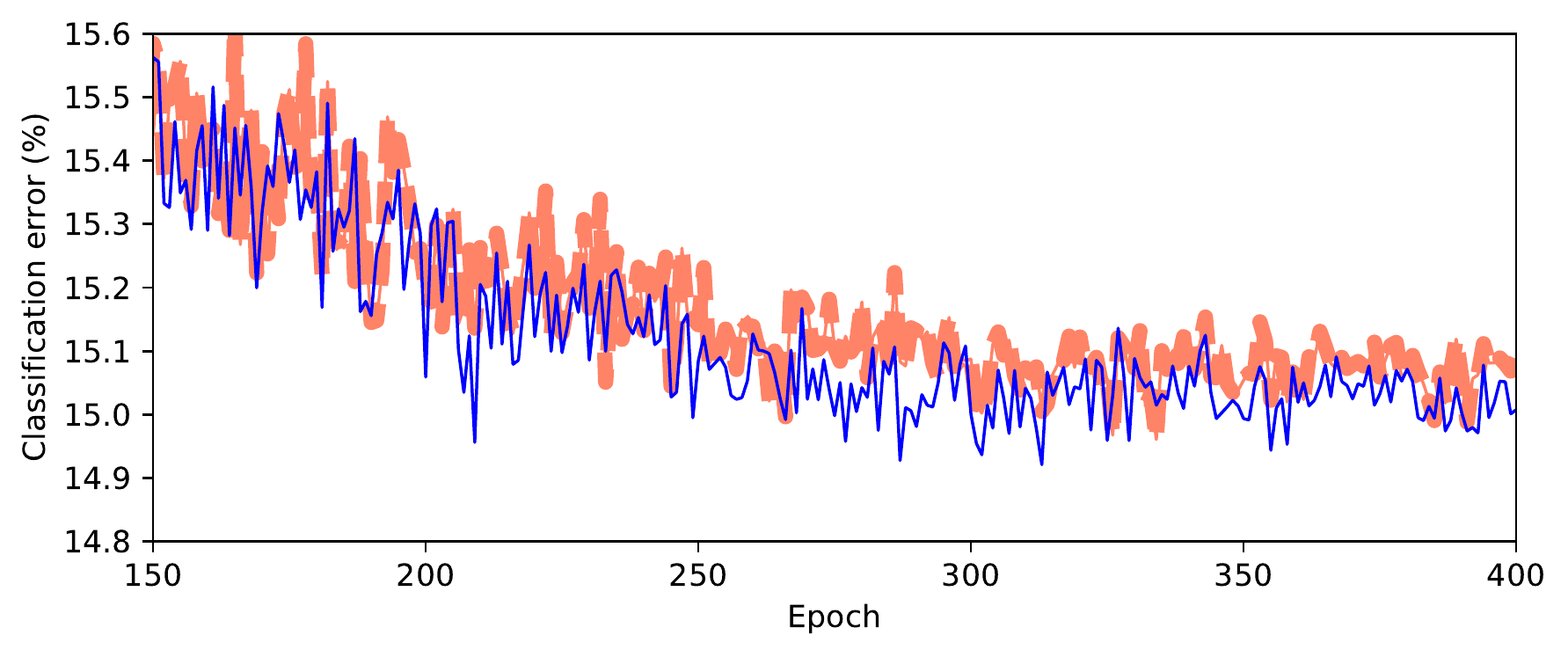} \\
  \multicolumn{2}{c}{SRHT initialization}\\
  \vspace{1em}
\end{tabular}
\caption{RP initialization in pretrained networks. Plots show median test error on \texttt{MNIST} and \texttt{NORB}
datasets for different RP initializations (dashed orange line) and the reference Gaussian initialization (solid blue
line).}
\label{tab:rp_init_mnist}
\end{figure}

For each dataset and initialization scheme, we trained ten network instances with different random number seeds. Results
from these experiments are reported in Fig.~\ref{tab:rp_init_mnist}. In each case, we report median test error as a
function of the finetuning epoch. The standard Gaussian initialization serves as the baseline result.

Unlike \glspl{CNN}, image classification with pretrained networks does not benefit from \gls{RP} initialization.
Specifically, Achlioptas' and \gls{SRHT} initialization yielded slightly worse results, compared to the reference
initialization. Li's initialization performed better on the \texttt{NORB} dataset but worse on \texttt{MNIST}. The Count
Sketch initialization yielded results significantly worse than reference, which can be attributed to its sparsity.

\subsection{Document retrieval with autoencoders}

\begin{figure}[!htb]
\begin{tabular}{cc}
  \includegraphics[width=0.5\textwidth]{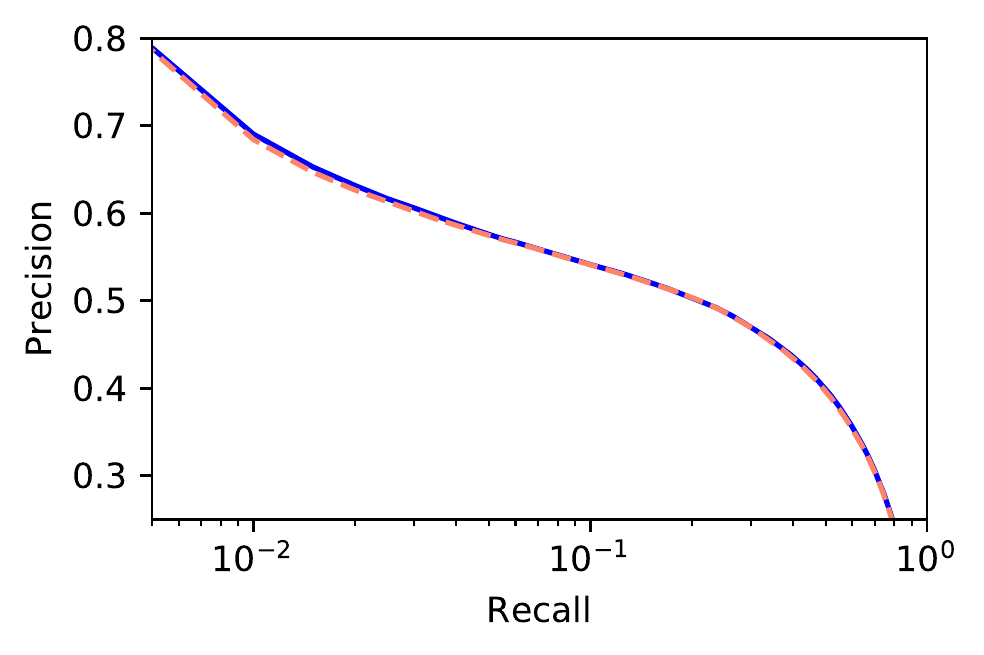} &
  \includegraphics[width=0.5\textwidth]{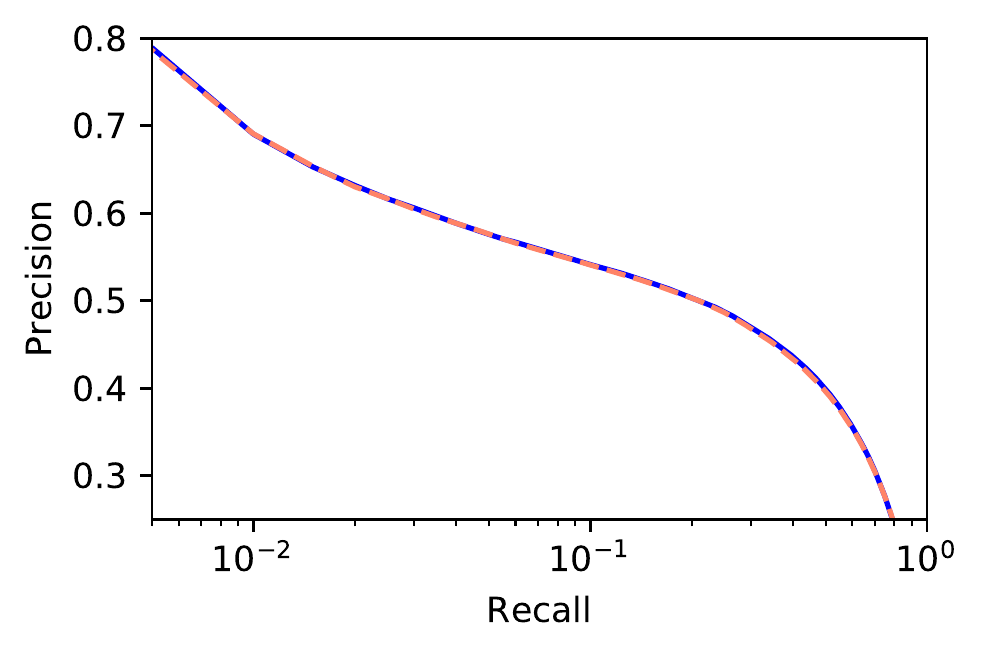} \\
  Achlioptas' initialization & Li's initialization \\
  \vspace{1.22em} \\

  \includegraphics[width=0.5\textwidth]{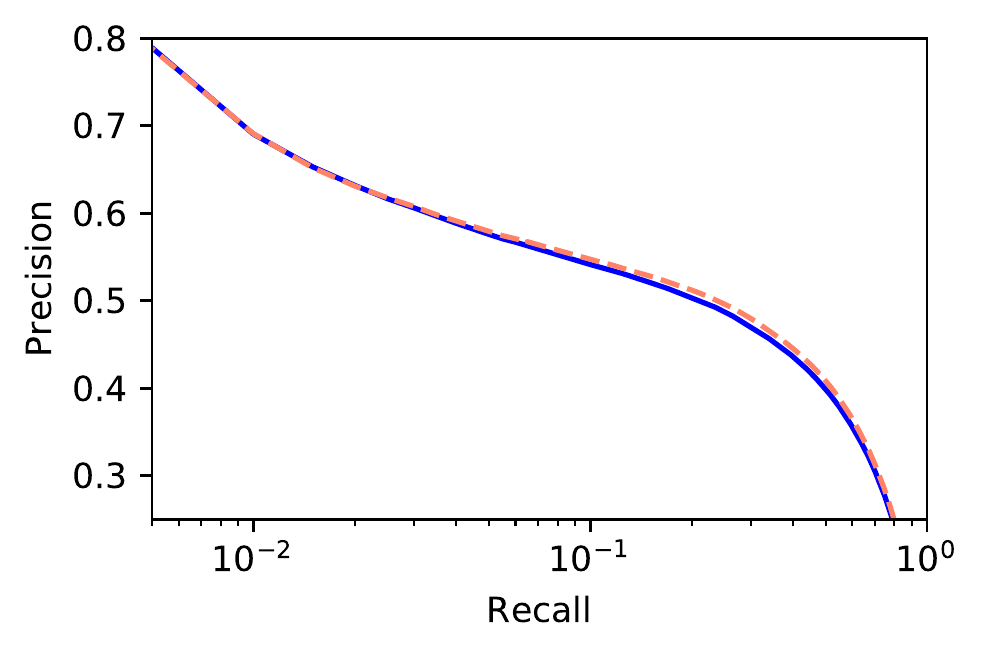} &
  \includegraphics[width=0.5\textwidth]{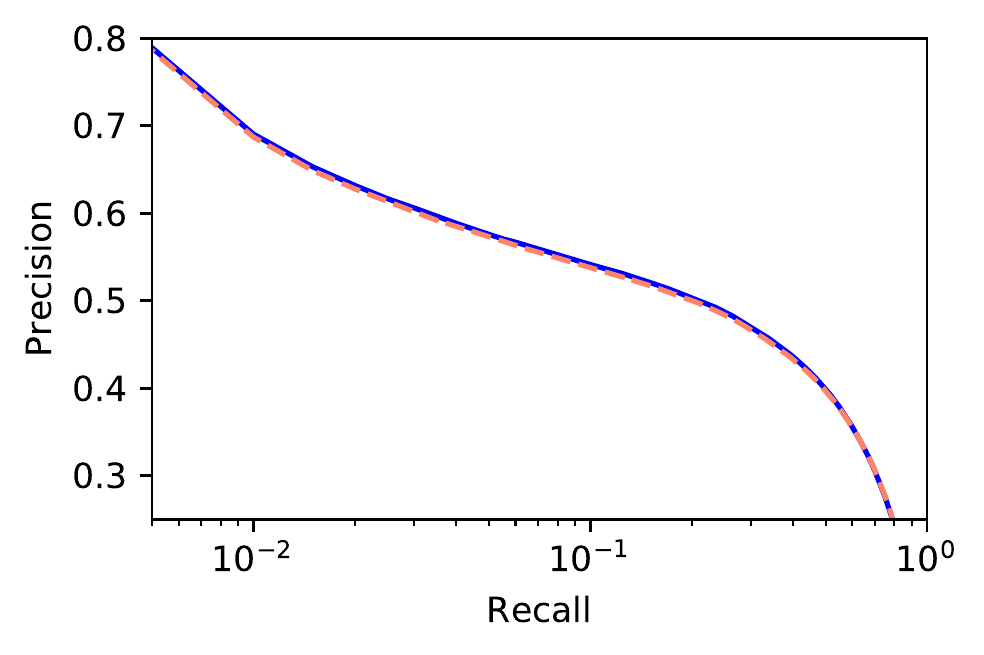} \\
  Count Sketch initialization & SRHT initialization \\
  \vspace{1em}
\end{tabular}
\caption{RP initialization in autoencoders. Plots show the precision-recall curves for results with median AUC on the
\texttt{TNG} dataset for different RP initializations (dashed orange line) and the reference Gaussian initialization
(solid blue line).}
\label{fig:rp_init_tng}
\end{figure}

\begin{figure}[!htb]
\begin{tabular}{cc}
  \includegraphics[width=0.5\textwidth]{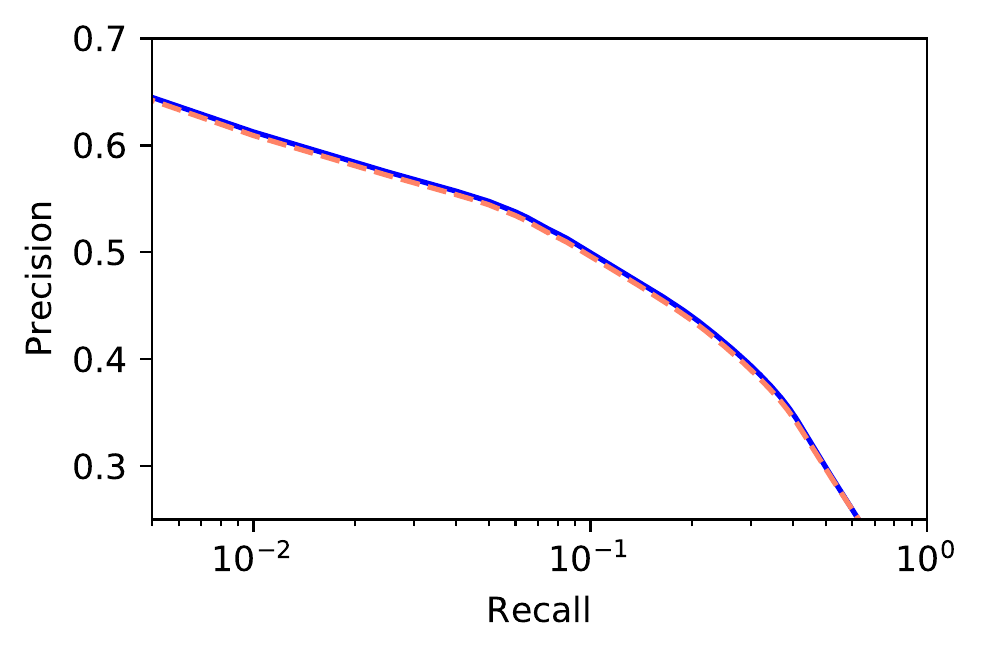} &
  \includegraphics[width=0.5\textwidth]{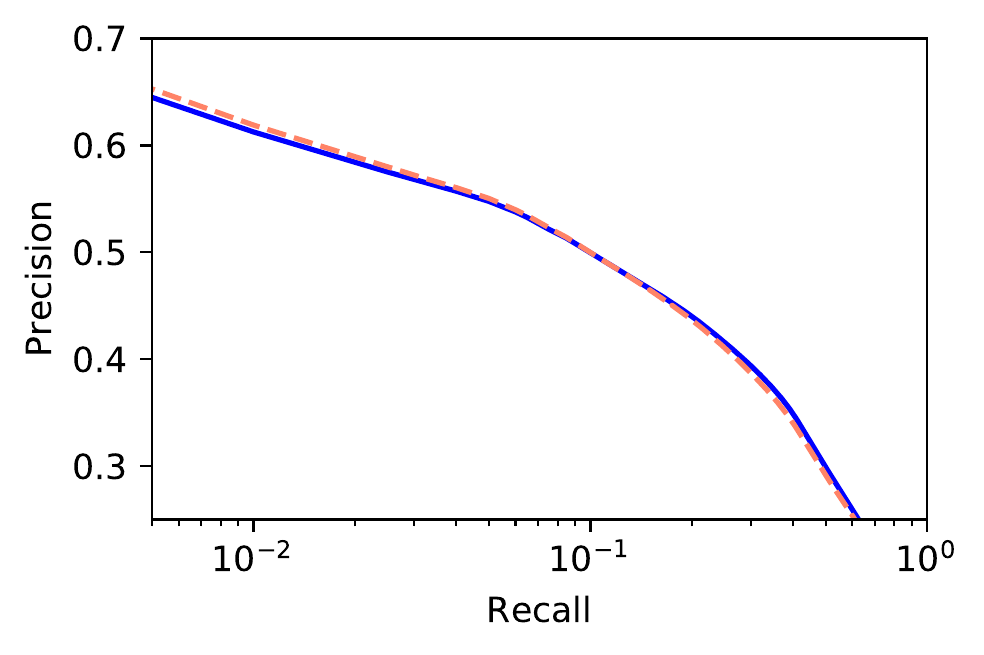} \\
  Achlioptas' initialization & Li's initialization \\
  \vspace{1.22em} \\

  \includegraphics[width=0.5\textwidth]{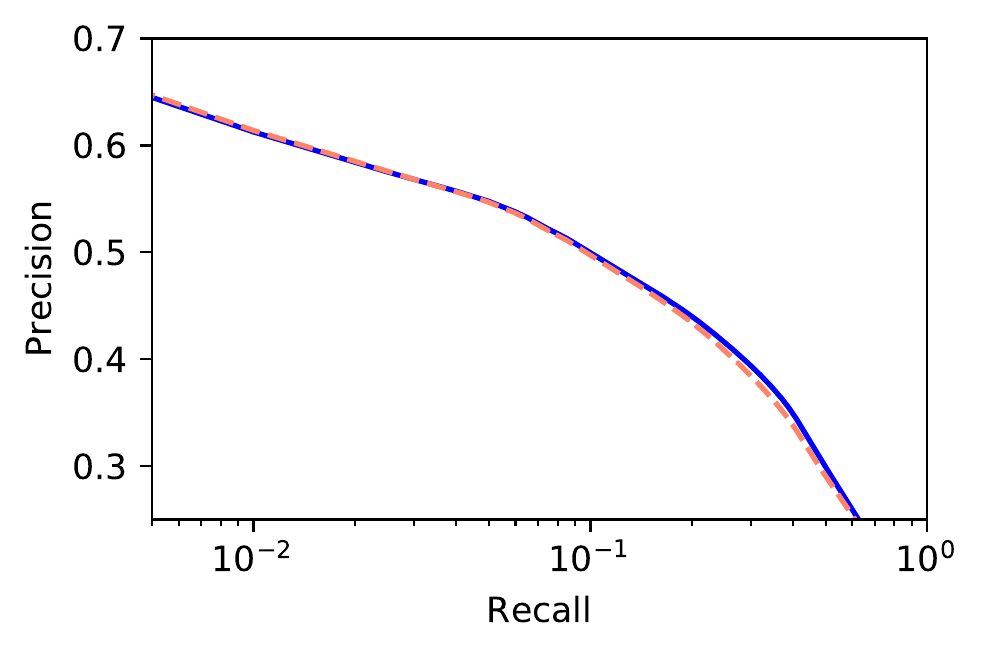} &
  \includegraphics[width=0.5\textwidth]{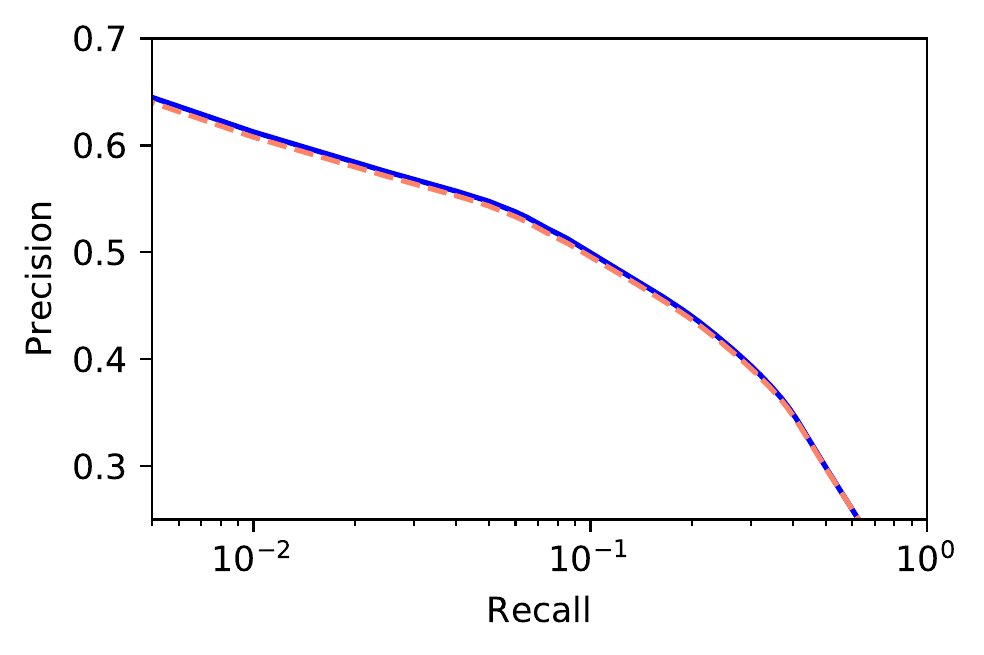} \\
  Count Sketch initialization & SRHT initialization \\
  \vspace{1em}
\end{tabular}
\caption{RP initialization in autoencoders. Plots show the precision-recall curves for results with median AUC on the
\texttt{RCV1} dataset for different RP initializations (dashed orange line) and the reference Gaussian initialization
(solid blue line).}
\label{fig:rp_init_rcv1}
\end{figure}

In the previous sections we presented the results of applying \gls{RP} initialization in image classification task. In
this section we evaluate \gls{RP} initialization in networks trained for the document retrieval task. In particular, we
carry out experiments with \gls{RP} initialization in deep autoencoders trained on the \texttt{TNG} and \texttt{RCV1}
corpus. As a baseline we use deep autoencoder architectures and training regime from~\citep{grzegorczyk2016encouraging}.
This is the same baseline as in Section~\ref{sec:rp_bow}. In reference \glspl{DBN} all layers were initialized with
small random numbers from a Gaussian distribution. We compare these baseline networks to \glspl{DBN} in which weights
were initialized with different \gls{RP} matrices. In this initialization we chose the scaling factor for Count Sketch
with experiments on the validation sets ($\gamma = 0.3$). We used the document codes inferred with the autoencoders in a
document retrieval task, similarly to the evaluation in Section~\ref{sec:rp_bow}. We use \gls{AUC} to compare the
performance of the trained networks. Similarly to the previous experiments with \gls{RP} initialization, for each
dataset and \gls{RP} initialization scheme we trained ten network instances with different random number seeds. In
Table~\ref{tab:rp_init_auc} we report median \gls{AUC} values for different initialization schemes. In
Fig.~\ref{fig:rp_init_tng} and Fig.~\ref{fig:rp_init_rcv1} we present the precision-recall curves for autoencoders with
the median value of \gls{AUC}. Each curve for a \gls{RP}-initialized autoencoder (plotted in dashed orange line) is
juxtaposed with the curve for the reference Gaussian-initialized autoencoder (plotted in solid blue line).

\renewcommand{\arraystretch}{1.1}
\setlength{\tabcolsep}{9pt}
\begin{table*}[htb]
  \caption{Median area under the precision-recall curve for deep autoencoders initialized with RP matrices.}
  \label{tab:rp_init_auc}
  \centering
  \begin{tabular}{rccc}
  \toprule
    Weight initialization scheme & \phantom{a} & \multicolumn{2}{c}{Dataset} \\ \cmidrule{3-4}
                     && \texttt{TNG} & \texttt{RCV1} \\ \midrule
    Gaussian (reference) && 0.373 & \textbf{0.315} \\
    Achlioptas'      && 0.372 & 0.314 \\
    Li's             && 0.373 & 0.311 \\
    SRHT             && 0.372 & 0.314 \\
    Count Sketch     && \textbf{0.381} & 0.310 \\
    \bottomrule
  \end{tabular}
\end{table*}

\renewcommand{\arraystretch}{0}
\setlength{\tabcolsep}{0pt}

\renewcommand{\arraystretch}{0}

\setlength{\tabcolsep}{0.5em} % for the horizontal padding
\renewcommand{\arraystretch}{1.1}

In general, \gls{RP} initialization had little effect on the document retrieval performance. Only for the \texttt{TNG}
dataset the Count Sketch initialization yielded slightly better network performance compared to the standard Gaussian
initialization.

\section{Conclusions}

In this chapter we explored the viability of initializing deep networks with different \gls{RP} matrices. We motivated
why \gls{RP} matrices that satisfy the Johnson–Lindenstrauss lemma may serve as good initial weights in deep networks.
We then experimentally evaluated the performance of modern networks initialized with \gls{RP} matrices. Specifically, we
experimented with \glspl{CNN} and with pretrained networks using Achlioptas', Li's, \gls{SRHT} and Count Sketch \gls{RP}
matrices.

Our results show that dense orthogonal \gls{RP} initialization schemes can improve the performance of deep
\acrlongpl{CNN}. In particular, in our evaluation, the initialization based on \gls{SRHT} outperformed the reference
He's initialization in state-of-the-art \glspl{ResNet} with stochastic depth. Sparse \gls{RP} initializations, i.e.,
Li's, Achlioptas' and Count Sketch, yielded results that were inconsistent among different benchmarks.

In pretrained networks \gls{RP} initialization usually yielded results close to the results obtained with the standard
Gaussian initialization. Only the Count Sketch initialization yielded significantly different results: it performed much
worse than the reference in image classification networks while in autoencoders used for document retrieval it performed
better. We argue that this poor performance in image classification networks is a consequence of too high sparsity of
the Count Sketch matrix. When initializing a fully-connected layer that has $d$ inputs and $r$ outputs with Count
Sketch, only $r$ elements are set to non-zero values. Therefore, the sparsity of the weight matrix is $\frac{1}{r}$. In
our evaluation of image classification networks, depending on the dataset, Count Sketch scheme initialized only~$0.1\%$
of all weights to non-zero values. While sparsely initialized pretrained networks have been shown to perform well
in~\citep{grzegorczyk2015effects}, the Martens' initialization used therein resulted in approximately~\mbox{10-times}
denser initial weight matrices. The autoencoder networks in our experiments were built with much smaller layers compared
to the image classification networks. As a result, the weight matrices in autoencoders were initialized more densely.

%% file: conclusions.tex
\cleardoublepage
\chapter{Conclusions}
\label{cha:conclusions}

In this work we studied areas in which \acrlongpl{DNN} can benefit from \acrlong{RP}. We started by reviewing the
challenges to training machine learning models on extremely high-dimensional data and by discussing the existing
approaches. We focused on a particularly difficult type of data -- data that is represented by millions of features, is
sparse and lacks a structure that could be exploited to simplify the learning model. We discussed why efficiently
training \glspl{DNN} on such type of data is a challenging task and how this challenge can be overcome by incorporating
a novel type of layer into the network architecture. Specifically, we propose to extend the network architecture with a
layer that incorporates a \acrlong{RP} operation. We consider two variants of the proposed \gls{RP} layer: one in which
its weights are fixed and one where they are learned during training. We found that training the weights in the \gls{RP}
layer, although computationally much more expensive, can significantly improve the overall network performance. We
proposed several modifications to the network architecture and the training regime that enabled efficient training
\glspl{DNN} with learnable \gls{RP} layer on data with as many as tens of millions of input features and examples.
Specifically:
\begin{itemize}
    \item we initialize the \gls{RP} layer weights with a sparse \acrlong{RP} scheme,
    \item we finetune only these weights in the \gls{RP} that were initialized to non-zero values, i.e., we employ
    sparse connectivity in the \gls{RP} layer,
    \item we batch normalize the activations of the \gls{RP} layer,
    \item we use linear activation function after the batch normalization,
    \item we update weights in the \gls{RP} layer only for a fraction of the training mini-batches.
\end{itemize}
We conducted an evaluation of \glspl{DNN} with the \gls{RP} layer on several large-scale synthetic and real-world
datasets. The evaluation showed that our approach is not only viable but also competitive in terms of performance with
the state-of-the-art techniques. In particular, incorporating \gls{RP} into the \gls{DNN} architecture allowed us to
improve the state-of-the-art classification error by over 30\% on two real-world benchmarks: \texttt{url} and
\texttt{webspam}. These results open a path to applying neural networks in tasks where directly learning from the data
was previously infeasible because of the overly high dimensionality of input examples. The main limitation of our
approach is that it is computationally more expensive than classic methods, such as linear \glspl{SVM}. However, with an
already trained model, the inference time of \glspl{DNN} with the \gls{RP} layer is small: feeding a training example
through the \gls{RP} layer can be realized with a single matrix multiplication. By using fast \gls{RP} schemes, this
operation can be performed in linear or nearly linear time. The transformations in subsequent layers can be implemented
efficiently on modern hardware, e.g., on \glspl{GPGPU}. Therefore, despite the high computational cost of training,
neural networks with \gls{RP} layer can be used to solve practical problems. We also found that \acrlong{RP} is useful
for initialization of weights in \glspl{DNN}. Specifically, we propose to initialize the weights in \glspl{DNN} with
scaled \gls{RP} matrices. This approach yielded deep \gls{CNN} models that perform better than networks initialized with
the current state-of-the-art He's method. Together, our results fully prove the thesis stated in the introduction:
\acrlong{RP} can be beneficial for training \acrlongpl{DNN} by enabling \glspl{DNN} to learn from sparse, unstructured,
high-dimensional data and by improving the network initialization.

In our evaluation, we tested five \gls{RP} matrix constructions: Gaussian, Achlioptas', Li's, \gls{SRHT} and Count
Sketch. Our experiments suggest that in neural network applications the crucial properties of an \gls{RP} construction
are its density and orthogonality. Specifically, in networks with fixed-weight \gls{RP} layer, orthogonal projection
schemes work best. For weight initialization the most successful schemes are the ones that are both orthogonal and
dense. Finally, sparse orthogonal schemes yield the best results in finetuned \gls{RP} layers. Out of the evaluated
schemes, the most useful for practical application are therefore the \gls{SRHT} and Count Sketch constructions. When
employed for training \glspl{DNN} with fixed-weight \gls{RP} layers, they combine the most efficient projection with the
best performance of the final models. Because of its sparsity, Count Sketch projection matrix is also suitable for
learnable \gls{RP} layer. Since networks with finetuned \gls{RP} layer outperform models with fixed-weight \acrlong{RP},
Count Sketch is the overall best \gls{RP} construction for \acrlongpl{DNN}. For weight initialization, the best
performing \gls{RP} scheme was \gls{SRHT}. In our experiments, it improved the performance of the state-of-the-art
\glspl{ResNet} on several benchmark datasets. All sparse \gls{RP} constructions performed poorly when used for weight
initialization.

In future work we plan to investigate novel \gls{RP} schemes proposed during the work on this thesis. In particular, we
are eager to explore applications in \glspl{DNN} of a new family of dense structured random matrices, which extends
constructions such as the circulant, Hankel or Toeplitz matrices~\citep{choromanski2016recycling}. This family, called
random ortho-matrices~(ROM)~\citep{choromanski2017unreasonable,felix2016orthogonal}, provides promising theoretical
guarantees on the embedding quality and, as their name suggests, are fully-orthogonal. These two properties should make
ROMs a perfect candidate for application in \glspl{DNN}, especially for weight initialization. We also plan to further
investigate training strategies for finetuned \gls{RP} layers and, especially, the feasibility of changing the
connectivity of neurons in the \gls{RP} layer during finetuning. In other words, we want to update not only the weights
that are initially non-zero, but also the weights that are initially set to zero but receive large gradients during
training. To prevent the number of learnable parameters from growing uncontrollably, the weights that consistently
receive small gradient can be removed from the set of updated weights. We hope that these approaches will further
improve performance of \glspl{DNN} applied to sparse, high-dimensional, unstructured data.

%% file: datasets.tex
\cleardoublepage
\chapter{Datasets}
\label{cha:datasets}

In this appendix we list and briefly describe the datasets used in this work. A summary of these datasets is given in
Table~\ref{tab:datasets}.

\setlength{\tabcolsep}{9pt}
\begin{table*}[htb]
  \caption{A summary of the datasets used in the conducted experiments. Density is the fraction of non-zero elements in
  the training set.}
  \label{tab:datasets}
  \centering
  \begin{tabular}{rrrrrc}
    \toprule
    Dataset            & \begin{tabular}{@{}c@{}} Training \\ set size \end{tabular} &
      \begin{tabular}{@{}c@{}} Test \\ set size \end{tabular} & Dimensionality & Classes & Density \\ \midrule
    \texttt{MNIST}     & $60,000$     & $10,000$  & $784$        & $10$  & $0.191$ \\
    \texttt{NORB}      & $291,600$    & $58,320$  & $2,048$      & $6$   & dense \\
    \texttt{CIFAR-10}  & $50,000$     & $10,000$  & $3,072$      & $10$  & dense \\
    \texttt{CIFAR-100} & $50,000$     & $10,000$  & $3,072$      & $100$ & dense \\
    \texttt{SVHN}      & $604,388$    & $26,032$  & $3,072$      & $10$  & dense \\
    \texttt{TNG}       & $11,314$     & $7,532$   & $2,000$      & $20$  & $0.034$ \\
    \texttt{RCV1}      & $402,207$    & $402,207$ & $2,000$      & $103$ & $0.035$ \\
    \texttt{url}       & $1,976,130$  & $420,000$ & $3,231,961$  & $2$   & $3.58 \cdot 10^{-5}$ \\ % (228964773 nonzeros)
    \texttt{webspam}   & $280,000$    & $70,000$  & $16,609,143$ & $2$   & $2.24 \cdot 10^{-4}$ \\
    \texttt{KDD2010-a} & $8,407,752$  & $510,302$ & $20,216,830$ & $2$   & $1.80 \cdot 10^{-6}$ \\ % (305613510 nonzeros)
    \texttt{KDD2010-b} & $19,264,097$ & $748,401$ & $29,890,095$ & $2$   & $9.84 \cdot 10^{-7}$ \\ % (566345888 nonzeros)
    synthetic          & $1,000,000$  & $250,000$ & $1,000,000$  & $2$   & 
      \begin{tabular}{@{}c@{}} from $10^{-6}$ \\ to $1.7 \cdot 10^{-4}$ \end{tabular} \\
    \bottomrule
  \end{tabular}
\end{table*}

\subsection*{MNIST}
\begin{figure}[htb]
  \centering
  \includegraphics[width=0.4\linewidth]{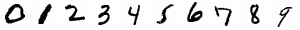}
  \caption{Example MNIST images.}
  \label{fig:mnist_example}
\end{figure}

The \texttt{MNIST} dataset~\citep{lecun1998gradient}\footnote{Available at \url{http://yann.lecun.com/exdb/mnist/}} is a
widely used benchmark for machine learning algorithms. It consists of images of handwritten
digits~(Figure~\ref{fig:mnist_example}).
In the original dataset the images are represented by~$256$ grey-scale levels, but in this work we use pixel intensities
rescaled to the~$[0, 1]$ interval. We use the permutation invariant version of the dataset, i.e., we randomly shuffle
the pixel order.

\subsection*{NORB}

\begin{figure}[htb]
  \centering
  \includegraphics[width=\linewidth]{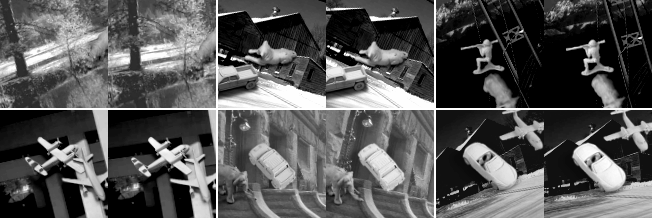}
  \caption{Example NORB images~(one from each of the six classes).}
  \label{fig:norb_example}
\end{figure}

The Jittered-Cluttered \texttt{NORB} dataset~\citep{lecun2004learning}\footnote{Available at
\url{https://cs.nyu.edu/~ylclab/data/norb-v1.0/}} consists of images depicting one of $50$ toys on diverse background
captured in stereo mode under variable lighting and viewpoints~(Figure~\ref{fig:norb_example}).
Following~\citet{nair2010rectified} we resized original images to $32 \times 64$ pixels, subtracted from each image its
mean pixel intensity, divided pixel intensities by the standard deviation of pixel intensities in the training set and
constructed a validation set consisting of $58,320$ cases from the training set.

\subsection*{20-newsgroups (TNG)}

The \texttt{TNG} dataset\footnote{Available at \url{http://qwone.com/~jason/20Newsgroups}} is a collection of posts from
20 Usenet newsgroups. Topics in \texttt{TNG} range from religion (e.g., \textit{talk.religion.misc, alt.atheism,
soc.religion.christian}) to computer hardware (e.g., \textit{comp.sys.ibm.pc.hardware, comp.sys.mac.hardware}).
Following \citet{salakhutdinov2009semantic} we preprocessed the corpus by removing the stop-words, stemming it and
constructing a \gls{BOW} representation using the most common words in the training set. For our experiments we created
several dataset variants with different vocabulary sizes, ranging from~$2000$ to~$10,000$. For the validation set we
use~3,000 documents randomly extracted from the training set.

\subsection*{Reuters Corpus Volume I (RCV1)}

Reuters Corpus Volume I~\citep{lewis2004rcv1} is an archive of over $800,000$ English newswire stories published between
August 1996 and August 1997. We use its corrected version RCV1-v2. Each newswire story in the corpus has been
categorized (multi-labeled) into 103 topics from four main groups: Corporate/Industrial, Economics, Government/Social
and Markets. Topics from each group form a hierarchical structure, typically with depth three. Following
\citet{salakhutdinov2009semantic} we define the relevance of two documents to be the fraction of their agreeing topics.
We apply the same train/test split and preprocessing scheme as \citet{salakhutdinov2009semantic}. Similarly to
\texttt{TNG} we experiment on \gls{BOW} dataset variants created over dictionaries with vocabulary sizes ranging
from~$2000$ to~$10,000$.

\subsection*{Malicious URL (url)}

\texttt{url}~\citep{ma2009identifying}\footnote{Available at \url{http://sysnet.ucsd.edu/projects/url/}} is a large
binary classification dataset often used for evaluating online learning methods. It consists of $3.2$M-dimensional
descriptors of $2.4$M URL addresses. Descriptors contain lexical features (\gls{BOW} representations of tokens in the
URL) and host-based features (WHOIS information, location, connection, speed, blacklist membership, etc.). The challenge
in this dataset is to recognize malicious addresses from benign addresses. Following \citet{wang2011trading} we use
examples from the first~$100$ days of data collection as the training set and remaining examples for testing.

\subsection*{KDD2010-a and KDD2010-b}

\texttt{KDD2010-a} and \texttt{KDD2010-b} are large student performance prediction datasets from the \textit{KDD Cup
2010} educational data mining competition. We use a preprocessed versions of these datasets made available by the
challenge winner~\citep{yu2010feature}\footnote{Available at
\url{https://www.csie.ntu.edu.tw/~cjlin/libsvmtools/datasets/}}. For validation we use random subsets of the training
set with the same size as the corresponding test sets.

\subsection*{webspam}

\texttt{webspam}~\citep{webb2006introducing} is a document dataset consisting of~$350,000$ descriptors of web pages,
which was originally used in Pascal Large Scale Learning Challenge~\citep{sonnenburg2008pascal}. The challenge in this
dataset is to detect examples of the, so called, \textit{Webspam} (or \textit{search spam}), i.e., web pages that are
designed to manipulate search engine results. We use the normalized trigram representation available at the LibSVM
dataset repository\footnote{Available at \url{https://www.csie.ntu.edu.tw/~cjlin/libsvmtools/datasets/}}. The original
dataset is not split into training and testing set. Therefore, following~\citep{wang2011trading} we use a random $80/20$
train/test split.

\subsection*{CIFAR-10 and CIFAR-100}

\begin{figure}[htb]
  \centering
  \includegraphics[width=0.4\linewidth]{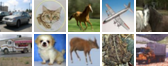}
  \caption{Example CIFAR images.}
  \label{fig:cifar_example}
\end{figure}

\texttt{CIFAR-10} and \texttt{CIFAR-100}~\citep{krizhevsky2009learning}\footnote{Available at
\url{https://www.cs.toronto.edu/~kriz/cifar.html}} are relatively small, widely used benchmarks in machine learning. The
task is to classify $32 \times 32$ RGB images~(Figure~\ref{fig:cifar_example}) across $10$ or $100$ categories (for
\texttt{CIFAR-10} and \texttt{CIFAR-100}, respectively).
The images are a subset of the $80$ million tiny images
dataset\footnote{\url{http://people.csail.mit.edu/torralba/tinyimages/}}. We preprocessed the datasets following
\citet{huang2016deep}. Specifically, we randomly extracted~$5000$ images from the training sets and used them as the
validation sets. We applied standard data augmentation steps, i.e., horizontal flipping and translation by $4$ pixels.

\subsection*{Street View House Number (SVHN)}

\begin{figure}[htb]
  \centering
  \includegraphics[width=0.4\linewidth]{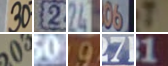}
  \caption{Example SVHN images.}
  \label{fig:svhn_example}
\end{figure}

Similarly to \texttt{MNIST}, \texttt{SVHN}\footnote{Available at \url{http://ufldl.stanford.edu/housenumbers/}} is a
digit classification dataset. We use its version that contains $32 \times 32$ cropped RGB images extracted from house
numbers from Google Street View photographs~(Figure~\ref{fig:svhn_example}).
We constructed the validation set and preprocessed the data following~\citet{huang2016deep}.

\subsection*{Synthetic datasets}

We prepared several~$10^6$-dimensional synthetic datasets, each consisting of~\mbox{$1.25 \cdot 10^6$} examples
belonging to two balanced classes. Each dataset was constructed by first generating a~$\rho$-dense matrix~$\mathbf{S}$
($\rho$ being the fraction of non-zero elements in~$\mathbf{S}$) and then selecting a fraction of features, $\phi$, that
would separate examples from the two classes. We refer to these features as the \textit{significant features}. Non-zero
elements in~$\mathbf{S}$ were drawn randomly from~$\mathcal{N}(0, 1)$. To separate the classes we picked examples from
one class and added a Gaussian noise with non-zero mean to all non-zero elements in significant features. Note that this
does not alter the sparsity of~$\mathbf{S}$. We generated two groups of such sparse datasets:
\begin{itemize}
  \item datasets with fixed fraction of significant features: $\phi = 0.2$ and density~$\rho$ ranging from $10^{-6}$ to
  $10^{-4}$,
  \item datasets with fixed density: $\rho = 10^{-4}$ and a fraction of significant features~$\phi$ ranging from $0.01$
  to $0.2$.
\end{itemize}
The above ranges for~$\rho$ and~$\phi$ were chosen so that the most difficult dataset variants had, on average, one or
two significant non-zero features per example. We randomly selected $80\%$ rows of~$\mathbf{S}$ as the training set and
the remaining $20\%$ as the test set.

%% file: publications.tex
\cleardoublepage
\phantomsection
\addcontentsline{toc}{chapter}{PhD candidate publications list}
\chapter*{PhD candidate publications list}

Below we list publications co-authored by the author of this thesis along with their MNiSW (Polish Ministry of Science
and Higher Education) classification. IF stands for Impact Factor as indexed on the Thomson Journal Citation Reports
list.

\begin{enumerate}
\setlength\parindent{24pt}

\item Karol Grzegorczyk, Marcin Kurdziel, Piotr Iwo Wójcik. 2016. Encouraging orthogonality between weight vectors in
pretrained deep neural networks. \textit{Neurocomputing}, 202 (2016), 84--90.

IF: 3.317, MNiSW: 30pts (List A)

\item Maciej Malawski, Maciej Kuźniar, Piotr Wójcik, Marian Bubak. 2013. How to use Google App Engine for free
computing. \textit{IEEE Internet Computing}, 17, 1 (2013), 50--59.

IF: 1.521, MNiSW: 40pts (List A)

\item Piotr Iwo Wójcik, Marcin Kurdziel. 2018. Training neural networks on high-dimensional data using random
projection. \textit{Pattern Analysis and Applications}, doi: 10.1007/s10044-018-0697-0.

IF: 1.352, MNiSW: 20pts (List A)

\item Piotr Iwo Wójcik, Thérèse Ouellet, Margaret Balcerzak, Witold Dzwinel. 2015. Identification of biomarker genes for
resistance to a pathogen by a novel method for meta-analysis of single-channel microarray datasets. \textit{Journal of
Bioinformatics and Computational Biology}, 13, 4 (2015), 1550013.

IF: 0.8, MNiSW: 15pts (List A)

\item Karol Grzegorczyk, Marcin Kurdziel, Piotr Iwo Wójcik. 2016. Implementing deep learning algorithms on graphics
processor units. \textit{Parallel Processing and Applied Mathematics. PPAM 2015. Lecture Notes in Computer Science},
vol. 9573, 473--482.

MNiSW: 15pts (Web of Science)

\item Karol Grzegorczyk, Marcin Kurdziel, Piotr Iwo Wójcik. 2015. Effects of sparse initialization in deep belief
networks. \textit{Computer Science}, 16, 4 (2015), 313--327.

MNiSW: 12pts (List B)

\item Chwastowski et al. 2012. The CC1 project -- system for Private Cloud Computing. \textit{Computer Science}, 13, 2
(2012), 103--111.

MNiSW: 12pts (List B)

\item Piotr Iwo Wójcik, Marcin Kurdziel. 2017. Random projection initialization for deep neural networks. \textit{25th
European Symposium on Artificial Neural Networks, Computational Intelligence and Machine Learning (ESANN'2017)},
April 26--28, 2017, Bruges, Belgium.

MNiSW: 10pts (Web of Science)

\item Joanna Kocot, Tomasz Szepieniec, Piotr Wójcik, Michał Trzeciak, Maciej Golik, Tomasz Grabarczyk, Hubert
Siejkowski, Mariusz Sterzel. 2014. A framework for domain-specific science gateways. \textit{eScience on Distributed
Computing Infrastructure. Lecture Notes in Computer Science}, vol 8500, 130--146.

MNiSW: 5pts

\item Joanna Kocot, Tomasz Szepieniec, Mariusz Sterzel, Daniel Harężlak, Maciej Golik, Tomasz Twaróg, Piotr Wójcik. 2012.
InSilicoLab: a domain-specific science gateway. \textit{Cracow'12 Grid Workshop}, October 22--24, 2012, Krakow, Poland.

\end{enumerate}